\documentclass[runningheads]{llncs}

\usepackage[utf8]{inputenc} 
\usepackage[T1]{fontenc}    
\usepackage{url}            
\usepackage{booktabs}       
\usepackage{amsfonts}       
\usepackage{nicefrac}       
\usepackage{microtype}      
\usepackage{multirow}
\usepackage{enumitem}
\usepackage{amsmath,amsfonts}
\usepackage{bm}
\usepackage{graphicx}
\usepackage{caption}
\usepackage{listings}
\usepackage{textcomp}
\usepackage{xcolor}
\usepackage{colortbl}
\usepackage{subcaption}
\usepackage{tcolorbox}
\usepackage{xspace}
\usepackage{flushend}
\usepackage{pifont}

\usepackage[ruled,linesnumbered]{algorithm2e}    
\usepackage{algpseudocode} 
\usepackage[flushleft]{threeparttable}

\newcommand{\PreserveBackslash}[1]{\let\temp=\\#1\let\\=\temp}
\newcolumntype{C}[1]{>{\PreserveBackslash\centering}p{#1}}
\newcolumntype{R}[1]{>{\PreserveBackslash\raggedleft}p{#1}}
\newcolumntype{L}[1]{>{\PreserveBackslash\raggedright}p{#1}}

\definecolor{dg}{rgb}{0,0.694,0.298}
\definecolor{purple}{rgb}{0.4,0.176,0.569}
\definecolor{colorsteps}{rgb}{0.83, 0.83, 0.83}
\definecolor{royalblue}{RGB}{65,105,225} 
\definecolor{lightgray}{HTML}{eeeeee}
\definecolor{tab_red}{rgb}{1,0.76,0.71}
\definecolor{rq}{HTML}{009444}

\usepackage{hyperref}
\hypersetup{
    colorlinks=true,
    linkcolor=red,
    citecolor=royalblue,
    filecolor=magenta,      
    urlcolor=magenta,
}

\makeatletter
\DeclareRobustCommand\onedot{\futurelet\@let@token\@onedot}
\def\@onedot{\ifx\@let@token.\else.\null\fi\xspace}
\def\eg{\emph{e.g}\onedot} 
\def\ie{\emph{i.e}\onedot}

\def\etal{\emph{et al}\onedot}
\makeatother

\begin{document}

\title{An Exploratory Study of AI System Risk Assessment from the Lens of Data Distribution and Uncertainty}
\titlerunning{An Exploratory Study of AI System Risk Assessment}
%

\author{Zhijie~Wang\inst{1}\and 
Yuheng Huang\inst{1} \and
Lei Ma\inst{1,2,3}\thanks{Corresponding author} \and\\
Haruki Yokoyama\inst{4} \and 
Susumu Tokumoto\inst{4} \and
Kazuki Munakata\inst{4}}

\authorrunning{Z. Wang, Y. Huang, L. Ma, H. Yokoyama, S. Tokumoto, and K. Munakata}
%
\institute{University of Alberta, Canada\and 
Alberta Machine Intelligence Institute (Amii), Canada\and
Kyushu University, Japan\and
Fujitsu Limited, Japan
}

\maketitle              

\begin{abstract}
In the past few years, deep learning (DL) has become a driving force and has been widely adopted in many domains and applications with competitive performance.  In practice, to solve the nontrivial and complicated tasks in real-world applications, DL is often not used standalone, but instead contributes as a piece of gadget of a larger complex system. Although there comes a fast increasing trend to study the quality issues of deep neural networks (DNNs) at the model (or unit) level, few studies have been performed to investigate the quality and reliability of DNNs at both the unit level and the potential impacts on the system level. More importantly, there is also a lack of systematic investigation on how to perform the risk assessment for data-driven AI systems from unit level to system level. 

To bridge this gap, this paper initiates an early exploratory study of AI system risk assessment from both the data distribution and uncertainty angles to address these issues. We propose a general framework and an exploratory study for analyzing AI systems. We first collect four representative AI systems from different industrial domains (\ie, {\em autonomous driving}, {\em object tracking}, {\em speech recognition}, {\em dialogue}) and utilize different common corruption patterns accordingly to enable the analysis of each AI system's robustness. Then, we explore the potential usefulness of two out-of-distribution (OOD) detection methods and two general uncertainty estimation metrics for AI systems. Based on this, we continue to leverage the OOD and uncertainty information collected from each individual module to investigate the error propagation effects inside the AI systems. Finally, we make an early attempt to explore simple combination strategies of OOD and uncertainty information to improve the reliability of AI systems. After large-scale (\ie, 700+ experimental configurations and 5000+ GPU hours) experiments and in-depth investigations, we have reached a few key interesting findings: (1) common corruption patterns can still pose threats to AI system's robustness, (2) OOD detection and uncertainty estimation techniques are still effective in analyzing different modules inside AI systems, (3) certain modules of an AI system could have a strong correlation to the system output error, and (4) even simple strategies of combining OOD- and uncertainty-analysis can significantly improve the quality and reliability of AI systems (e.g., system performance against corrupted data improved by up to 38\%). Our results highlight the practical need for more in-depth investigations into AI systems and the industrial demands for new quality assurance and risk assessment techniques for AI systems.
\end{abstract}


\section{Introduction}\label{sec:introduction}

Over the past decade, deep learning (DL) has been a driving force behind the success of many intelligent systems, across various industrial applications and domains, \eg, face recognition~\cite{wang2018cosface,deng2019arcface}, dialogue system~\cite{wen2016network,li2017end}, speech recognition~\cite{synnaeve2019end,kahn2020self}, autonomous driving \cite{bojarski2020nvidia,feng2018towards}, object tracking~\cite{li2015deeptrack,wang2015visual}. Some of these systems have already been widely deployed into real-world environments that impact our society and daily life~\cite{ravanelli2021speechbrain,peng2020first}.

With the excitement of celebrating the achievement to approach the human competitive level of intelligence by these data-driven AI systems, recent concerns are raised regarding the quality and reliability of these AI systems~\cite{Tesla,TeslaComment}, especially in the context of safety-critical domains and applications, \eg, autonomous driving~\cite{peng2020first}. Different from traditional rule-based software, the data-driven nature of the DL fundamentally changes the software development paradigm, in which the decision logic is automatically learned from the data instead of being directly coded by a human developer. However, the probabilistic nature and lacking interpretability~\cite{lipton2018mythos} make the quality and risk assessment of Deep Neural Networks (DNNs) rather difficult, even at the unit (model) level.

Recently, various techniques~\cite{pei2017deepxplore,ma2018deepgauge,kim2019guiding,xie2019deephunter,ma2018deepmutation,ma2019deepct,zhou2020deepbillboard,tian2018deeptest,zhang2018deeproad,wang2019repairing,sotoudeh2019correcting,zhang2019apricot,Yu2021,xie2021rnnrepair} have been proposed to address DNNs' quality assurance and risk assessments at the unit level. For example, quite a few studies(\eg, \cite{pei2017deepxplore,ma2018deepgauge,kim2019guiding,xie2019deephunter,ma2018deepmutation,ma2019deepct,zhou2020deepbillboard,tian2018deeptest,zhang2018deeproad}) have been working on the DNN testing, which aims to identify more inputs that a DNN can not handle before deploying it. Meanwhile, some progress (\eg, \cite{wang2019repairing,sotoudeh2019correcting,zhang2019apricot,Yu2021,xie2021rnnrepair}) has also been made on the DNN debugging and repairing, which tries to analyze why a DNN fails to make a correct prediction and then repair this DNN model accordingly. In parallel, there is another line of work on DNNs' risk assessments, where two representative types of techniques exist: (1) out-of-distribution (OOD) detection and (2) uncertainty estimation. OOD detection assesses the model's prediction risk from the \textit{data distribution} angle. This is in line with the fundamental assumption of machine learning in that only on future unseen data that share a similar distribution with the training data, the learned machine learning model could exhibit a certain level of prediction accuracy. However, in practice, the test data might now always fit well with the training data in terms of distribution, due to various reasons, e.g., data collection, operational environment changes. OOD detection could be helpful to identify and filter out those potentially risky inputs that might fall out of the training data distribution, on which the model could be incapable. In particular, OOD detection techniques~\cite{hendrycks2016baseline,liang2018enhancing,hsu2020generalized,liu2020energy,lee2018training,liang2018enhancing,kim2019guiding} are proposed to capture the relation of data instances against the distribution of training data. Based on this, developers can interpret how different the test data is compared with training data to assess the potential risk of the model's prediction on test data. Different from OOD detection, uncertainty estimation techniques~\cite{blundell2015weight,gal2016dropout,malinin2018predictive,van2020uncertainty,zhang2020towards} are also proposed to assess prediction risks from various sources of uncertainty, e.g., a simple and effective way to estimate the uncertainty for models' prediction via Bayesian inference.
By interpreting the uncertainty score, developers can be aware of the confidence in a model's prediction in terms of uncertainty, before making the decision to accept or reject the model's prediction.

Although these DNN quality assurance and risk assessment techniques have been shown to be useful at the unit (model) level. To the best of our knowledge, limited studies have been on the risk assessment of DNN-enabled AI systems at the system level. So far, to the best of our knowledge, it is even not clear what could be a promising direction for quality assurance and risk assessment of AI systems, and whether existing techniques at the unit level could still work at a higher system level, and to what extent.
In practice, real-world AI applications are usually not only based on a single standalone DNN model. 
Instead, one or multiple DNN models are often used as the components, forming the parts of a larger system~\cite{zhu2020ConvLab,udacitychallenge,zhang2021bytetrack,ravanelli2021speechbrain} that often also contain the traditional software components. Some simplified common patterns that an AI system constitutes could be: 

\begin{itemize}[leftmargin=*]
	\item Multiple ML/DL models are designed to process input in parallels, the results of which are aggregated as the system to solve one specific task (\eg, Fig.~\ref{fig:sys_arch} (A)).
	\item Multiple ML/DL models are used to form a task pipeline, the input of which ML/DL model could depend on a former one, together solving a complex task (\eg, Fig.~\ref{fig:sys_arch} (B)).
	\item ML/DL models are integrated with traditional software components with complex internal interactions (\eg, Fig.~\ref{fig:sys_arch} (C) and (D)).
\end{itemize}

A recent study~\cite{peng2020first} has shown that a real-world AI system, \eg, Baidu Apollo autonomous driving system~\cite{apollo} that includes 30+ ML/DL models and 10+ traditional software components, can be as complex as covering all these circumstances.
The complexity of AI systems hence brings more challenges to their analysis, compared to the model level. How to understand and analyze the behavior of individual modules as well as different modules' interactions inside a complex AI system, and how to perform the risk assessment could be some of the key pillars towards building reliable and trustworthy AI systems. However, to the best of our knowledge, there is no extensive study on analyzing the AI systems at the system level. In particular,

\begin{itemize}[leftmargin=*]
    \item \textbf{AI systems often have limited availability.} Although DL techniques and AI systems built with such techniques have been used in various domains for a long period, most AI systems used by industry could be intellectual properties and not open-source. As the key artifacts of an AI system, training data played an essential role in the development and decision logic learning of models, however, is often not publically available due to commercial and privacy restrictions. Therefore, collecting open-sourced AI systems for the study is not easy, especially for those with training data.
    \item \textbf{No general techniques have been established for the analysis of complex AI systems at the system level.} Existing work of quality assurance mainly focuses on analyzing a single DNN (unit-level). However, there still lack study to investigate their effectiveness in analyzing a complex AI system at the system level.
    \item \textbf{Evaluating the AI system's capability in the wild could be challenging.} AI system's performance could significantly drop after being deployed to real-world environments due to the data distribution shift between training data and test data~\cite{koh2021wilds}. Therefore it is important to understand the potential impacts from the operational environment to the system prediction capability and risks. However, collecting and labeling new data for an existing system with pre-trained ML/DL models could be both hard and time-consuming.
\end{itemize}

Therefore, towards addressing these issues and enabling further research on AI systems, in this paper, we propose  an exploratory study of AI system risk assessment. As an early attempt, this paper performs analysis and risk assessment of an AI system through characterizing data distribution and uncertainty inside the system. To be specific, we first investigate a wide range of AI-enabled systems (mainly based on the criteria of artifact availability) and come to four representative ones from different tasks and data domains: ConvLab-2~\cite{zhu2020ConvLab}, SpeechBrain~\cite{ravanelli2021speechbrain}, Udacity~\cite{udacitychallenge}, and ByteTrack~\cite{zhang2021bytetrack}. Since collecting and labeling new data could be challenging, we make use of common corruption patterns in different domains' data to synthesize real-world corrupted datasets for evaluating the AI system's performance (\ie, robustness) in the wild. Then, to utilize the existing OOD detection techniques and uncertainty estimation more generally, we propose two general OOD detection and two general uncertainty estimation metrics techniques. Overall, we mainly investigate the following research questions and identify several challenges and potential opportunities for AI system's risk assessment:
\ding{202} \textbf{What are the potential impacts of common corruption patterns on AI systems in the wilds?} We first evaluate AI systems' performance degradation against different real-world corruption patterns. Our experimental results show that these corruption patterns could significantly affect AI systems' performance, indicating an urgent need for addressing AI systems' robustness issues, which could be of great importance, especially for those AI systems that are designed to be deployed in the wild with various noisy factors in the operational environment. 
\ding{203} \textbf{What is the OOD awareness of different modules in AI systems?} We continue to study the effectiveness of two proposed general OOD detection methods based on the corrupted datasets by comparing the OOD scores on clean and corrupted data. Our experimental results reveal that the proposed OOD detection methods could detect data distribution changes, showing the potential for assessing the AI system's prediction risks. Our investigation results also reveal that different modules in an AI system could have different sensitivities to OOD data.
\ding{204} \textbf{How does the uncertainty estimation perform on different modules in AI systems?} In addition to OOD detection, we further study the effectiveness of two proposed general uncertainty estimation metrics on different AI systems by comparing clean data's uncertainty scores and corrupted counterparts. We found that the proposed metrics could well reflect the AI system's prediction uncertainty on the corrupted datasets through large-scale experiments. Furthermore, we also find that one module’s uncertainty score change does not always correlate to another module’s uncertainty score change in an AI system.
\ding{205} \textbf{How do the errors propagate among modules in AI systems?} In practice, when AI systems fail to make correct predictions, it is often of great importance to analyze to find out the potential reasons, \eg, whether a single module has triggered the bug or if there is any erroneous information propagation inside the AI system. Based on the OOD detection and uncertainty estimation results from RQ2~\&~3, we further perform an in-depth analysis of how each module's OOD and uncertainty information in an AI system has correlated to the output error. Overall, we find that not every module's OOD/uncertainty information strongly correlates to the system output error. 
\ding{206} \textbf{How do simple combinations of OOD analysis- and uncertainty-based method improve the reliability of AI systems?} Finally, we make an early exploratory investigation on whether using OOD score or uncertainty information could assess an AI system's prediction risks, and further reduce the system prediction risks by rejecting those uncertain and unreliable cases. Surprisingly, our results indicate that even a simple combination strategy could enhance the system performance by up to 38\%.

In summary, this paper makes the following contributions:

\begin{itemize}[leftmargin=*]
    
    \item {We design a research process for the analysis and risk assessment of AI systems at the system level. Then we conduct an exploratory study on several AI systems based on this process. In order to analyze system quality and reliability, we leverage common corruption patterns from different domains to generate and prepare multiple large-scale corrupted datasets to facilitate our study.}
    
    \item We study the internal mechanism (\ie, interactions among different modules inside a system) and the external mechanism (\ie, the system's performance against corrupted data) of AI-enabled systems from the lens of data distribution and uncertainty through large-scale experiments (total experimental time is more than 5,000 GPU hours).
    
    \item As an early attempt, we utilize five simple strategies of combining OOD- and uncertainty-analysis to make risk assessments for AI systems.
    
    \item Based on our study results and in-depth analysis, we further propose discussions on the implications of our exploratory study as well as several future research opportunities towards building reliable and trustworthy AI systems, \eg, system-level quality assurance, fault localization in AI system.
\end{itemize}

To the best of our knowledge, this is an early paper that initiates an exploratory study on analyzing complex AI systems at the system level. Our empirical study results reveal the current status of analyzing AI systems, demonstrating multiple demands and potential research opportunities around AI systems. The results of this study call the attention of researchers and industrial practitioners, that similar to the traditional software system, there could be many challenges and potential issues that need to be addressed at a higher level of the AI-enabled systems, instead of only focusing on the unit (model) level, where most existing work centers around.
We hope our work could enable better understanding, establish the basis, and pave the path towards further quality assurance research to build reliable and trustworthy AI systems. 

\section{Related Work}\label{sec:related}

In this section, we discuss previous work that could be most relevant to this paper, \ie, {\em unit-level AI module analysis}, along with three directions: (1) out-of-distribution detection, (2) uncertainty estimation, and (3) testing, debugging and repairing of DNNs. Making risk assessments for a standalone DNN has been studied for many years. Existing work that is relevant to our study mostly focuses on a single neural network with one specific task like image classification. However, few works have been made for AI systems involving multiple AI-enabled modules as well as traditional software. In the following, we survey related work for unit-level AI module analysis and try to find those applicable for analyzing complex AI systems. 

\subsection{Out-of-distribution detection.}

One fundamental assumption of machine learning is that training and test data share a similar distribution. In other words, machine learning often does not make any guarantee for those data that fall out of the distribution of training data. Therefore, DNNs often fail to offer correct predictions for out-of-distribution (OOD) data and can even provide high-confidence predictions even if they're incorrect~\cite{goodfellow2014explaining,amodei2016concrete}. OOD detection aims to alleviate this problem by detecting test data that are distributionally different from the training data. Generally speaking, OOD detection seeks to compute an OOD score for a given input instance [61], and if, it is below a defined threshold, it is in-distribution (ID) data. Otherwise, it is out-of-distribution (OOD) one~\cite{morteza2022provable}. The score computation is thus important for OOD detection, and it can be roughly categorized into two types: (1) methods requiring auxiliary data, and (2) methods that can be applied for pre-trained models without extra data.

If the model is accessible at the training stage and the training data are given, regularization specifically designed for OOD detection can be employed so that the OOD score obtained from the model can better handle OOD cases~\cite{geifman2019selectivenet,malinin2018predictive,lee2018training,papadopoulos2021outlier,hendrycks2018deep}. However, in this study, as we aim to analyze existing AI systems before deployment, redesigning the models or retraining the models is beyond our scope, making such OOD approaches inappropriate for our study.

There is also some work focusing on analyzing pre-trained DNN models. Hendrycks~\etal\cite{hendrycks2016baseline} first establishes a common OOD detection baseline by directly analyzing the softmax probability at the last layer. Along the line, there are multiple methods trying to improve OOD detection~\cite{liang2018enhancing,hsu2020generalized,liu2020energy}. One major limitation of these methods is that they specifically focus on the image classification problem only, while the systems we study involve multiple input domains and have various tasks. There is also some work trying to extract hidden features from intermediate layers of DNNs. Lee~\etal\cite{lee2018simple} propose to use {\em Mahalanobis distance} to measure intermediate hidden features of ID data and OOD data. Kim~\etal\cite{kim2019guiding} propose {\em Surprise Adequacy} in which {\em kernel density estimation} is used to estimate the probability density function of ID data and OOD data. Once the estimation is done, we can know how surprised a model is for a given input and thus discriminate ID and OOD cases. Although they originally targeted classification tasks, these two metrics can be generalizable to either regression or other tasks as well since they pay attention to the intermediate output instead of classification confidence at the last layer. In our work, we apply {\em Surprise Adequacy} and {\em kernel density estimation} to identify possible OOD cases at test time. However, we would like to emphasize that OOD cases are just one aspect of risk assessments in our study. There still exist possible ID (in-distribution) test instances while the systems make erroneous predictions. 
 
\subsection{Uncertainty estimation techniques.}
Estimating uncertainty inherent in DNNs can help developers understand how much confidence can be put in the models. Uncertainty analysis can help identify not only OOD cases~\cite{ulmer2021know} but also probably erroneous predictions for ID cases~\cite{blundell2015weight}. In the past few years, uncertainty quantification methodologies have been studied extensively for DNNs~\cite{gawlikowski2021survey,abdar2021review}. Gawlikowsk~\etal~\cite{gawlikowski2021survey} categorized these methods into four types: single deterministic methods, Bayesian methods, ensemble methods and test-time augmentation methods. 

{\em Single deterministic} and {\em ensemble} methods require extra attention for uncertainty measurement early at the model architecture design stage. The former involves training a model explicitly to quantify uncertainties~\cite{sensoy2018evidential,malinin2018predictive} or using additional components to estimate the uncertainty~\cite{raghu2019direct,ramalho2020density}. The latter needs the inference result of several different networks trained for one task to quantify the uncertainty~\cite{lakshminarayanan2017simple,ovadia2019can,gustafsson2020evaluating}. As our goal is to measure the uncertainty of existing complex AI systems, we do not assume there is any explicit design for uncertainty or ensemble in these systems, and thus these methods are unsuitable. 

In contrast, the {\em test-time augmentation} and {\em Bayesian} methods can both be non-parametric and extensible.
Test-time augmentation usually generates $T$ augmented examples per input and feeds them to DNN so that a predictive distribution can be obtained to measure the uncertainty. The augmentation can be achieved through traditional data augmentation techniques (e.g., image rotation, translation). { While this augmentation is easy to implement, it is usually used in the image domain~\cite{wang2019aleatoric,ayhan2018test} and the quality of the generated images can be hard to control if the changes are made inappropriately~\cite{shanmugam2020and}. For example, rotating or translating an image might result in the main part of an image being outside of the image's bounds. Adding too much noise might block the main part of an image, and the augmented image is thus invalid since even humans cannot easily interpret it.} As we are studying AI systems with different input domains, this kind of method is not generalizable as well. Bayesian methods estimate uncertainty inherent in inference by predicting the output distribution rather than offering a deterministic result. Although Bayesian modelling often requires Bayesian Neural Networks (BNNs) for the uncertainty measurement, Gal~\etal\cite{Gal2016Bayesian} propose an easy yet effective technique called  Monte Carlo Dropout (MC Dropout) to approximate Bayesian inference for any neural networks with dropout layers. This measurement calculates uncertainty estimates by enabling dropout layers at test time and approximating variational inference. As it is applicable for a wide range of DNNs and has been proved successful in identifying adversarial and misclassified examples~\cite{zhang2020uncertainty,zhang2020towards,ma2021test}, we believe it is more suitable for our study and thus take it as the primary approach to measure uncertainty. Although MC dropout is applicable for any DNNs with dropout layers, no previous work uses it to access the uncertainty for classical logical components inside AI systems and how this impacts the system behavior. However, this kind of component plays a crucial role in some AI systems like tracking and dialogue generation. Our study performs the first in-depth analysis of the uncertainty measurement of these components and their potential impacts on the whole system.

\subsection{Testing, debugging and repairing of DNNs} 

In recent years there has been an ongoing effort to improve the quality and robustness of DNNs from the perspective of software engineering. DNN testing aims to reveal the vulnerabilities of DNNs before deployment. Instead of solely relying on the accuracy of test samples to determine whether the models are ready to deploy, various techniques have been proposed to probe the boundaries of DNNs by looking inside the black boxes. For example, a series of coverage criteria specifically designed for DNNs have been proposed to guide the testing and help understand the behavior of DNNs~\cite{pei2017deepxplore,ma2018deepgauge,kim2019guiding}. Based on this, some testing techniques are developed to synthesize test inputs to thoroughly test the DNNs~\cite{xie2019deephunter,ma2018deepmutation,ma2019deepct}. Following this line of work, multiple test case generation techniques for specific application scenarios are proposed~\cite{zhou2020deepbillboard,tian2018deeptest,zhang2018deeproad}. One major difference between our work and theirs is that we study DNNs across different domains at the system level, while they mainly focus on unit-level AI modules under certain scenarios. DNN debugging and repairing move one step further, trying to repair DNN models after finding why they fail to make correct predictions. Some previous work proposes to change model architecture directly. For example, Wang \etal~\cite{wang2019repairing} repair the model by integrating an additional layer to pre-process error inputs, Sotoudeh \etal~\cite{sotoudeh2019correcting} and Zhang \etal~\cite{zhang2019apricot} correct DNNs by changing network weights. From another perspective, error-guided data augmentation can also help repair through re-training/fine-tuning. For example, DeepRepair~\cite{Yu2021} and RNNRepair~\cite{xie2021rnnrepair} synthesize new training data by analyzing errors made by the trained model. Our work is orthogonal to them in two aspects—first, our work focus on the risk assessment at the system level instead of the module level. Instead of proposing a method to improve the robustness of DNNs, we aim to reveal the effectiveness and limitations of current approaches and identify the possible challenges to guide future research at the AI system level.

\section{Overview of Study Design}\label{sec:design}
\begin{figure}[t]
    \centering
    \includegraphics[width=\textwidth]{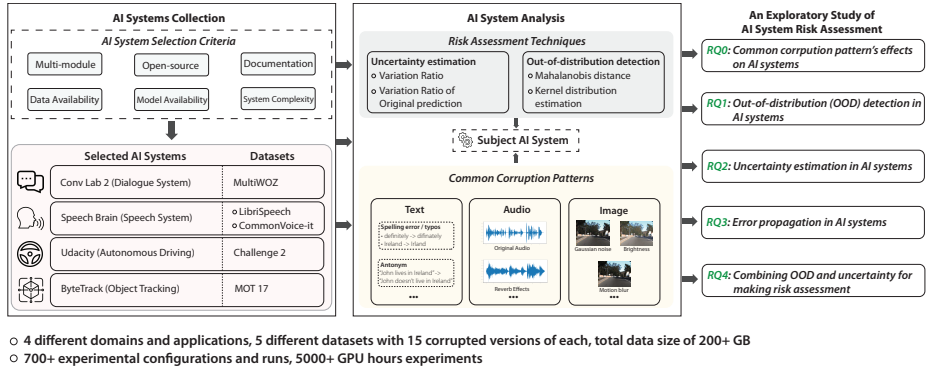}    
    \caption{Overview of our study workflow. } 
    \label{fig:workflow}
\end{figure}

In this section, we introduce the study design in this paper. Figure~\ref{fig:workflow} shows the overview of workflow our proposed framework and its three major components. The first step is the collection of AI systems (the left-side of Fig.~\ref{fig:workflow}). As discussed in Sec.~\ref{sec:introduction}, collecting suitable AI systems for our study could be challenging. Therefore, we mainly focus on selecting AI systems covering various tasks and applications in our study. The middle part of Fig.~\ref{fig:workflow} illustrates this study's analysis techniques. According to domains of selected AI systems, we first use {\em common corruption patterns} to synthesize large-scale corrupted datasets, then we use risk assessment techniques from two different aspects: (1) out-of-distribution (OOD) detection and (2) uncertainty estimation. The last component (the right-side of Fig.~\ref{fig:workflow}) summarizes the research questions to be investigated in our study. With the corrupted datasets, we first examine how AI systems can be risky in the wilds (\textcolor{rq}{RQ0}). Since the corrupted datasets are usually with shifted distribution, we then investigate the effectiveness of OOD detection in the AI systems (\textcolor{rq}{RQ1}), followed by the effectiveness of uncertainty estimation (\textcolor{rq}{RQ2}). These two techniques form the basis for our next step's investigation of error propagation in the AI systems (\textcolor{rq}{RQ3}) and make it possible to make an early attempt on the AI system's risk assessments (\textcolor{rq}{RQ4}).

Now we summarize the design of each part in our study in Sec.~\ref{subsec:aisystems}, \ref{subsec:analysis_techniques}, and \ref{subsec:research_questions}, respectively. We also introduce the details of our experimental configurations in Sec.~\ref{subsec:experimental_configurations}.

\subsection{Subject AI Systems}
\label{subsec:aisystems}

\begin{figure}[t]
    \centering
    \includegraphics[width=0.7\textwidth]{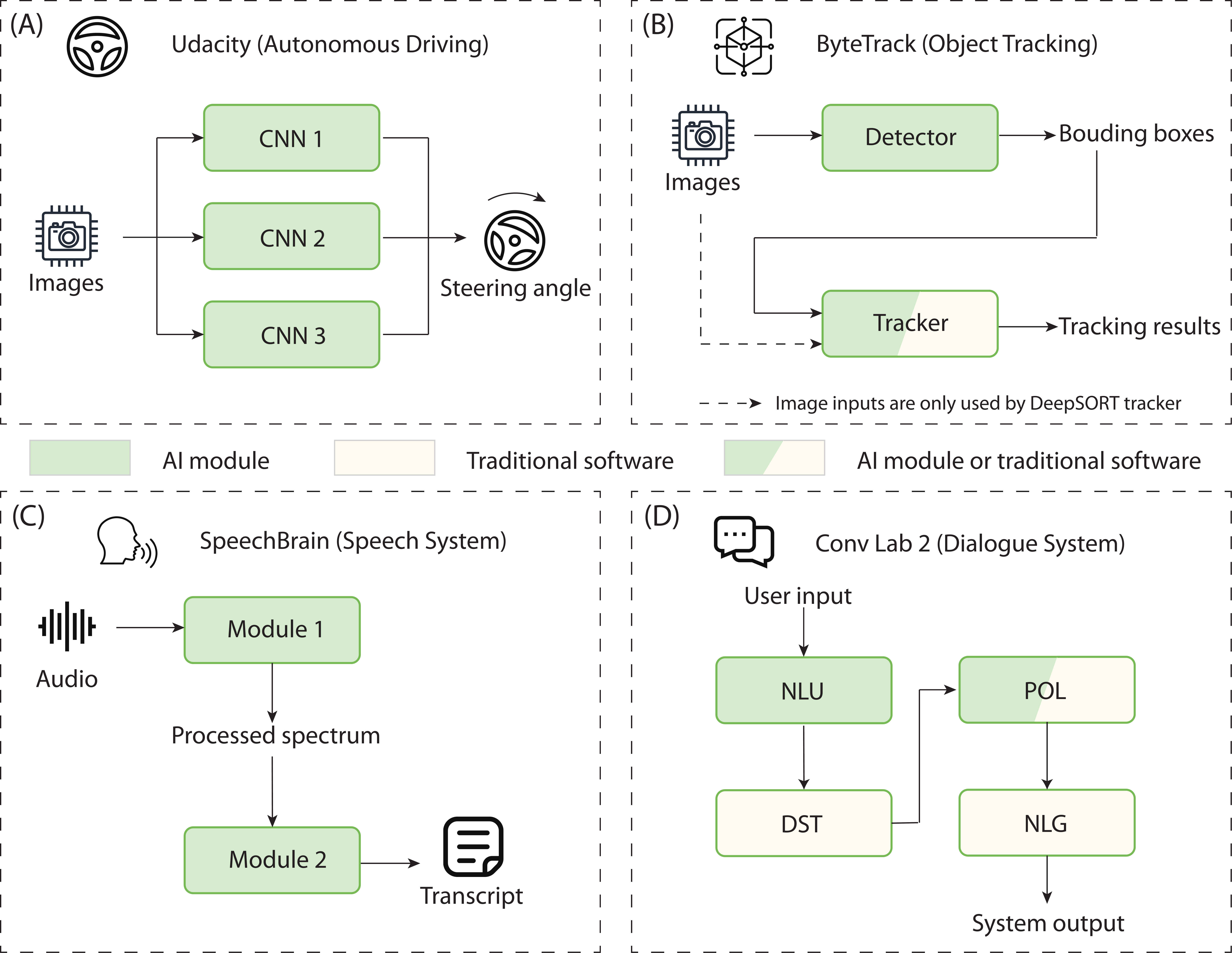}
    \caption{Simplified workflows of four subject AI systems used in this study.} 
    \label{fig:sys_arch}
\end{figure}

In real-world applications, multiple AI models are usually combined with traditional software to together solve complicated tasks as a complex system. These AI-enabled systems are explicitly designed for different tasks and application scenarios. To conduct a relatively comprehensive study of these AI-enabled systems, we try to cover a wide range of tasks including (1) autonomous driving, (2) object tracking, (3) speech recognition, and (4) dialogue system. Fig.~\ref{fig:sys_arch} shows the simplified workflows of each subject system. These four tasks also cover the three different data domains: image, text, and audio. As most AI systems are treated as private intellectual properties and are kept confidential (especially for the artifact of training data), it could be rather challenging to collect many subject AI systems for our study. In this paper, we mainly collect systems that (1) have been used in previous work and (2) have promising performance on related tasks, (3) have their key artifacts publicly available. Specifically, we select subject AI systems from different domains based on the following criteria:

\begin{itemize}[leftmargin=*]
    \item {\bf Multi-module}. An AI system should have multiple modules instead of a single AI model, where at least one module should be an AI-enabled one.
    
    \item {\bf Open-source}. An AI system must be open-sourced so that we can access AI models (\eg, training program, model architecture) inside it to quantitatively evaluate the output of every single module in the system.
    
    \item {\bf Documentation}. An AI system must have comprehensive documentation so that we can understand its system structure and pipelines.
    
    \item {\bf Data availability}. An AI system must have corresponding publicly available data (\eg, training data and test data).
    
    \item {\bf Pre-trained model availability}. An AI system should have pre-trained models available. With the pre-trained models released by authors, we can ensure the performance of each AI system and reduce the reproduction efforts.
    
    \item {\bf Feasible system complexity}. As for an early exploratory study, the system should not be too complex to make the evaluation feasible and reasonably achievable, from the perspectives of our available computational resources and time.
\end{itemize}

Next, we briefly introduce each of our subject AI systems. 

{
    \subsubsection{Autonomous Driving System} In recent years autonomous driving has attracted much attention in both academia and industry. One of the major tasks of autonomous driving is steering angle control. NVIDIA first proposed a simple CNN to solve the problem in 2016~\cite{bojarski2016end}, representing one of the early works to bring the power of DNNs to autonomous driving. Afterwards, they publish a report demonstrating a more sophisticated system called PilotNet~\cite{bojarski2020nvidia}. Along the line, there are many efforts to tackle the autonomous steering problem~\cite{du2019self,chi2017deep,yang2018end,hou2019learning,yang2018real}. 
}

{\noindent \textbf{The system we choose: Udacity.}} Udacity system~\cite{udacitychallenge} is derived from a well-known self-driving competition \textit{Udacity self-driving challenge 2}. The goal of this task is to predict autonomous driving steering angles. Specifically, it takes two consecutive images as input and outputs the steering angle. However, the control decisions made by the systems built on this challenge are independent across different frames, and as a result, the dependencies between adjacent images are ignored. Because of this, we can take them as systems in image processing. In this work, we select the ``Rambo'' system (which ranked 2nd in this competition) as the candidate system for analysis. { This system is a parallel-executed system where three separately trained CNNs (see Fig.~\ref{fig:sys_arch} (A)) process two consecutive images individually and predict the steering angle value independently. After then, their prediction results are merged as the final output by taking average.} The three CNNs are trained separately and therefore we consider this as a minimal and simple example of an AI system as it would be interesting to see what the results might be for simple and complex cases of an AI system in this study. The performance of the Udacity system is measured by \textit{root mean square error} (RMSE, Eq.~\ref{eq:rmse}) between the predicted steering angle value and ground truth value. 

\begin{equation}
    \label{eq:rmse}
    RMSE = \sqrt{\frac{\sum_{i=1}^N (\hat{y}_i-y_i)^2}{N}}
\end{equation}
where $\hat{y}_i$ and $y_i$ denote the predicted and ground truth steering angle value, respectively. $N$ denotes the total number of images.

{
    \subsubsection{Object Tracking System} Multi-object tracking (MOT) is one of the popular domains nowadays in the object tracking field. It involves automatically identifying multiple objects in a video and often requires one module to detect the objects and another module to track the objects. The detector is often a powerful object detection model~\cite{lin2017focal,zhou2019objects,redmon2018yolov3,ge2021yolox}, while the tracker can be either DNN-based~\cite{wojke2017simple,liang2020rethinking} or solely powered by traditional software~\cite{zhang2021bytetrack,bewley2016simple}.
}

{\noindent \textbf{The system we choose: ByteTrack.}} ByteTrack system~\cite{zhang2021bytetrack} is designed for object tracking, which has two serial-executed modules: (1) object detector, and (2) object tracker. The object detector takes two consecutive images as input and outputs bounding boxes for object instances detected in each image. The object tracker then takes the bounding boxes detected in two images as input and outputs the association between the two images' bounding boxes. Unlike the Udacity challenge systems, ByteTrack explicitly records the states of multiple objects across all frames in order to precisely track each of them in the video. As the cross-frame dependencies are considered, the target system is thus more complex, representing systems in the video processing. Note that when using DeepSORT as the tracker in the ByteTrack system, the tracker also takes images as input (dashed line in Fig.~\ref{fig:sys_arch} (B)). The performance of the ByteTrack system is measured by \textit{multiple object tracking accuracies} (MOTA, Eq.~\ref{eq:mota})~\cite{bernardin2006multiple}.

{
\begin{equation}
    \label{eq:mota}
    MOTA = 1 - \frac{\sum_t(m_t+fp_t+mmme_t)}{\sum_tg_t}
\end{equation}
where $m_t$, $fp_t$ and $mme_t$ are the number of misses (the tracker misses an object), of false positives (the tracker reports non-existing objects) and of mismatches (the tracking result between two bounding boxes is incorrect) respectively for time $t$. $g_t$ denotes the number of groundtruth bounding boxes tracked at time $t$.} For two consecutive frames, the equation first computes the error rate by taking three different possible errors (i.e., miss, false positive, and mismatch) into account over the number of objects, then performs ``1-\text{error rate}'' to calculate the MOTA.

\subsubsection{Speech Recognition System} Speech recognition is to identify an acoustic input sequence and convert it into a text transcript. Early in 2011, Dahl \etal~\cite{dahl2011context} proposed (CD)-DNN-HMM as an end-to-end system that involves DNN to do the Automatic Speech Recognition (ASR) task. Since then, using deep learning as a key component in ASR systems has been widely studied. Deepspeech series~\cite{hannun2014deep,amodei2016deep}, as an end-to-end method, has achieved great success and is still widely adopted now. As an alternative, employing the encoder-decoder structure with a classifier can be more modularized~\cite{graves2014towards,graves2012sequence,watanabe2017hybrid,baevski2020wav2vec}. We prefer this solution as we can study the relationship between modules more easily. 

{\noindent \textbf{The system we choose: SpeechBrain.}} SpeechBrain system~\cite{ravanelli2021speechbrain} is a toolkit for processing multiple speech-oriented tasks. It offers various kinds of ASR systems with two-stage architectures (as shown in Fig.~\ref{fig:sys_arch} (C)) where the first module process the speech spectrum, and the second module transforms the processed spectrum into text. The input to SpeechBrain is audio which involves more complex pre-processing and post-processing compared to image input. The competitive performance of SpeechBrain makes it an ideal choice to study systems in \textit{audio processing}. The performance of the SpeechBrain system is measured by \textit{word error rate} (WER, Eq.\ref{eq:wer}) between the predicted text and ground truth transcript. 


\begin{equation}
    \label{eq:wer}
    WER=\frac{S+D+I}{N}
\end{equation}
where $S, D, I$ are the number of substitutions, the number of deletions, and the number of insertions, respectively. $N$ is the number of words in the reference. {Note that WER can also be seen as an edit distance.}

\subsubsection{Dialogue System} Dialogue generation aims to generate a response given a post by the user automatically in order to finish specific tasks, \eg, recommendation. A dialogue system involves a complex Natural Language Processing (NLP) task. Wen~\etal~\cite{wen2015semantically} made an early attempt to use LSTM to solve it. Recently it has experienced rapid growth. Nowadays, an AI-enabled dialogue system usually includes: (1) a natural language understanding component~\cite{mairesse2009spoken,lee2019convlab} which is used to parse the users' intent, (2) a dialogue policy~\cite{takanobu2019guided}, which decides the response of the system, and (3) a natural language generation component~\cite{wen2016network} which is used to transform dialogue acts into a natural language sentence. 

{\noindent \textbf{The system we choose: ConvLab2.}} ConvLab2 system~\cite{zhu2020ConvLab} is a dialogue system that has four connected modules (Fig.~\ref{fig:sys_arch} (D)): NLU (natural language understanding), DST (dialogue state tracking), POL (dialogue policy), and NLG (natural language generation). NLU is used to parse the user’s intent, takes an utterance as input and outputs the corresponding dialogue acts. DST updates the belief state, which contains the constraints and requirements of the user. POL receives the belief state and outputs system dialogue acts. NLG transforms dialogue acts into a natural language sentence. Note that the ConvLab2 system is a dialogue recommendation system, where each dialogue is oriented with a specific task, e.g., hotel booking. The performance of the ConvLab2 system is measured by whether the system agent (who was receiving the call) successfully performs a specific task in a dialogue (\textit{success rate}, Eq.~\ref{eq:sr}). Suppose the task for a dialogue is hotel booking. If the agent successfully record the information about hotel booking (e.g., preferred day and hotel name), then the task is seen as successful.

\begin{equation}
    \label{eq:sr}
    success~rate = \frac{|D|_{success}}{|D|}
\end{equation}
where $|D|_{success}$ and $|D|$ are the number of successful dialogues and total dialogues, respectively.

\subsection{Analysis Methodology}
\label{subsec:analysis_techniques}

Our analysis methodology for AI systems consists of two parts (1) Synthesizing real-world corrupted datasets by common corruption patterns of the specific input domain. (2)Based on the pair-wise data of clean and corrupted data, we then leverage two different aspects of risk assessment techniques: {\em out-of-distribution detection} and {\em uncertainty estimation} as the second part of our analysis.

\begin{table}[htbp]
\caption{Summary of corruptions used in the proposed study and its corresponding parameters.} 
\scriptsize
\centering
\def\arraystretch{1.5}
\begin{tabular}{L{0.07\textwidth}L{0.15\textwidth}L{0.25\textwidth}L{0.42\textwidth}}
    \toprule
    \textbf{Domain}    &   \textbf{Corruption}  & \textbf{Effects} & \textbf{Parameters} \\ \midrule
    \multirow{9}{*}{Image}   &    Gaussian noise (GN) & Add Gaussian noise pattern on an image. & $c \in \left\{0.08, 0.18, 0.38\right\}$, where
    $c$ is the scale of noise (variance of a zero-mean Gaussian distribution). \\
    & \cellcolor{lightgray}Motion blur (MB)  &  \cellcolor{lightgray}Add motion blur effect on an image. & \cellcolor{lightgray}$(c_0, c_1) \in \left\{(10,3), (15,8), (20,15)\right\}$, where $c_0$ and $c_1$ denote the radius and the scale of motion blur.  \\
    & Snow (SN)   & Add snow effect on an image. & $(c_0,c_1,c_2,c_3,c_4) \in~ $\{$(0.1, 0.3, 3, 0.5, 10, 4, 0.8)$, $(0.55, 0.3, 4, 0.9, 12, 8, 0.7)$, $(0.55,0.3,2.5,0.85,12,12, 0.55)$\}, where $c_i$ is the parameter for snow effects.    \\ 
    & \cellcolor{lightgray}Contrast (CO)   & \cellcolor{lightgray}Change the contrast of an image.  & \cellcolor{lightgray}$c\in\left\{0.4, 0.2, 0.05\right\}$, where $c$ is the faxwctor when changing image's contrast. \\
    & Pixelate (PI)   & Add pixelate effect on an image.  & $c\in\left\{0.6, 0.4, 0.25\right\}$, where $c$ is the factor of pixelization.\\
    \midrule
    \multirow{7}{*}{Audio}   & Reverb (RE) & Add reverb effect to an audio clip. & $c\in\left\{0.33,0.67,1.0\right\}$, where $c$ is the wet-level of reverb effect.\\
    & \cellcolor{lightgray}Speed (SP) & \cellcolor{lightgray}Change the speed of an audio clip. & \cellcolor{lightgray}$c\in\left\{0.8, 1.5, 2.0\right\}$, where $c$ is the factor of speed.\\
    & Distortion (DI) & Add distorion effect to an audio clip. & $c\in\left\{15, 20, 30\right\}$, where $c$ is the level of distortion.\\
    & \cellcolor{lightgray}Background noise (BN) & \cellcolor{lightgray}Add Gaussian white noise to an audio clip. & \cellcolor{lightgray}$c\in\left\{10, 5, 2\right\}$, where $c$ is the target signal-to-noise ratio.\\
    & Pitch shift (PI) & Shift the pitch of an audio clip. & $c\in\left\{2, 4, 6\right\}$, where $c$ is the number of semitones when shifting an audio.\\
    \midrule
    \multirow{10}{*}{Text}    & Digits2words (DW) & Change the digits in a text corpus into corresponding words. & $c\in\left\{0.2, 0.6, 1.0\right\}$, where $c$ is the probability to change.\\
    & \cellcolor{lightgray}Remove char (RC) & \cellcolor{lightgray}Randomly ($p=0.1$) remove a character in a word in a text corpus. & \cellcolor{lightgray}$c\in\left\{0.15, 0.45, 0.75\right\}$, where $c$ is the probability to change a word in a text corpus.\\
    & Change char (RC) & Randomly ($p=0.1$) replace a character with similar one in a word in a text corpus. & $c\in\left\{0.15, 0.45, 0.75\right\}$, where $c$ is the probability to change a word in a text corpus.\\
    & \cellcolor{lightgray}Misspelling (MS) & \cellcolor{lightgray}Replace a word with its commonly misspelling one in a text corpus. & \cellcolor{lightgray}$c\in\left\{0.2, 0.6, 1.0\right\}$, where $c$ is the probability to replace a word in a text corpus.\\
    & Swap (SW) & Randomly swap two characters in a word in a text corpus. & $c\in\left\{0.15, 0.45, 0.75\right\}$, where $c$ is the probability to swap a word in a text corpus.\\
    \bottomrule
\end{tabular}
\label{tab:corruption}
\end{table}

\subsubsection{Common corruption patterns.} Common corruption patterns~\cite{hendrycks2019robustness} refer to patterns that (1) naturally exist in the real world and (2) could significantly affect DNN's performance. AI models' performance against such common corruption patterns could reflect their robustness and help developers assess potential risks before deploying to complex real-world environments. Therefore, this paper adopts a similar idea to analyze AI systems by using different corruption patterns for data from different domains. We give the introductions of corruption for each domain of data in Table~\ref{tab:corruption}.

\begin{itemize}[leftmargin=*]

    \item {\bf Image data corruption}. Among 15 common image corruption patterns proposed by Hendrycks \etal~\cite{hendrycks2019robustness}, we select the following five representative ones: \textbf{Gaussian noise (GN)}, \textbf{motion blur (MB)}, \textbf{snow (SN)}, \textbf{contrast (CO)}, and \textbf{pixelate (PI)}. These five corruption patterns are selected from different categories in order to make our selection diverse enough. \textbf{GN} and \textbf{MB} are common noise patterns and blur effect, respectively. \textbf{SN} is related to a natural weather effect while \textbf{CO} usually occurs due to the changes in light conditions. The last one \textbf{PI} usually occurs during the image up-sampling process.
    
    \item {\bf Audio data corruption}. We select the following five audio corruption patterns: \textbf{reverb (RE)}, \textbf{speed (SP)}, \textbf{distortion (DI)}, \textbf{background noise (BN)}, and \textbf{pitch shift (PS)} from~\cite{papakipos2022augly,pedalboard2022}. \textbf{RE} adds reverb effect to an audio clip. \textbf{SP} changes the speed of an audio clip. \textbf{DI} adds distortion effects to the audio clip. \textbf{BN} adds Gaussian noise to the audio clip. \textbf{PS} shifts the pitch of an audio clip.
    
    \item {\bf Text data corruption}. We select the following five text corruption patterns among 11 text corruption patterns proposed by Rychalska \etal~\cite{rychalska2019models}: \textbf{digit2words (DW)}, \textbf{remove chars (RC)}, \textbf{change chars (CC)}, \textbf{misspelling (MS)}, and \textbf{swap (SW)}. \textbf{DW} replaces the numbers in a text corpus as corresponding words. \textbf{RC} randomly removes a character of some words in a text corpus, while \textbf{CC} randomly replaces a character with another. \textbf{MS} replaces words with common misspelling ones. \textbf{SW} swap two characters in a word.

\end{itemize}

\subsubsection{OOD detection technique.} 

As introduced in Sec.~\ref{sec:related}, many of the existing OOD detection techniques are designed specifically for classification problems~\cite{hendrycks2016baseline,liang2018enhancing,hsu2020generalized,liu2020energy}. However, some of the related work relies on measuring the distribution difference of the intermediate representations (i.e. hidden states in DNNs) to determine whether the given input is OOD or not. As intermediate representations are generic across classification and regression tasks, in this paper, we adopt Surprise Adequacy~\cite{kim2019guiding} as our OOD detection technique, which is well suited for measuring complex AI systems under our setting.

\begin{algorithm}
\caption{OOD detection using Surprise Adequacy~\cite{kim2019guiding}}
\label{alg:ood}

\KwIn{An AI module $\mathbf{M}$, selected layer $L$, training data $X_{train}$, test data instance $\hat{x}$, metric $g(\cdot)$}
\KwOut{OOD score $p$}
$Z_{train} \gets \emptyset$\;
\For{$x_i~\leftarrow~X_{train}$}{
    $z_i \gets \mathbf{M}(x_i, L)$\;
    $Z_{train} = Z_{train}~\bigcup~\{z_i\}$\;
}
$p \gets g(Z_{train}, \hat{x})$\;
\KwRet{$p$}\;
\end{algorithm}

We formally describe the OOD detection workflow in Algorithm~\ref{alg:ood}. Specifically, we first select a target feature layer $L$ of a DL model in the system and collect all this layer’s output features $Z_{train}$ by feeding training data into a trained model (Line 2~-~4). Then, for each new test data instance, we measure the distance between its feature on $L$ and the distribution of collected training features $Z_{train}$ using a specific metric $g(\cdot)$ (Line 6). In this work, we use two representative metrics to measure such distribution distance, \ie, \textbf{Mahalanobis distance (Maha-d)} and \textbf{kernel distribution estimation (KDE)} adopted from existing work~\cite{lee2018simple,kim2019guiding}.

\subsubsection{Uncertainty analysis.} We make use of the Bayesian methods with the following two uncertainty metrics: Variation Ratio (VR) and Variation Ratio of Original prediction (VRO). These two metrics are originally proposed for classification problems~\cite{zhang2020towards}. Though metrics like standard deviation can be applied to simple regression task, however, most AI systems focus on more complex problems, e.g., object tracking (which could be a combination of classification and regression). Hence, to use generalizable analysis techniques, we make adaptions to the calculation of VR and VRO and define them as follows.

\begin{definition}
    Extended variation ratio (VR)
\end{definition}

\begin{equation}
    \label{eq:evr}
    VR = 1 - \frac{\sum_{i=1}^T w*\frac{\sum_{j=1, j \neq i}^{j=T} (1 - dist(p_i, p_j))}{T-1}}{T} 
\end{equation}

where $T$ is the inference time, $dist(\cdot)$ represents the distance function between two points. $p_i$ represents the inference result at $i$ inference time. $w$ is the weight. 

\begin{definition}
    Extended variation ratio for original prediction (VRO)
\end{definition}

\begin{equation}
\label{eq:evro}
    VRO = 1 - \frac{\sum_{i=1}^T (1 - dist(p_i, p_{L_M}))}{T}
\end{equation}

where $T$ is the inference time, $dist(\cdot)$ represents the distance function between two points. $p_i$ represents the inference result at $i$ inference time. $p_{L_M}$ is the prediction result from the original model $M$.


We argue that these adaptions are necessary for our study. Unlike classification problems, it's hard to find two identical outputs through multiple runs in other problems AI systems face, \eg, regression problems. Therefore we use a distance function to measure the similarity between two outputs instead of checking if the two outputs are numerically equal.

\subsection{Research Questions}
\label{subsec:research_questions}

\begin{figure}
    \centering
    \includegraphics[width=0.95\linewidth]{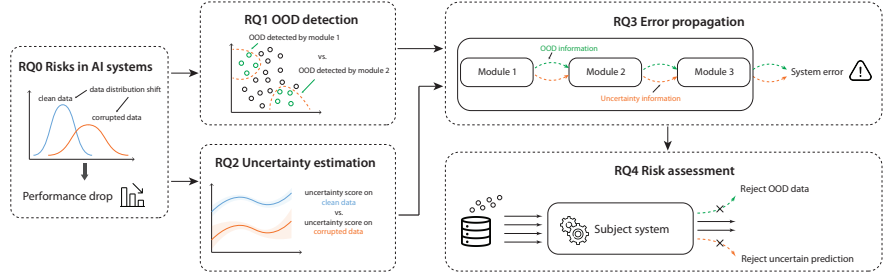}
    \caption{The workflow and relationship of research questions in the study design in this paper.}
    \label{fig:rq}
\end{figure}

We facilitate our study by investigating five different research questions . Fig.~\ref{fig:rq} shows the logic flow in our design of research questions. We first create large-scale corrupted datasets and investigate the potential risks brought by these common corruption patterns in RQ0, serving as the basis of this study. Then we discuss the effectiveness of two aspects' risk assessment techniques (\ie, OOD detection and uncertainty estimation) in RQ1 and RQ2 by leveraging the corrupted data generated in RQ0. After a deeper understanding of these techniques, we further investigate their potential as system error indicators in RQ3. Finally, we make an early attempt to combine these techniques for making risk assessments on AI systems. We detail the design of each research question in the following.

\begin{itemize}[leftmargin=*]
    \item \textbf{RQ0. How common corruption patterns increase risks in the wild for AI systems?} While common corruption patterns' effects on a single AI model have been widely studied, yet it's still unclear whether common corruption patterns could also affect AI systems' performance or not. In this research question, we first create 15 corrupted versions (5 corruption patterns~$\times$~3 severity levels) for each system's corresponding dataset. Then, we evaluate the performance of each AI system against different corruption patterns and severity levels with related performance metrics as introduced in Sec.~\ref{subsec:aisystems}. 
    
    \item \textbf{RQ1. What is the OOD awareness of different modules in AI systems?} Based on the corrupted dataset created in \textbf{RQ1}, we evaluate our OOD detection methods by comparing the OOD score on clean data and corrupted data created in \textbf{RQ1}. We first use the \textit{Wilcoxon signed-rank test} to measure if the OOD score's distribution on clean and corrupted data is significantly different. Then, we further measure if different AI modules are sensitive to different OOD data.
    
    \item \textbf{RQ2. How does the uncertainty information behave on different modules in AI systems?} Similar to \textbf{RQ2}, in \textbf{RQ3} we evaluate our uncertainty estimation methods by comparing the uncertainty score on clean data and corrupted data in this research question. Specifically, we use \textit{Wilcoxon signed-rank test} to measure if the uncertainty score's distribution on clean data and corrupted data are significantly different. Then, we use Kendall's $\tau$ test to measure if there is any correlation between different modules' uncertainty inside an AI system.
    
    \item \textbf{RQ3. How do the errors propagate among modules in AI systems?} In this research question, we investigate the potential propagation of erroneous information among modules in AI systems. Specifically, we use \textit{Kendall's $\tau$} test to measure if there is any correlation between different modules' OOD/uncertainty information and the system's output error.
    
    \item \textbf{RQ4. How do simple combinations of OOD analysis- and uncertainty-based method improve the reliability of AI systems?} In this RQ, we make an early attempt to combine OOD analysis and uncertainty information to make risk assessments of AI systems by rejecting unreliable predictions through five different strategies. Then, we evaluate systems' performance improvement after utilizing each strategy.
    
\end{itemize}

\subsection{Experimental Configurations}
\label{subsec:experimental_configurations}

In this section, we introduce the common configurations that are shared by multiple RQs in our study, \eg, model and dataset configurations, parameters of OOD detection and uncertainty estimation, and hardware and software dependencies. We will further detail the specific configurations (if any) of each individual research question in Sec.~\ref{sec:empirical}

\subsubsection{Model \& dataset configurations.}

We summarize the selection of each AI system's configurations and datasets for evaluation in Table~\ref{tab:model-dataset}. A complex AI system could have multiple configurations for each corresponding dataset, and each configuration could also have different settings for both AI models and traditional software. In this study, we choose all available configurations (with pre-trained models available) in the Udacity, ByteTrack, and SpeechBrain systems. While in the ConvLab-2 system, we select four configurations with the highest success rate on clean data.
Note that we create fifteen corrupted versions (5 corruption patterns $\times$ 3 severity levels) for each dataset. Therefore, the dataset size in Table~\ref{tab:model-dataset} refers to the total size of clean data and corrupted data.

\begin{table}[htbp]
    \centering
    \scriptsize
    \caption{Model~\&~corresponding dataset used for evaluating different AI systems}
    \begin{tabular}{llll}
    \toprule
    \textbf{System}              & \textbf{Configuration}                 & \textbf{Dataset}      & \textbf{Size of Dataset} \\ \midrule
    Udacity                      & CNN \& CNN \& CNN                      & Challenge2~\cite{udacitychallenge}  & 48,000 images      \\ \midrule
    \multirow{3}{*}{ByteTrack}   & YOLO+Track                         & \multirow{3}{*}{MOT17~\cite{milan2016mot16}}          & \multirow{3}{*}{3 hours video clips} \\
                                 & YOLO+SORT                              &                                 & \\
                                 & YOLO+DeepSORT                          &                                 & \\ \midrule
    \multirow{4}{*}{SpeechBrain} & \cellcolor{lightgray}CRDNN+RNN         & \cellcolor{lightgray} & \cellcolor{lightgray}      \\
                                 & \cellcolor{lightgray}CRDNN+Transformer & \multirow{-2}{*}{\cellcolor{lightgray}LibriSpeech~\cite{panayotov2015librispeech}}     & \multirow{-2}{*}{\cellcolor{lightgray}100 hours audio clips}\\
                                 & CRDNN+RNN                              & \multirow{2}{*}{CommonVoice-it~\cite{ardila2020common}} & \multirow{2}{*}{150 hours audio clips}\\
                                 & Wav2Vec2+RNN                           &                                 & \\ \midrule
    \multirow{4}{*}{ConvLab2}    & BERT+RULE                              & \multirow{4}{*}{MultiWOZ~\cite{zang2020multiwoz}}       & \multirow{4}{*}{16,000 dialogues} \\
                                 & MILU+RULE                              &                                 & \\
                                 & BERT+PPO                               &                                 & \\
                                 & MILU+PPO                               &                                 & \\ \bottomrule
    \end{tabular}
    \label{tab:model-dataset}
\end{table}

\begin{table}[htbp]
    \centering
    \scriptsize
    \caption{The choices of $dist(\cdot)$ function for different output formats in different systems.}
    \begin{tabular}{lll}
    \toprule
    \textbf{Output Format} & $dist(\cdot)$ & \textbf{Related System(s)} \\ \midrule
    Numerical value & {\em Standard deviation} & Udacity \\
    Numerical vector & {\em Cosine distance} & SpeechBrain \\
    Detection box & IOU & ByteTrack\\
    Tracklet & MOTA~\cite{bernardin2006multiple} & ByteTrack\\
    Text & BLEU~\cite{papineni2002bleu} & ConvLab-2 \& SpeechBrain\\
    \bottomrule
    \end{tabular}
    \label{tab:dist_function}
\end{table}

\subsubsection{OOD detection configurations~\&~uncertainty estimation.}

For \textbf{OOD detection}, each AI model could have multiple choices on the feature layer. In our experiments, we select the {\em penultimate} layer (the layer before the final output layer) as the target layer $L$, which is also adopted by existing work~\cite{liu2020energy}. 
For \textbf{uncertainty estimation}, we set the inference times of Monte Carlo dropout as 10 and the dropout rate as 0.1 throughout our experiments. Since different AI systems and modules might have different tasks and output formats, we adjust the $dist(\cdot)$ function (Eq.~\ref{eq:evr}~\&~\ref{eq:evro}) according to the output format of a system or an individual module. For example, the detector of a tracking system outputs bounding boxes, therefore we use IOU as the $dist(\cdot)$ function.
We summarize the $dist(\cdot)$ function used in our study in Table~\ref{tab:dist_function} and briefly describe {\em Cosine distance} in eq.~\ref{eq:cos}, IOU in eq.~\ref{eq:IOU}, MOTA in eq.~\ref{eq:mota}, BLEU in eq.~\ref{eq:blue_1} and eq.~\ref{eq:blue_2}. We also refer audience to Sec.~\ref{subsec:aisystems} for the details of MOTA.

{

    \begin{align}
        \label{eq:cos}
        Cos(\mathbf{x_1}, \mathbf{x_2}) &= \frac{x_1 \cdot x_2}{||x_1|| \times ||x_2||}
    \end{align}
    
    where $x_1$ and $x_2$ stand for the output (hidden representations in the format of vectors) of the neural networks with respect to two data points.

    \begin{align} 
        \label{eq:IOU}
        IOU &= \frac{|A \cap B|}{|A \cup B|} 
    \end{align}
    
    where $A$ and $B$ stand for the bounding box of two different predictions. $IOU$ computes the intersection over the union, which represents the overlapping degree of two predictions. This criterion is commonly used to evaluate object detection systems where a higher value means more overlapping between predicted results and ground truth. 
    
    \begin{align}
        \label{eq:blue_1}
        BLUE = BP \cdot exp(\sum_{n=1}^{N}w_n logp_n)
    \end{align}
    
    The BLUE score aims to compute how one sentence resembles the other, which usually serves as a criterion to evaluate the machine translation system. It is based on brevity penalty BP and n-gram precision, where $p_n$ is n-gram precision, and $w_n$ is positive weights summing to one. $w_n$ is usually set to uniform weights $w_n = 1 / N$. Specifically, BP is computed by:
    
    \begin{equation}
        \label{eq:blue_2}
        BP = \left\{\begin{array}{lr}
            1 &\quad \text{if} \mkern9mu c > r \\
            e ^ {(1 - r/c)} & \quad \text{if} \mkern9mu c \leq r \\
            \end{array}\right\} \\
    \end{equation}
    
    where $c$ is the length of the candidate translation and $r$ is the effective reference corpus length. $BP$ is used to ensure that the length of high-scoring translation matches the reference translations. 
}

\subsubsection{Hardware \& software dependencies.} Given that our experiments are highly computationally intensive, all the evaluations were run on a small cluster of four specialized Deep Learning servers. Each server has an AMD 3955WX CPU (3.9GHz), 256GB RAM, and four NVIDIA A4000 GPUs (16GB VRAM of each). Overall, our evaluation takes more than 5,000 GPU hours to complete. 

\section{Empirical Study Results}\label{sec:empirical}

In this section, we present the experiment and study results of each research question. In the paper, we only report our main results and summarized findings.

\subsection{RQ0. How common corruption patterns increase risks in the wild for AI systems?}

To answer RQ0, we investigate the common corruption patterns' effects on four AI systems by measuring the corresponding performance drop. We show the system's performance of Udacity, ByteTrack, ConvLab-2, and SpeechBrain against corrupted test data in Table~\ref{tab:udacity}, Table~\ref{tab:tracking}, Table~\ref{tab:convlab}, and Table~\ref{tab:speech}, respectively. In these tables, $\Delta$ indicates the relative performance decrease after applying common corruption patterns. A larger $\Delta$ indicates a bigger performance drop and suggests that the corresponding setting (\ie, specific corruption pattern and severity level) could threaten the system's robustness.

\begin{table}[htbp]
    \scriptsize
    \centering
    \caption{Udacity system's performance against different corruption patterns on Challenge2 dataset (each cell of results in the table is averaged over 3000 images' runs).}
    \begin{threeparttable}
    \begin{tabular}{lcccccccccccc}
    \toprule
    \multirow{2}{*}{\textbf{Corruption}} & \multirow{2}{*}{\textbf{Severity}} & \multicolumn{2}{c}{\textbf{Module 1}} && \multicolumn{2}{c}{\textbf{Module 2}} && \multicolumn{2}{c}{\textbf{Module 3}} && \multicolumn{2}{c}{\textbf{System}}\\
    \cmidrule{3-4} \cmidrule{6-7} \cmidrule{9-10} \cmidrule{12-13}
    & & \multicolumn{1}{c}{RMSE} & \multicolumn{1}{c}{$\Delta (\%)^\dagger$} && \multicolumn{1}{c}{RMSE} & \multicolumn{1}{c}{$\Delta (\%)^\dagger$} && \multicolumn{1}{c}{RMSE} & \multicolumn{1}{c}{$\Delta (\%)^\dagger$} && \multicolumn{1}{c}{RMSE} & \multicolumn{1}{c}{$\Delta (\%)^\dagger$} \\
    \midrule
    Original  & N/A      & 0.065 & \multicolumn{1}{c}{---} &  & 0.054 & \multicolumn{1}{c}{---} & & 0.048 & \multicolumn{1}{c}{---} & & 0.045 & \multicolumn{1}{c}{---} \\
    \cellcolor{lightgray} & \cellcolor{lightgray}1  & \cellcolor{lightgray}0.071 & \cellcolor{lightgray}9.2 &\cellcolor{lightgray}& \cellcolor{lightgray}0.054 & \cellcolor{lightgray}<0.1 &\cellcolor{lightgray}& \cellcolor{lightgray}0.049 & \cellcolor{lightgray}2.0 &\cellcolor{lightgray}& \cellcolor{lightgray}0.049 & \cellcolor{lightgray}8.9\\
    \cellcolor{lightgray} & \cellcolor{lightgray}2  & \cellcolor{lightgray}0.087 & \cellcolor{lightgray}33.8 &\cellcolor{lightgray}& \cellcolor{lightgray}0.089 & \cellcolor{lightgray}64.8 & \cellcolor{lightgray} & \cellcolor{lightgray}0.068 & \cellcolor{lightgray}41.7 & \cellcolor{lightgray} & \cellcolor{lightgray}0.072 & \cellcolor{lightgray}60.0\\
    \multirow{-3}{*}{\cellcolor{lightgray}Gaussian noise (GN)} & \cellcolor{lightgray}3 & \cellcolor{lightgray}0.164 & \cellcolor{lightgray}152.3 & \cellcolor{lightgray} & \cellcolor{lightgray}0.178 & \cellcolor{lightgray}229.6 & \cellcolor{lightgray} & \cellcolor{lightgray}0.144 & \cellcolor{lightgray}200.0 & \cellcolor{lightgray} & \cellcolor{lightgray}0.159 & \cellcolor{lightgray}253.3 \\
    \multirow{3}{*}{Motion blur (MB)}  & 1        & 0.068 & 4.6 & & 0.062 & 14.8 & & 0.055 & 14.6 && 0.051 & 13.3 \\
    & 2        & 0.081 & 24.6 & & 0.123 & 127.8 & & 0.092 & 91.6 & & 0.081 & 80.0 \\
    & 3        & 0.104 & 60.0 & & 0.233 & 331.5 & & 0.183 & 281.3 & & 0.143 & 217.8\\
    \cellcolor{lightgray} & \cellcolor{lightgray}1 & \cellcolor{lightgray}0.064 & \cellcolor{lightgray}-1 &\cellcolor{lightgray}& \cellcolor{lightgray}0.052 &\cellcolor{lightgray}3.7 & \cellcolor{lightgray} & \cellcolor{lightgray}0.048 & \cellcolor{lightgray}<0.1 & \cellcolor{lightgray} & \cellcolor{lightgray}0.044 & \cellcolor{lightgray}-2.2 \\
    \cellcolor{lightgray} & \cellcolor{lightgray}2 & \cellcolor{lightgray}0.070 & \cellcolor{lightgray}7.7 & \cellcolor{lightgray} & \cellcolor{lightgray}0.072 & \cellcolor{lightgray}33.3 & \cellcolor{lightgray} & \cellcolor{lightgray}0.056 &\cellcolor{lightgray}16.7 & \cellcolor{lightgray} & \cellcolor{lightgray}0.056 & \cellcolor{lightgray}24.4\\
    \multirow{-3}{*}{\cellcolor{lightgray}Snow (SN)} & \cellcolor{lightgray}3 & \cellcolor{lightgray}0.091 &  \cellcolor{lightgray}40.0 & \cellcolor{lightgray} & \cellcolor{lightgray}0.099 &\cellcolor{lightgray}83.3 & \cellcolor{lightgray} & \cellcolor{lightgray}0.088 & \cellcolor{lightgray}83.3 & \cellcolor{lightgray} & \cellcolor{lightgray}0.086 & \cellcolor{lightgray}91.1\\
    \multirow{3}{*}{Contrast (CO)}  & 1 & 0.077 & 18.5 & & 0.081 & 50.0 &  & 0.072 & 50.0 & & 0.064 & 42.2 \\
    & 2        & 0.112 & 72.3 & & 0.107 & 98.1 & & 0.113 & 135.4 &  & 0.093 & 106.7 \\
    & 3        & 0.167 & 156.9 & & 0.167 & 209.3 & & 0.182 & 279.2 &  & 0.170 & 277.8 \\
    \cellcolor{lightgray} & \cellcolor{lightgray}1 & \cellcolor{lightgray}0.068 & \cellcolor{lightgray}4.6 & \cellcolor{lightgray} & \cellcolor{lightgray}0.061 & \cellcolor{lightgray}13.0 & \cellcolor{lightgray} & \cellcolor{lightgray}0.056 & \cellcolor{lightgray}16.7 & \cellcolor{lightgray} & \cellcolor{lightgray}0.052 & \cellcolor{lightgray}15.6\\
    \cellcolor{lightgray} & \cellcolor{lightgray}2 & \cellcolor{lightgray}0.071 & \cellcolor{lightgray}9.2 & \cellcolor{lightgray} & \cellcolor{lightgray}0.071 & \cellcolor{lightgray}31.5 & \cellcolor{lightgray} & \cellcolor{lightgray}0.064 & \cellcolor{lightgray}33.3 & \cellcolor{lightgray} & \cellcolor{lightgray}0.058 & \cellcolor{lightgray}28.9\\
    \multirow{-3}{*}{\cellcolor{lightgray}Pixelate (PI)} & \cellcolor{lightgray}3 & \cellcolor{lightgray}0.075 & \cellcolor{lightgray}15.4 & \cellcolor{lightgray} & \cellcolor{lightgray}0.088 & \cellcolor{lightgray}63.0 & \cellcolor{lightgray} & \cellcolor{lightgray}0.074 & \cellcolor{lightgray}54.2 & \cellcolor{lightgray} & \cellcolor{lightgray}0.068 & \cellcolor{lightgray}51.1\\
    \bottomrule
    \end{tabular}
    \begin{tablenotes}
        \item[$\dagger$] $\Delta=\frac{\text{RMSE}}{\text{RMSE}_{\text{Original}}} - 1$; $\Delta$ suggests the performance decrease against different corruption patterns.
    \end{tablenotes}
    \end{threeparttable}
    \label{tab:udacity}
\end{table}

\begin{table}[htbp]
    \scriptsize
    \centering
    \caption{ByteTrack's performance against different corruption patterns on MOT17 dataset (each cell of results
in the table is averaged over 21,600 frames of videos' runs).}
    \begin{threeparttable}
        \centering
        \begin{tabular}{lccccccccc}
        \toprule
        \multirow{2}{*}{\textbf{Corruption}} & \multirow{2}{*}{\textbf{Severity}} & \multicolumn{2}{c}{\textbf{Configuration $1^\star$}} && \multicolumn{2}{c}{\textbf{Configuration $2^\star$}} && \multicolumn{2}{c}{\textbf{Configuration $3^\star$}}\\
        \cmidrule{3-4} \cmidrule{6-7} \cmidrule{9-10}
        & & \multicolumn{1}{c}{MOTA} & $\Delta (\%)^\dagger$ & & \multicolumn{1}{c}{MOTA} & $\Delta (\%)^\dagger$ & & \multicolumn{1}{c}{MOTA} & $\Delta (\%)^\dagger$ \\
        \midrule
        Original  & N/A      & 0.765 & \multicolumn{1}{c}{---} && 0.746 & \multicolumn{1}{c}{---} && 0.754 & \multicolumn{1}{c}{---}  \\
        \cellcolor{lightgray} & \cellcolor{lightgray}1  & \cellcolor{lightgray}0.728 & \cellcolor{lightgray}4.8 & \cellcolor{lightgray} & \cellcolor{lightgray}0.704 & \cellcolor{lightgray}5.6 & \cellcolor{lightgray} & \cellcolor{lightgray}0.725 & \cellcolor{lightgray}3.8 \\
        \cellcolor{lightgray} & \cellcolor{lightgray}2  & \cellcolor{lightgray}0.662 & \cellcolor{lightgray}13.5 & \cellcolor{lightgray} & \cellcolor{lightgray}0.622 & \cellcolor{lightgray}16.6 & \cellcolor{lightgray} & \cellcolor{lightgray}0.651 & \cellcolor{lightgray}13.6\\
        \multirow{-3}{*}{\cellcolor{lightgray}Gaussian noise (GN)} & \cellcolor{lightgray}3 & \cellcolor{lightgray}0.362 & \cellcolor{lightgray}52.7 & \cellcolor{lightgray} & \cellcolor{lightgray}0.302 & \cellcolor{lightgray}59.5 & \cellcolor{lightgray} & \cellcolor{lightgray}0.355 & \cellcolor{lightgray}52.9 \\
        \multirow{3}{*}{Motion blur (MB)}  & 1        & 0.745 & 2.6 & & 0.722 & 3.2 & & 0.736 & 2.4 \\
        & 2        & 0.645 & 5.1 & & 0.619 & 17.0 & & 0.636 & 15.6\\
        & 3        & 0.441 & 42.4  & & 0.413 & 44.6 & & 0.441 & 41.5 \\
        \cellcolor{lightgray} & \cellcolor{lightgray}1 & \cellcolor{lightgray}0.750 & \cellcolor{lightgray}2.0 & \cellcolor{lightgray} & \cellcolor{lightgray}0.736 & \cellcolor{lightgray}1.3 & \cellcolor{lightgray} & \cellcolor{lightgray}0.746 & \cellcolor{lightgray}1.1\\
        \cellcolor{lightgray} & \cellcolor{lightgray}2 & \cellcolor{lightgray}0.726 & \cellcolor{lightgray}5.1 & \cellcolor{lightgray} & \cellcolor{lightgray}0.704 & \cellcolor{lightgray}5.6 & \cellcolor{lightgray} & \cellcolor{lightgray}0.717 & \cellcolor{lightgray}4.9\\
        \multirow{-3}{*}{\cellcolor{lightgray}Contrast (CO)} & \cellcolor{lightgray}3 & \cellcolor{lightgray}0.283 & \cellcolor{lightgray}63.0 & \cellcolor{lightgray} & \cellcolor{lightgray}0.240 & \cellcolor{lightgray}67.8 & \cellcolor{lightgray} & \cellcolor{lightgray}0.259 & \cellcolor{lightgray}65.6\\
        \multirow{3}{*}{Snow (SN)}  & 1 & 0.725 & 5.2 && 0.686 & 8.0 && 0.722 & 4.2 \\
        & 2        & 0.682 & 10.8 && 0.604 & 19.0 && 0.670 & 11.1 \\
        & 3        & 0.613 & 19.9 && 0.538 & 27.9 && 0.606 & 19.6\\
        \cellcolor{lightgray} & \cellcolor{lightgray}1 & \cellcolor{lightgray}0.757 & \cellcolor{lightgray}1.0 & \cellcolor{lightgray} & \cellcolor{lightgray}0.735 & \cellcolor{lightgray}1.5 & \cellcolor{lightgray} & \cellcolor{lightgray}0.747 & \cellcolor{lightgray}0.9\\
        \cellcolor{lightgray} & \cellcolor{lightgray}2 & \cellcolor{lightgray}0.721 & \cellcolor{lightgray}5.8 & \cellcolor{lightgray} & \cellcolor{lightgray}0.692 & \cellcolor{lightgray}7.2 & \cellcolor{lightgray} & \cellcolor{lightgray}0.710 & \cellcolor{lightgray}5.8\\
        \multirow{-3}{*}{\cellcolor{lightgray}Pixelate (PI)} & \cellcolor{lightgray}3 & \cellcolor{lightgray}0.624 & \cellcolor{lightgray}18.4 & \cellcolor{lightgray} & \cellcolor{lightgray}0.602 & \cellcolor{lightgray}19.3 & \cellcolor{lightgray} & \cellcolor{lightgray}0.618 & \cellcolor{lightgray}18.0\\
        \bottomrule
        \end{tabular}
    \begin{tablenotes}
        \item[$\dagger$] $\Delta= 1 - \frac{\text{MOTA}}{\text{MOTA}_{\text{Original}}}$; $\Delta$ suggests the performance decrease against different corruption patterns.
        \item[$\star$] Configuration 1: YOLO + Track; Configuration 2: YOLO + SORT; Configuration 3: YOLO + DeepSORT.
    \end{tablenotes}
    \end{threeparttable}
    \label{tab:tracking}
\end{table}

\begin{table}[htbp]
    \scriptsize
    \centering
    \caption{ConvLab-2's performance against different corruption patterns on MultiWOZ dataset (each cell of
results in the table is averaged over 1000 dialogues' runs).}
    \begin{threeparttable}
        \centering
        \resizebox{\linewidth}{!}{
        \begin{tabular}{lcccccccccccc}
        \toprule
        \multirow{2}{*}{\textbf{Corruption}} & \multirow{2}{*}{\textbf{Severity}} & \multicolumn{2}{c}{Configuration $1^\star$} && \multicolumn{2}{c}{Configuration $2^\star$} && \multicolumn{2}{c}{Configuration $3^\star$} && \multicolumn{2}{c}{Configuration $4^\star$}\\
        \cmidrule{3-4} \cmidrule{6-7} \cmidrule{9-10} \cmidrule{12-13}
        & & SR (\%) & $\Delta (\%)^\dagger$ && SR (\%) & $\Delta (\%)^\dagger$ && SR (\%) & $\Delta (\%)^\dagger$ && SR (\%) & $\Delta (\%)^\dagger$\\ 
        \midrule
        Original  & N/A      & 70.5 & --- && 63.2 & --- && 49.7 & --- && 44.4 & ---\\
        \cellcolor{lightgray} & \cellcolor{lightgray}1  & \cellcolor{lightgray}63.8 & \cellcolor{lightgray}9.5 & \cellcolor{lightgray} & \cellcolor{lightgray}56.5 & \cellcolor{lightgray}10.6 & \cellcolor{lightgray} & \cellcolor{lightgray}43.9 & \cellcolor{lightgray}11.7 & \cellcolor{lightgray} & \cellcolor{lightgray}36.6 & \cellcolor{lightgray}17.6\\
        \cellcolor{lightgray} & \cellcolor{lightgray}2  & \cellcolor{lightgray}60.3 & \cellcolor{lightgray}14.5 & \cellcolor{lightgray} & \cellcolor{lightgray}53.3 & \cellcolor{lightgray}15.7 & \cellcolor{lightgray} & \cellcolor{lightgray}43.0 & \cellcolor{lightgray}13.5 & \cellcolor{lightgray} & \cellcolor{lightgray}36.4 & \cellcolor{lightgray}18.0\\
        \multirow{-3}{*}{\cellcolor{lightgray}Digits2words (DW)} & \cellcolor{lightgray}3 & \cellcolor{lightgray}55.6 & \cellcolor{lightgray}21.1 & \cellcolor{lightgray} & \cellcolor{lightgray}50.2 & \cellcolor{lightgray}20.6 & \cellcolor{lightgray} & \cellcolor{lightgray}41.9 & \cellcolor{lightgray}15.7 & \cellcolor{lightgray} & \cellcolor{lightgray}35.9 & \cellcolor{lightgray}19.1  \\
        \multirow{3}{*}{Change char (CC)}  & 1        & 57.7 & 18.2 & & 51.1 & 19.1 & & 31.1 & 37.4 & & 33.2 & 25.2 \\
        & 2        & 45.0 & 36.2 & & 33.0 & 47.8 & & 30.7 & 38.2 & & 23.8 & 46.4\\
        & 3        & 29.7 & 57.9 & & 19.1 & 69.8 & & 16.9 & 66.0 & & 11.2 & 74.8 \\
        \cellcolor{lightgray} & \cellcolor{lightgray}1 & \cellcolor{lightgray}52.5 & \cellcolor{lightgray}25.5 & \cellcolor{lightgray} & \cellcolor{lightgray}47.8 & \cellcolor{lightgray}24.4 &\cellcolor{lightgray} & \cellcolor{lightgray}40.4 & \cellcolor{lightgray}18.7 & \cellcolor{lightgray} & \cellcolor{lightgray}35.9 & \cellcolor{lightgray}19.1\\
        \cellcolor{lightgray} & \cellcolor{lightgray}2 & \cellcolor{lightgray}31.5 & \cellcolor{lightgray}55.3 & \cellcolor{lightgray} & \cellcolor{lightgray}30.3 & \cellcolor{lightgray}52.1 & \cellcolor{lightgray} & \cellcolor{lightgray}22.6 & \cellcolor{lightgray}54.5 & \cellcolor{lightgray} & \cellcolor{lightgray}20.2 & \cellcolor{lightgray}54.5\\
        \multirow{-3}{*}{\cellcolor{lightgray}Remove char (RC)} & \cellcolor{lightgray}3 & \cellcolor{lightgray}13.7 & \cellcolor{lightgray}80.6 & \cellcolor{lightgray} & \cellcolor{lightgray}13.2 & \cellcolor{lightgray}79.1 & \cellcolor{lightgray} & \cellcolor{lightgray}5.5 & \cellcolor{lightgray}88.9 & \cellcolor{lightgray} & \cellcolor{lightgray}4.7 & \cellcolor{lightgray}89.4\\
        \multirow{3}{*}{Misspelling (MS)}  & 1 & 63.3 & 10.2 & & 54.5 & 13.8 & & 44.0 & 11.5 & & 36.4 & 18.0\\
        & 2        & 56.2 & 20.3 & & 44.1 & 30.2 & & 41.0 & 17.5 & & 34.7 & 21.8\\
        & 3        & 29.7 & 57.9 & & 21.4 & 66.1 & & 31.2 & 37.2 & & 25.5 & 42.6\\
        \cellcolor{lightgray} & \cellcolor{lightgray}1 & \cellcolor{lightgray}47.5 & \cellcolor{lightgray}32.6 & \cellcolor{lightgray} & \cellcolor{lightgray}44.3 & \cellcolor{lightgray}29.9 & \cellcolor{lightgray} & \cellcolor{lightgray}36.0 & \cellcolor{lightgray}27.6 & \cellcolor{lightgray} & \cellcolor{lightgray}33.3 & \cellcolor{lightgray}25.0\\
        \cellcolor{lightgray} & \cellcolor{lightgray}2 & \cellcolor{lightgray}20.6 & \cellcolor{lightgray}70.8 & \cellcolor{lightgray} & \cellcolor{lightgray}14.8 & \cellcolor{lightgray}76.6 & \cellcolor{lightgray} & \cellcolor{lightgray}8.6 & \cellcolor{lightgray}82.7 & \cellcolor{lightgray} & \cellcolor{lightgray}4.8 & \cellcolor{lightgray}89.2\\
        \multirow{-3}{*}{\cellcolor{lightgray}Swap (SW)} & \cellcolor{lightgray}3 & \cellcolor{lightgray}3.2 & \cellcolor{lightgray}95.5 & \cellcolor{lightgray} & \cellcolor{lightgray}1.4 & \cellcolor{lightgray}97.8 & \cellcolor{lightgray} & \cellcolor{lightgray}3.0 & \cellcolor{lightgray}94.0 & \cellcolor{lightgray} & \cellcolor{lightgray}2.1 & \cellcolor{lightgray}95.3\\
        \bottomrule
        \end{tabular}
        }
    \begin{tablenotes}
        \item[$\dagger$] $\Delta= 1 - \frac{\text{SR}}{\text{SR}_{\text{Original}}}$, SR denotes {\em success rate}; $\Delta$ suggests the performance decrease against different corruption patterns.
        \item[$\star$] Configuration 1: BERT NLU + Rule POL; Configuration 2: MILU NLU + Rule POL; Configuration 3: BERT NLU + PPO POL; Configuration 4: MILU NLU + PPO POL.
    \end{tablenotes}
    \end{threeparttable}
    \label{tab:convlab}
\end{table}

\begin{table}[htbp]
    \scriptsize
    \centering
    \caption{SpeechBrain’s performance against different corruption patterns on LibriSpeech (each cell of results
in the table is averaged over 2600 audio samples' runs)~\&~CommonVoice-it (each cell of results in the table is averaged over 8940 audio samples' runs) dataset.}
    \begin{threeparttable}
        \centering
        \resizebox{\linewidth}{!}{
        \begin{tabular}{lcccccccccccc}
        \toprule
        \multirow{2}{*}{\textbf{Corruption}} & \multirow{2}{*}{\textbf{Severity}} & \multicolumn{2}{c}{Configuration $1^\star$} && \multicolumn{2}{c}{Configuration $2^\star$} && \multicolumn{2}{c}{Configuration $3^\star$} && \multicolumn{2}{c}{Configuration $4^\star$}\\
        \cmidrule{3-4} \cmidrule{6-7} \cmidrule{9-10} \cmidrule{12-13}
        & & WER & $\Delta (\%)^\dagger$ && WER & $\Delta (\%)^\dagger$ && WER & $\Delta (\%)^\dagger$ && WER & $\Delta (\%)^\dagger$\\ 
        \midrule
        Original  & N/A      & 0.212 & --- && 0.035 & --- && 0.249 & --- && 0.299 & ---\\
        \cellcolor{lightgray} & \cellcolor{lightgray}1  & \cellcolor{lightgray}0.185 & \cellcolor{lightgray}-12.7 & \cellcolor{lightgray} & \cellcolor{lightgray}0.100 & \cellcolor{lightgray}185.7 & \cellcolor{lightgray} & \cellcolor{lightgray}0.379 & \cellcolor{lightgray}26.8 & \cellcolor{lightgray} & \cellcolor{lightgray}0.295 & \cellcolor{lightgray}18.5\\
        \cellcolor{lightgray} & \cellcolor{lightgray}2  & \cellcolor{lightgray}0.334 & \cellcolor{lightgray}57.5 & \cellcolor{lightgray} & \cellcolor{lightgray}0.560 & \cellcolor{lightgray}1500.0 & \cellcolor{lightgray} & \cellcolor{lightgray}0.645 & \cellcolor{lightgray}115.7 & \cellcolor{lightgray} & \cellcolor{lightgray}0.482 & \cellcolor{lightgray}93.6\\
        \multirow{-3}{*}{\cellcolor{lightgray}Reverb (RE)} & \cellcolor{lightgray}3 & \cellcolor{lightgray}0.766 & \cellcolor{lightgray}261.3 & \cellcolor{lightgray} & \cellcolor{lightgray}0.852 & \cellcolor{lightgray}2334.3 & \cellcolor{lightgray} & \cellcolor{lightgray}0.883 & \cellcolor{lightgray}195.3 & \cellcolor{lightgray} & \cellcolor{lightgray}0.749 & \cellcolor{lightgray}200.8\\
        \multirow{3}{*}{Speed (SP)}  & 1        & 0.214 & 0.9 & & 0.047 & 34.3 & & 0.455 & 52.2 & & 0.288 & 15.7\\
        & 2        & 0.608 & 196.8 & & 0.339 & 868.6 & & 0.676 & 126.1 & & 0.360 & 44.6 \\
        & 3        & 1.159 & 446.7 & & 0.981 & 2702.9 & & 1.013 & 238.8 & & 0.991 & 298.0 \\
        \cellcolor{lightgray} & \cellcolor{lightgray}1 & \cellcolor{lightgray}0.195 & \cellcolor{lightgray}-8.0 & \cellcolor{lightgray} & \cellcolor{lightgray}0.041 & \cellcolor{lightgray}17.1 & \cellcolor{lightgray} & \cellcolor{lightgray}0.344 & \cellcolor{lightgray}15.1 & \cellcolor{lightgray} & \cellcolor{lightgray}0.270 & \cellcolor{lightgray}8.4\\
        \cellcolor{lightgray} & \cellcolor{lightgray}2 & \cellcolor{lightgray}0.247 & \cellcolor{lightgray}16.5 & \cellcolor{lightgray} & \cellcolor{lightgray}0.052 & \cellcolor{lightgray}48.6 & \cellcolor{lightgray} & \cellcolor{lightgray}0.390 & \cellcolor{lightgray}30.4 & \cellcolor{lightgray} & \cellcolor{lightgray}0.308 & \cellcolor{lightgray}23.7\\
        \multirow{-3}{*}{\cellcolor{lightgray}Distortion (DI)} & \cellcolor{lightgray}3 & \cellcolor{lightgray}0.374 & \cellcolor{lightgray}76.4 & \cellcolor{lightgray} & \cellcolor{lightgray}0.216 & \cellcolor{lightgray}517.1 & \cellcolor{lightgray} & \cellcolor{lightgray}0.553 & \cellcolor{lightgray}84.9 & \cellcolor{lightgray} & \cellcolor{lightgray}0.473 & \cellcolor{lightgray}90.0\\
        \multirow{3}{*}{Pitch Shift (PS)}  & 1 & 0.114 & -46.2 & & 0.043 & 22.9 & & 0.336 & 12.4 & & 0.264 & 6.0\\
        & 2        & 0.170 & -19.8 & & 0.065 & 85.7 & & 0.385 & 28.8 & & 0.291 & 16.9 \\
        & 3        & 0.245 & 15.6 & & 0.174 & 397.1 & & 0.496 & 65.9 & & 0.347 & 39.4\\
        \cellcolor{lightgray} & \cellcolor{lightgray}1 & \cellcolor{lightgray}0.164 & \cellcolor{lightgray}-22.6 & \cellcolor{lightgray} & \cellcolor{lightgray}0.330 & \cellcolor{lightgray}842.9 & \cellcolor{lightgray} & \cellcolor{lightgray}0.387 & \cellcolor{lightgray}29.4 & \cellcolor{lightgray} & \cellcolor{lightgray}0.291 & \cellcolor{lightgray}16.9 \\
        \cellcolor{lightgray} & \cellcolor{lightgray}2 & \cellcolor{lightgray}0.194 & \cellcolor{lightgray}-8.5 & \cellcolor{lightgray} & \cellcolor{lightgray}0.720 & \cellcolor{lightgray}1957.1 & \cellcolor{lightgray} & \cellcolor{lightgray}0.479 & \cellcolor{lightgray}60.2 & \cellcolor{lightgray} & \cellcolor{lightgray}0.350 & \cellcolor{lightgray}40.6\\
        \multirow{-3}{*}{\cellcolor{lightgray}Background noise (BN)} & \cellcolor{lightgray}3 & \cellcolor{lightgray}0.243 & \cellcolor{lightgray}14.6 &\cellcolor{lightgray} & \cellcolor{lightgray}0.896 & \cellcolor{lightgray}2460.0 & \cellcolor{lightgray} & \cellcolor{lightgray}0.568 & \cellcolor{lightgray}90.0 & \cellcolor{lightgray} & \cellcolor{lightgray}0.416 & \cellcolor{lightgray}67.1 \\
        \bottomrule
        \end{tabular}
        }
    \begin{tablenotes}
        \item[$\dagger$] $\Delta= \frac{\text{WER}}{\text{WER}_{\text{Original}}} - 1$, WER denotes {\em word error rate}; $\Delta$ suggests the performance decrease against different corruption patterns.
        \item[$\star$] Configuration 1: CRDNN + RNN (LibriSpeech); Configuration 2: CRDNN + Transformer (LibriSpeech); Configuration 3: CRDNN + RNN (CommonVoice-it); Configuration 4: Wav2Vec2 + RNN (CommonVoice-it).
    \end{tablenotes}
    \end{threeparttable}
    \label{tab:speech}
\end{table}

Overall, we confirm that AI systems' robustness and reliability can be threatened by the common corruption patterns. { In the first image-oriented system, \ie, Udacity system, we observe that the highest severity level of corruption could drop the system's performance by 178.2\% on average across different configurations, respectively. Specifically, we find that the system performs worse when against the highest severity level of Gaussian noise (GN), motion blur (MB) and contrast change (CO). Meanwhile, we also find that in some cases, even the lowest severity level of corruption could affect systems performance significantly (\eg, level 1 of \textbf{CO} could decrease the Udacity system's performance by 42.2\%).
In the ByteTrack system, we find that with the highest level of severity, the system performance is dropped by $59.5\%$ across five different corruption patterns and three system configurations.} The performance of the dialogue system, \ie, ConvLab-2, is also significantly affected by common text corruption patterns. On average, even the lowest severity of corruption can drop the systems' performance by 20.1\% across different configurations and corruption patterns, respectively. Specifically, we also find that the \textbf{SW} corruption poses the severest threat to the ConvLab-2 system's robustness. On average, the system's performance dropped by 22.0\%, 79.8\%, and 95.7\% (from lowest severity level to highest one, respectively) across four configurations when facing \textbf{SW}. Audio corruption patterns also affect the performance of the speech recognition system, \ie, SpeechBrain. On average, when facing the lowest and highest severity level of corruption, the system's performance drop is by 60.8\% and 529.8\% across different system configurations and corruption patterns, respectively. Surprisingly, in the SpeechBrain system, we find that the performance on clean data can't indicate the system's robustness. For instance, on the LibriSpeech dataset, configuration 1 (CRDNN+RNN) has a much higher WER than configuration 2 (CRDNN+Transformer) on the clean data (0.212 vs. 0.035). However, when facing different corruption patterns, there is a significant performance drop with configuration 2. For instance, the WER with configuration 2 increases by 1500.0\% when facing the level-2 \textbf{RE} effect. By contrast, configuration 1 only drops by 57.5\% when facing the same corruption. In fact, different from the system with configuration 2, the performance of the system with configuration 1 does not even drop when being against 4 out of 5 level-1 corruption patterns.

We also find some possibilities to improve the AI system's reliability based on the different performances of each AI system with different configurations against common corruption patterns. Recall that the Udacity system's output is determined by three independently executed modules (by taking the average). We find that different modules can be robust to different corruption patterns. For instance, when increasing the severity level of \textbf{MB} corruption, the performance (\ie, RMSE) of Module 2 and Module 3 are significantly affected (dropped by 331.5\% and 281.3\%, respectively). However, the performance drop of Module 1 is only 60.0\%. Therefore, we can further improve the system's performance if we can reject the bad prediction results from Modules 2 and 3 instead of always taking the average of all three modules. This also motivates us to investigate whether there is a good indicator to detect such bad prediction results. Similarly, in the SpeechBrain system, we observe that \textbf{BN} corruption does not have many effects on configuration 1 (even level 3 corruption only increases WER from 0.212 to 0.243). However, the effects of \textbf{BN} are significant in configuration 2, where WER increased from 0.035 to 0.896 when facing level 3 corruption. Hence, given that different AI system configurations can be robust against different corruption patterns, a comprehensive understanding of each system is needed in order to achieve better system robustness.


\begin{tcolorbox}[size=title]
{
\textbf{Answer to RQ0:} In general, we confirm that common corruption patterns also significantly affect the AI system's performance, posing threats to the AI system's prediction. A higher severity level of corruption usually has more substantial effects on the AI system's performance. In addition, we also find AI systems with different configurations can be resilient against different corruption patterns, revealing possibilities of improving system reliability by the ensemble.
}
\end{tcolorbox}

\subsection{RQ1. What is the OOD awareness of different modules in AI systems?}

In this research question, we utilize the corrupted datasets created in RQ1 to further evaluate the usefulness of two OOD detection techniques in this study---Mahalanobis distance (Maha-d) and kernel distribution estimation (KDE). We first evaluate the OOD score's distribution differences between clean data and corrupted data in Sec.~\ref{sec:ood_distribution}, then we investigate if different modules inside an AI system could have different OOD awareness in Sec.~\ref{sec:ood_diff}.

\subsubsection{OOD detection against common corruption patterns.}
\label{sec:ood_distribution}

\begin{table}[htbp]
    \scriptsize
    \centering
    \caption{Wilcoxon signed-rank test on ByteTrack system's OOD score distribution difference between clean data and corrupted data on MOT17 dataset.}
    \begin{threeparttable}
    \resizebox{\linewidth}{!}{
    \begin{tabular}{llccccc}
    \toprule
    \multirow{2}{*}{\textbf{Corruption}} & \multirow{2}{*}{\textbf{Settings$^\diamond$}} & \multicolumn{2}{c}{${\mathcal{D}_\textbf{Maha-d}}^\dagger$} && \multicolumn{2}{c}{${\mathcal{D}_\textbf{KDE}}^\dagger$}\\
    \cmidrule{3-4} \cmidrule{6-7}
    & & \multicolumn{1}{c}{YOLO detector} & \multicolumn{1}{c}{DeepSORT tracker} && \multicolumn{1}{c}{YOLO detector} & \multicolumn{1}{c}{DeepSORT tracker}\\
    \midrule
    \cellcolor{lightgray} & \cellcolor{lightgray}L1-Ori & \cellcolor{lightgray}$0.85$ & \cellcolor{lightgray}$0.73$ &\cellcolor{lightgray}& \cellcolor{lightgray}$1.00$ & \cellcolor{lightgray}$0.06*$\\
    \cellcolor{lightgray} & \cellcolor{lightgray}L2-L1 & \cellcolor{lightgray}$0.67$ & \cellcolor{lightgray}$0.84$ &\cellcolor{lightgray}& \cellcolor{lightgray}$0.74$ & \cellcolor{lightgray}$0.08*$\\
    \multirow{-3}{*}{\cellcolor{lightgray}Gaussian noise (GN)} & \cellcolor{lightgray}L3-L2 & \cellcolor{lightgray}$0.96$ & \cellcolor{lightgray}$0.62$ &\cellcolor{lightgray}& \cellcolor{lightgray}$0.98$ & \cellcolor{lightgray}$0.19*$\\
    \multirow{3}{*}{Motion blur (MB)} & L1-Ori &$0.30*$ & $0.67$ && $1.00$ & $0.49*$\\
    & L2-L1 & $0.93$ & $0.72$ && $0.96$ & $0.29*$\\
    & L3-L2 & $0.90$ & $0.69$ && $0.89$ & $0.27*$\\
    \cellcolor{lightgray} & \cellcolor{lightgray}L1-Ori & \cellcolor{lightgray}$0.35*$ & \cellcolor{lightgray}$0.78$ &\cellcolor{lightgray}& \cellcolor{lightgray}$1.00$ & \cellcolor{lightgray}$0.17*$\\
    \cellcolor{lightgray} & \cellcolor{lightgray}L2-L1 & \cellcolor{lightgray}$0.96$ & \cellcolor{lightgray}$0.84$ &\cellcolor{lightgray}& \cellcolor{lightgray}$0.92$ & \cellcolor{lightgray}$0.10*$\\
    \multirow{-3}{*}{\cellcolor{lightgray}Contrast (CO)} & \cellcolor{lightgray}L3-L2 & \cellcolor{lightgray}$1.00$ & \cellcolor{lightgray}$0.82$ &\cellcolor{lightgray}& \cellcolor{lightgray}$0.97$ & \cellcolor{lightgray}$0.09*$\\
    \multirow{3}{*}{Snow (SN)} & L1-Ori & $0.69$ & $0.93$ && $1.00$ & $0.22*$\\
    & L2-L1 & $0.56$ & $0.88$ && $0.60$ & $0.21*$\\
    & L3-L2 & $0.72$ & $0.70$ && $0.68$ & $0.23*$\\
    \cellcolor{lightgray} & \cellcolor{lightgray}L1-Ori & \cellcolor{lightgray}$0.30*$ & \cellcolor{lightgray}$0.45*$ &\cellcolor{lightgray}& \cellcolor{lightgray}$1.00$ & \cellcolor{lightgray}$0.31*$\\
    \cellcolor{lightgray} & \cellcolor{lightgray}L2-L1 & \cellcolor{lightgray}$0.91$ & \cellcolor{lightgray}$0.46*$ &\cellcolor{lightgray}& \cellcolor{lightgray}$0.93$ & \cellcolor{lightgray}$0.31*$\\
    \multirow{-3}{*}{\cellcolor{lightgray}Pixelate (PI)} & \cellcolor{lightgray}L3-L2 & \cellcolor{lightgray}$0.97$ & \cellcolor{lightgray}$0.40*$ &\cellcolor{lightgray}& \cellcolor{lightgray}$0.97$ & \cellcolor{lightgray}$0.18*$\\
    \bottomrule
    \end{tabular}
    }
    \begin{tablenotes}
        \item[$\diamond$] A setting of ``A-B'' indicates that the test is conducted between uncertainty score on A and B, \eg, ``L1-Ori'' means the test is conducted between level 1 corrupted data and clean data.
        \item[$\dagger$] An effect size $\mathcal{D}$ indicates the probability that uncertainty score on corrupted data is larger than it on clean data; larger the effect size, more significant the result is.
        \item[$^*$] This result is not statistically significant ($p>0.0001$).
    \end{tablenotes}
    \end{threeparttable}
    \label{tab:tracking_ood}
\end{table}

\begin{table}[htbp]
    \scriptsize
    \centering
    \caption{Wilcoxon signed-rank test on ConvLab-2 system's OOD score distribution difference between clean data and corrupted data on MultiWOZ dataset.}
    \begin{threeparttable}
    \resizebox{\linewidth}{!}{
    \begin{tabular}{llccccccc}
    \toprule
    \multirow{2}{*}{\textbf{Corruption}} & \multirow{2}{*}{\textbf{Settings$^\diamond$}} & \multicolumn{3}{c}{${\mathcal{D}_\textbf{Maha-d}}^\dagger$} && \multicolumn{3}{c}{${\mathcal{D}_\textbf{KDE}}^\dagger$}\\
    \cmidrule{3-5} \cmidrule{7-9}
    & & \multicolumn{1}{c}{BERT NLU} & \multicolumn{1}{c}{MILU NLU} & \multicolumn{1}{c}{PPO Pol} && \multicolumn{1}{c}{BERT NLU} & \multicolumn{1}{c}{MILU NLU} & \multicolumn{1}{c}{PPO Pol}\\
    \midrule
    \cellcolor{lightgray} & \cellcolor{lightgray}L1-Ori & \cellcolor{lightgray}$0.55*$ & \cellcolor{lightgray}$0.75$ & \cellcolor{lightgray}$0.42*$ &\cellcolor{lightgray}& \cellcolor{lightgray}$0.60$ & \cellcolor{lightgray}$0.72$ & \cellcolor{lightgray}$0.80$\\
    \cellcolor{lightgray} & \cellcolor{lightgray}L2-L1 & \cellcolor{lightgray}$0.53*$ & \cellcolor{lightgray}$0.55*$ & \cellcolor{lightgray}$0.44*$ &\cellcolor{lightgray}& \cellcolor{lightgray}$0.57$ & \cellcolor{lightgray}$0.54*$ & \cellcolor{lightgray}$0.49*$\\
    \multirow{-3}{*}{\cellcolor{lightgray}Digits2words (DW)} & \cellcolor{lightgray}L3-L2 & \cellcolor{lightgray}$0.48*$ & \cellcolor{lightgray}$0.49*$ & \cellcolor{lightgray}$0.44*$ &\cellcolor{lightgray}& \cellcolor{lightgray}$0.55$ & \cellcolor{lightgray}$0.54*$ & \cellcolor{lightgray}$0.42*$\\
    \multirow{3}{*}{Change char (CC)} & L1-Ori &$0.96$ & $0.96$ & $0.48*$ && $0.98$ & $0.84$ & $0.88$\\
    & L2-L1 & $0.81$ & $0.97$ & $0.43*$ && $0.81$ & $0.87$ & $0.75$\\
    & L3-L2 & $0.23*$ & $0.27*$ & $0.33*$ && $0.78$ & $0.90$ & $0.45*$\\
    \cellcolor{lightgray} & \cellcolor{lightgray}L1-Ori & \cellcolor{lightgray}$0.98$ & \cellcolor{lightgray}$0.96$ & \cellcolor{lightgray}$0.46*$ &\cellcolor{lightgray}& \cellcolor{lightgray}$0.98$ & \cellcolor{lightgray}$0.86$ & \cellcolor{lightgray}$0.89$\\
    \cellcolor{lightgray} & \cellcolor{lightgray}L2-L1 & \cellcolor{lightgray}$0.80$ & \cellcolor{lightgray}$0.91$ & \cellcolor{lightgray}$0.32*$ &\cellcolor{lightgray}& \cellcolor{lightgray}$0.80$ & \cellcolor{lightgray}$0.84$ & \cellcolor{lightgray}$0.73$\\
    \multirow{-3}{*}{\cellcolor{lightgray}Remove char (RC)} & \cellcolor{lightgray}L3-L2 & \cellcolor{lightgray}$0.16*$ & \cellcolor{lightgray}$0.21*$ & \cellcolor{lightgray}$0.28*$ &\cellcolor{lightgray}& \cellcolor{lightgray}$0.80$ & \cellcolor{lightgray}$0.90$ & \cellcolor{lightgray}$0.38*$\\
    \multirow{3}{*}{Misspelling (MS)} & L1-Ori & $0.75$ & $0.82$ & $0.43*$ && $0.87$ & $0.75$ & $0.85$\\
    & L2-L1 & $0.80$ & $0.68$ & $0.44*$ && $0.84$ & $0.61$ & $0.56$\\
    & L3-L2 & $0.23*$ & $0.29*$ & $0.29*$ && $0.88$ & $0.78$ & $0.55$\\
    \cellcolor{lightgray} & \cellcolor{lightgray}L1-Ori & \cellcolor{lightgray}$0.97$ & \cellcolor{lightgray}$0.93$ & \cellcolor{lightgray}$0.45*$ &\cellcolor{lightgray}& \cellcolor{lightgray}$0.97$ & \cellcolor{lightgray}$0.89$ & \cellcolor{lightgray}$0.89$\\
    \cellcolor{lightgray} & \cellcolor{lightgray}L2-L1 & \cellcolor{lightgray}$0.80$ & \cellcolor{lightgray}$0.93$ & \cellcolor{lightgray}$0.05*$ &\cellcolor{lightgray}& \cellcolor{lightgray}$0.80$ & \cellcolor{lightgray}$0.93$ & \cellcolor{lightgray}$0.41*$\\
    \multirow{-3}{*}{\cellcolor{lightgray}Swap (SW)} & \cellcolor{lightgray}L3-L2 & \cellcolor{lightgray}$0.28*$ & \cellcolor{lightgray}$0.31*$ & \cellcolor{lightgray}$0.48*$ &\cellcolor{lightgray}& \cellcolor{lightgray}$0.78$ & \cellcolor{lightgray}$0.98$ & \cellcolor{lightgray}$0.21*$\\
    \bottomrule
    \end{tabular}
    }
    \begin{tablenotes}
        \item[$\diamond$] A setting of ``A-B'' indicates that the test is conducted between uncertainty score on A and B, \eg, ``L1-Ori'' means the test is conducted between level 1 corrupted data and clean data.
        \item[$\dagger$] An effect size $\mathcal{D}$ indicates the probability that uncertainty score on corrupted data is larger than it on clean data; larger the effect size, more significant the result is.
        \item[$*$] This result is not statistically significant ($p>0.0001$).
    \end{tablenotes}
    \end{threeparttable}
    \label{tab:convlab_ood}
\end{table}

\begin{table}[htbp]
    \scriptsize
    \centering
    \caption{Wilcoxon signed-rank test  on SpeechBrain system's OOD score distribution difference between clean data and corrupted data on LibriSpeech dataset.}
    \begin{threeparttable}
    \resizebox{\linewidth}{!}{
    \begin{tabular}{llccccccccccc}
    \toprule
    \multirow{3}{*}{\textbf{Corruption}} & \multirow{3}{*}{\textbf{Settings$^\diamond$}} & \multicolumn{5}{c}{\textbf{Configuration 1}} & & \multicolumn{5}{c}{\textbf{Configuration 2}}\\
    & & \multicolumn{2}{c}{${\mathcal{D}_\textbf{Maha-d}}^\dagger$} && \multicolumn{2}{c}{${\mathcal{D}_\textbf{KDE}}^\dagger$} && \multicolumn{2}{c}{${\mathcal{D}_\textbf{Maha-d}}^\dagger$} && \multicolumn{2}{c}{${\mathcal{D}_\textbf{KDE}}^\dagger$}\\
    \cmidrule{3-4} \cmidrule{6-7} \cmidrule{9-10} \cmidrule{12-13}
    & & \multicolumn{1}{c}{CRDNN} & \multicolumn{1}{c}{RNN} && \multicolumn{1}{c}{CRDNN} & \multicolumn{1}{c}{RNN} & & \multicolumn{1}{c}{CRDNN} & \multicolumn{1}{c}{Trans-} && \multicolumn{1}{c}{CRDNN} & \multicolumn{1}{c}{Trans-}\\
    \midrule
    \cellcolor{lightgray} & \cellcolor{lightgray}L1-Ori & \cellcolor{lightgray}$1.00$ & \cellcolor{lightgray}$0.67$ &\cellcolor{lightgray}& \cellcolor{lightgray}$0.74$ & \cellcolor{lightgray}$1.00$ &\cellcolor{lightgray}& \cellcolor{lightgray}$1.00$ & \cellcolor{lightgray}$0.46^*$ &\cellcolor{lightgray}& \cellcolor{lightgray}$0.99$ & \cellcolor{lightgray}$1.00$\\
    \cellcolor{lightgray} & \cellcolor{lightgray}L2-L1 & \cellcolor{lightgray}$1.00$ & \cellcolor{lightgray}$0.45^*$ &\cellcolor{lightgray}& \cellcolor{lightgray}$0.69$ & \cellcolor{lightgray}$0.68$ &\cellcolor{lightgray}& \cellcolor{lightgray}$1.00$ & \cellcolor{lightgray}$0.45^*$ &\cellcolor{lightgray}& \cellcolor{lightgray}$1.00$ & \cellcolor{lightgray}$0.74$\\
    \multirow{-3}{*}{\cellcolor{lightgray}Reverb (RE)} & \cellcolor{lightgray}L3-L2 & \cellcolor{lightgray}$1.00$ & \cellcolor{lightgray}$0.56$ &\cellcolor{lightgray}& \cellcolor{lightgray}$0.64$ & \cellcolor{lightgray}$0.66$ &\cellcolor{lightgray}& \cellcolor{lightgray}$1.00$ & \cellcolor{lightgray}$0.58$ &\cellcolor{lightgray}& \cellcolor{lightgray}$1.00$ & \cellcolor{lightgray}$0.52^*$\\
    \multirow{3}{*}{Speed (SP)} & L1-Ori &$1.00$ & $0.72$ && $0.70$ & $1.00$ && $1.00$ & $0.51^*$ && $1.00$ & $1.00$\\
    & L2-L1 & $0.46^*$ & $0.58$ && $0.61$ & $0.56$ && $0.19^*$ & $0.41^*$ && $0.52^*$ & $0.83$\\
    & L3-L2 & $0.84$ & $0.54^*$ && $0.75$ & $0.57$ && $1.00$ & $0.65$ && $1.00$ & $0.66$\\
    \cellcolor{lightgray} & \cellcolor{lightgray}L1-Ori & \cellcolor{lightgray}$1.00$ & \cellcolor{lightgray}$0.72$ &\cellcolor{lightgray}& \cellcolor{lightgray}$0.93$ & \cellcolor{lightgray}$1.00$ &\cellcolor{lightgray}& \cellcolor{lightgray}$1.00$ & \cellcolor{lightgray}$0.50^*$ &\cellcolor{lightgray}& \cellcolor{lightgray}$1.00$ & \cellcolor{lightgray}$1.00$\\
    \cellcolor{lightgray} & \cellcolor{lightgray}L2-L1 & \cellcolor{lightgray}$0.92$ & \cellcolor{lightgray}$0.49^*$ &\cellcolor{lightgray}& \cellcolor{lightgray}$0.74$ & \cellcolor{lightgray}$0.48^*$ &\cellcolor{lightgray}& \cellcolor{lightgray}$1.00$ & \cellcolor{lightgray}$0.43^*$ &\cellcolor{lightgray}& \cellcolor{lightgray}$1.00$ & \cellcolor{lightgray}$0.73$\\
    \multirow{-3}{*}{\cellcolor{lightgray}Distortion (DI)} & \cellcolor{lightgray}L3-L2 & \cellcolor{lightgray}$0.99$ & \cellcolor{lightgray}$0.44^*$ &\cellcolor{lightgray}& \cellcolor{lightgray}$0.83$ & \cellcolor{lightgray}$0.52^*$ &\cellcolor{lightgray}& \cellcolor{lightgray}$1.00$ & \cellcolor{lightgray}$0.38^*$ &\cellcolor{lightgray}& \cellcolor{lightgray}$1.00$ & \cellcolor{lightgray}$0.81$\\
    \multirow{3}{*}{Pitch shift (PS)} & L1-Ori & $1.00$ & $0.73$ && $0.55$ & $1.00$ && $1.00$ & $0.50^*$ && $0.73$ & $1.00$\\
    & L2-L1 & $0.71$ & $0.48^*$ && $0.58$ & $0.48^*$ && $1.00$ & $0.44^*$ && $0.95$ & $0.65$\\
    & L3-L2 & $0.62$ & $0.48^*$ && $0.60$ & $0.50^*$ && $0.99$ & $0.45^*$ && $0.97$ & $0.70$\\
    \cellcolor{lightgray} & \cellcolor{lightgray}L1-Ori & \cellcolor{lightgray}$1.00$ & \cellcolor{lightgray}$0.69$ &\cellcolor{lightgray}& \cellcolor{lightgray}$0.91$ & \cellcolor{lightgray}$1.00$ &\cellcolor{lightgray}& \cellcolor{lightgray}$1.00$ & \cellcolor{lightgray}$0.50^*$ &\cellcolor{lightgray}& \cellcolor{lightgray}$1.00$ & \cellcolor{lightgray}$1.00$\\
    \cellcolor{lightgray} & \cellcolor{lightgray}L2-L1 & \cellcolor{lightgray}$0.80$ & \cellcolor{lightgray}$0.46^*$ &\cellcolor{lightgray}& \cellcolor{lightgray}$0.67$ & \cellcolor{lightgray}$0.48^*$ &\cellcolor{lightgray}& \cellcolor{lightgray}$1.00$ & \cellcolor{lightgray}$0.49^*$ &\cellcolor{lightgray}& \cellcolor{lightgray}$1.00$ & \cellcolor{lightgray}$0.59$\\
    \multirow{-3}{*}{\cellcolor{lightgray}Background noise (BN)} & \cellcolor{lightgray}L3-L2 & \cellcolor{lightgray}$0.68$ & \cellcolor{lightgray}$0.47^*$ &\cellcolor{lightgray}& \cellcolor{lightgray}$0.63$ & \cellcolor{lightgray}$0.51^*$ &\cellcolor{lightgray}& \cellcolor{lightgray}$1.00$ & \cellcolor{lightgray}$0.51^*$ &\cellcolor{lightgray}& \cellcolor{lightgray}$1.00$ & \cellcolor{lightgray}$0.52^*$\\
    \bottomrule
    \end{tabular}
    }
    \begin{tablenotes}
        \item[$\diamond$] A setting of ``A-B'' indicates that the test is conducted between uncertainty score on A and B, \eg, ``L1-Ori'' means the test is conducted between level 1 corrupted data and clean data.
        \item[$\dagger$] An effect size $\mathcal{D}$ indicates the probability that uncertainty score on corrupted data is larger than it on clean data; larger the effect size, more significant the result is.
        \item[$*$] This result is not statistically significant ($p>0.0001$).
    \end{tablenotes}
    \end{threeparttable}
    \label{tab:speech_ood1}
\end{table}

\begin{table}[htbp]
    \scriptsize
    \centering
    \caption{Wilcoxon signed-rank test on SpeechBrain system's OOD score distribution difference between clean data and corrupted data on CommonVoice-it dataset.}
    \begin{threeparttable}
    \resizebox{\linewidth}{!}{
    \begin{tabular}{llccccccccccc}
    \toprule
    \multirow{3}{*}{\textbf{Corruption}} & \multirow{3}{*}{\textbf{Settings$^\diamond$}} & \multicolumn{5}{c}{\textbf{Configuration 3}} & & \multicolumn{5}{c}{\textbf{Configuration 4}}\\
    & & \multicolumn{2}{c}{${\mathcal{D}_\textbf{Maha-d}}^\dagger$} && \multicolumn{2}{c}{${\mathcal{D}_\textbf{KDE}}^\dagger$} && \multicolumn{2}{c}{${\mathcal{D}_\textbf{Maha-d}}^\dagger$} && \multicolumn{2}{c}{${\mathcal{D}_\textbf{KDE}}^\dagger$}\\
    \cmidrule{3-4} \cmidrule{6-7} \cmidrule{9-10} \cmidrule{12-13}
    & & \multicolumn{1}{c}{CRDNN} & \multicolumn{1}{c}{RNN} && \multicolumn{1}{c}{CRDNN} & \multicolumn{1}{c}{RNN} & & \multicolumn{1}{c}{Wav2Vec2} & \multicolumn{1}{c}{RNN} && \multicolumn{1}{c}{Wav2Vec2} & \multicolumn{1}{c}{RNN}\\
    \midrule
    \cellcolor{lightgray} & \cellcolor{lightgray}L1-Ori & \cellcolor{lightgray}$1.00$ & \cellcolor{lightgray}$0.39^*$ &\cellcolor{lightgray}& \cellcolor{lightgray}$1.00$ & \cellcolor{lightgray}$1.00$ &\cellcolor{lightgray}& \cellcolor{lightgray}$0.93$ & \cellcolor{lightgray}$0.79$ &\cellcolor{lightgray}& \cellcolor{lightgray}$0.85$ & \cellcolor{lightgray}$1.00$\\
    \cellcolor{lightgray} & \cellcolor{lightgray}L2-L1 & \cellcolor{lightgray}$1.00$ & \cellcolor{lightgray}$0.18^*$ &\cellcolor{lightgray}& \cellcolor{lightgray}$0.97$ & \cellcolor{lightgray}$0.99$ &\cellcolor{lightgray}& \cellcolor{lightgray}$0.94$ & \cellcolor{lightgray}$0.87$ &\cellcolor{lightgray}& \cellcolor{lightgray}$0.92$ & \cellcolor{lightgray}$0.98$\\
    \multirow{-3}{*}{\cellcolor{lightgray}Reverb (RE)} & \cellcolor{lightgray}L3-L2 & \cellcolor{lightgray}$1.00$ & \cellcolor{lightgray}$0.29^*$ &\cellcolor{lightgray}& \cellcolor{lightgray}$0.85$ & \cellcolor{lightgray}$0.81$ &\cellcolor{lightgray}& \cellcolor{lightgray}$0.78$ & \cellcolor{lightgray}$0.86$ &\cellcolor{lightgray}& \cellcolor{lightgray}$0.93$ & \cellcolor{lightgray}$0.94$\\
    \multirow{3}{*}{Speed (SP)} & L1-Ori &$1.00$ & $0.49^*$ && $1.00$ & $1.00$ && $0.47^*$ & $0.70$ && $0.49^*$ & $1.00$\\
    & L2-L1 & $0.92$ & $0.19^*$ && $0.60$ & $0.91$ && $0.90$ & $0.67$ && $0.70$ & $0.82$\\
    & L3-L2 & $1.00$ & $0.32^*$ && $0.91$ & $0.76$ && $0.99$ & $0.97$ && $0.85$ & $0.99$\\
    \cellcolor{lightgray} & \cellcolor{lightgray}L1-Ori & \cellcolor{lightgray}$1.00$ & \cellcolor{lightgray}$0.53$ &\cellcolor{lightgray}& \cellcolor{lightgray}$1.00$ & \cellcolor{lightgray}$1.00$ &\cellcolor{lightgray}& \cellcolor{lightgray}$0.99$ & \cellcolor{lightgray}$0.68$ &\cellcolor{lightgray}& \cellcolor{lightgray}$0.72$ & \cellcolor{lightgray}$1.00$\\
    \cellcolor{lightgray} & \cellcolor{lightgray}L2-L1 & \cellcolor{lightgray}$0.98$ & \cellcolor{lightgray}$0.45^*$ &\cellcolor{lightgray}& \cellcolor{lightgray}$0.63$ & \cellcolor{lightgray}$0.92$ &\cellcolor{lightgray}& \cellcolor{lightgray}$1.00$ & \cellcolor{lightgray}$0.66$ &\cellcolor{lightgray}& \cellcolor{lightgray}$0.81$ & \cellcolor{lightgray}$0.86$\\
    \multirow{-3}{*}{\cellcolor{lightgray}Distortion (DI)} & \cellcolor{lightgray}L3-L2 & \cellcolor{lightgray}$0.97$ & \cellcolor{lightgray}$0.29^*$ &\cellcolor{lightgray}& \cellcolor{lightgray}$0.70$ & \cellcolor{lightgray}$0.97$ &\cellcolor{lightgray}& \cellcolor{lightgray}$1.00$ & \cellcolor{lightgray}$0.84$ &\cellcolor{lightgray}& \cellcolor{lightgray}$0.91$ & \cellcolor{lightgray}$0.96$\\
    \multirow{3}{*}{Pitch shift (PS)} & L1-Ori & $1.00$ & $0.52^*$ && $1.00$ & $1.00$ && $0.60$ & $0.65$ && $0.73$ & $1.00$\\
    & L2-L1 & $0.96$ & $0.42^*$ && $0.98$ & $0.91$ && $0.84$ & $0.61$ && $0.78$ & $0.76$\\
    & L3-L2 & $0.92$ & $0.36^*$ && $0.78$ & $0.95$ && $0.78$ & $0.70$ && $0.81$ & $0.87$\\
    \cellcolor{lightgray} & \cellcolor{lightgray}L1-Ori & \cellcolor{lightgray}$1.00$ & \cellcolor{lightgray}$0.36^*$ &\cellcolor{lightgray}& \cellcolor{lightgray}$1.00$ & \cellcolor{lightgray}$1.00$ &\cellcolor{lightgray}& \cellcolor{lightgray}$1.00$ & \cellcolor{lightgray}$0.73$ &\cellcolor{lightgray}& \cellcolor{lightgray}$1.00$ & \cellcolor{lightgray}$1.00$\\
    \cellcolor{lightgray} & \cellcolor{lightgray}L2-L1 & \cellcolor{lightgray}$1.00$ & \cellcolor{lightgray}$0.28^*$ &\cellcolor{lightgray}& \cellcolor{lightgray}$0.80$ & \cellcolor{lightgray}$0.94$ &\cellcolor{lightgray}& \cellcolor{lightgray}$1.00$ & \cellcolor{lightgray}$0.70$ &\cellcolor{lightgray}& \cellcolor{lightgray}$1.00$ & \cellcolor{lightgray}$0.90$\\
    \multirow{-3}{*}{\cellcolor{lightgray}Background noise (BN)} & \cellcolor{lightgray}L3-L2 & \cellcolor{lightgray}$1.00$ & \cellcolor{lightgray}$0.28^*$ &\cellcolor{lightgray}& \cellcolor{lightgray}$0.75$ & \cellcolor{lightgray}$0.85$ &\cellcolor{lightgray}& \cellcolor{lightgray}$0.99$ & \cellcolor{lightgray}$0.70$ &\cellcolor{lightgray}& \cellcolor{lightgray}$1.00$ & \cellcolor{lightgray}$0.87$\\
    \bottomrule
    \end{tabular}
    }
    \begin{tablenotes}
        \item[$\diamond$] A setting of ``A-B'' indicates that the test is conducted between uncertainty score on A and B, \eg, ``L1-Ori'' means the test is conducted between level 1 corrupted data and clean data.
        \item[$\dagger$] An effect size $\mathcal{D}$ indicates the probability that uncertainty score on corrupted data is larger than it on clean data; larger the effect size, more significant the result is.
        \item[$^*$] This result is not statistically significant ($p>0.0001$).
    \end{tablenotes}
    \end{threeparttable}
    \label{tab:speech_ood2}
\end{table}

Table~\ref{tab:tracking_ood}, Table~\ref{tab:convlab_ood}, and Table~\ref{tab:speech_ood1}\&\ref{tab:speech_ood2} show the Wilcoxon signed-rank test results on significance between the OOD score of clean data and corrupted data on ByteTrack system, ConvLab-2 system, and SpeechBrain system, respectively. A positive effect size in these tables indicates that this module's OOD score on corrupted data is significantly larger than on clean data. The larger the value, the more significant the result can be. In the ByteTrack system, we find that no matter whether using Maha-d or KDE, the YOLO detector's OOD score measured on corrupted data is larger than it is on clean data. When the severity level of corruption increases, such difference also becomes more significant. This suggests that both these two OOD detection techniques can well capture the data distribution between clean data and corrupted data on the YOLO detector module. However, it becomes harder for these two methods to measure the data distribution difference on the second module of the ByteTrack system, \ie, the DeepSORT tracker. Specifically, we find that KDE failed to capture the data distribution difference between clean and corrupted data. In the ConvLab-2 system, we find that both these two methods can well measure the data distribution difference on either BERT or MILU NLU module (which is the first module in the system). Nevertheless, Maha-d fails to measure it on the PPO policy module, while KDE succeeds. The observation of these two systems suggests that for a serial execution system, OOD detection techniques are better at measuring the data distribution difference on the first module of the system. We further verify this finding on the SpeechBrain system, where we find that Maha-d method fails on the second module of Configuration 3 and 4. One plausible explanation for these observations is: that the first module in a serial-executed AI system usually directly takes the corrupted data as input, therefore, it can better capture the data distribution difference.

\subsubsection{OOD awareness of different modules inside AI systems.}
\label{sec:ood_diff}

\begin{table}[htbp]
    \scriptsize
    \centering
    \caption{Similarity of top-1\% OOD data detected according two different modules in SpeechBrain system.}
    \begin{threeparttable}
    \resizebox{\linewidth}{!}{
    \begin{tabular}{lcccccccccccc}
    \toprule
    \multirow{2}{*}{\textbf{Corruption}} & \multirow{2}{*}{\textbf{Severity}} & \multicolumn{2}{c}{\textbf{Configuration 1}$^\dagger$} && \multicolumn{2}{c}{\textbf{Configuration 2}$^\dagger$} && \multicolumn{2}{c}{\textbf{Configuration 3}$^\diamond$} && \multicolumn{2}{c}{\textbf{Configuration 4}$^\diamond$}\\
    \cmidrule{3-4} \cmidrule{6-7} \cmidrule{9-10} \cmidrule{12-13}
    & & \multicolumn{1}{c}{Maha-d} & \multicolumn{1}{c}{KDE} && \multicolumn{1}{c}{Maha-d} & \multicolumn{1}{c}{KDE} && \multicolumn{1}{c}{Maha-d} & \multicolumn{1}{c}{KDE} && \multicolumn{1}{c}{Maha-d} & \multicolumn{1}{c}{KDE}\\
    \midrule
    \cellcolor{lightgray} & \cellcolor{lightgray}1  & \cellcolor{lightgray}$0.0\%$ & \cellcolor{lightgray}$0.0\%$ &\cellcolor{lightgray}& \cellcolor{lightgray}$7.7\%$ & \cellcolor{lightgray}$3.8\%$ &\cellcolor{lightgray}& \cellcolor{lightgray}$1.1\%$ & \cellcolor{lightgray}$0.0\%$ & \cellcolor{lightgray} & \cellcolor{lightgray}$0.0\%$ & \cellcolor{lightgray}$0.0\%$\\
    \cellcolor{lightgray} & \cellcolor{lightgray}2  & \cellcolor{lightgray}$0.0\%$ & \cellcolor{lightgray}$0.0\%$ &\cellcolor{lightgray}& \cellcolor{lightgray}$0.0\%$ & \cellcolor{lightgray}$0.0\%$ &\cellcolor{lightgray}& \cellcolor{lightgray}$1.1\%$ & \cellcolor{lightgray}$0.0\%$ & \cellcolor{lightgray} & \cellcolor{lightgray}$0.0\%$ & \cellcolor{lightgray}$0.0\%$\\
    \multirow{-3}{*}{\cellcolor{lightgray}Reverb (RE)} & \cellcolor{lightgray}3 & \cellcolor{lightgray}$0.0\%$ & \cellcolor{lightgray}$0.0\%$ &\cellcolor{lightgray}& \cellcolor{lightgray}$0.0\%$ & \cellcolor{lightgray}$3.8\%$ &\cellcolor{lightgray}& \cellcolor{lightgray}$1.1\%$ & \cellcolor{lightgray}$0.0\%$ & \cellcolor{lightgray} & \cellcolor{lightgray}$1.1\%$ & \cellcolor{lightgray}$1.1\%$\\
    \multirow{3}{*}{Speed (SP)}  & 1        & $3.8\%$ & $0.0\%$ & & $3.8\%$ & $0.0\%$ && $2.2\%$ & $0.0\%$ & & $1.1\%$ & $5.6\%$\\
    & 2        & $0.0\%$ & $11.5\%$ & & $11.5\%$ & $3.8\%$ && $1.1\%$ & $0.0\%$ & & $0.0\%$ & $0.0\%$\\
    & 3        & $0.0\%$ & $0.0\%$ & & $0.0\%$ & $3.8\%$ && $4.5\%$ & $0.0\%$ & & $2.2\%$ & $1.1\%$\\
    \cellcolor{lightgray} & \cellcolor{lightgray}1  & \cellcolor{lightgray}$0.0\%$ & \cellcolor{lightgray}$0.0\%$ &\cellcolor{lightgray}& \cellcolor{lightgray}$11.5\%$ & \cellcolor{lightgray}$0.0\%$ &\cellcolor{lightgray}& \cellcolor{lightgray}$4.5\%$ & \cellcolor{lightgray}$0.0\%$ & \cellcolor{lightgray} & \cellcolor{lightgray}$4.5\%$ & \cellcolor{lightgray}$4.5\%$\\
    \cellcolor{lightgray} & \cellcolor{lightgray}2  & \cellcolor{lightgray}$0.0\%$ & \cellcolor{lightgray}$1.1\%$ &\cellcolor{lightgray}& \cellcolor{lightgray}$3.8\%$ & \cellcolor{lightgray}$0.0\%$ &\cellcolor{lightgray}& \cellcolor{lightgray}$5.6\%$ & \cellcolor{lightgray}$1.1\%$ & \cellcolor{lightgray} & \cellcolor{lightgray}$4.5\%$ & \cellcolor{lightgray}$4.5\%$\\
    \multirow{-3}{*}{\cellcolor{lightgray}Distortion (DI)} & \cellcolor{lightgray}3 & \cellcolor{lightgray}$0.0\%$ & \cellcolor{lightgray}$3.4\%$ &\cellcolor{lightgray}& \cellcolor{lightgray}$0.0\%$ & \cellcolor{lightgray}$0.0\%$ &\cellcolor{lightgray}& \cellcolor{lightgray}$2.2\%$ & \cellcolor{lightgray}$3.4\%$ & \cellcolor{lightgray} & \cellcolor{lightgray}$3.4\%$ & \cellcolor{lightgray}$4.5\%$\\
    \multirow{3}{*}{Pitch shift (PS)}  & 1        & $0.0\%$ & $0.0\%$ & & $7.7\%$ & $0.0\%$ && $0.0\%$ & $0.0\%$ & & $0.0\%$ & $3.4\%$\\
    & 2        & $0.0\%$ & $0.0\%$ & & $3.8\%$ & $3.8\%$ && $4.5\%$ & $0.0\%$ & & $1.1\%$ & $2.2\%$\\
    & 3        & $0.0\%$ & $0.0\%$ & & $7.7\%$ & $3.8\%$ && $2.2\%$ & $0.0\%$ & & $1.1\%$ & $1.1\%$\\
    \cellcolor{lightgray} & \cellcolor{lightgray}1  & \cellcolor{lightgray}$0.0\%$ & \cellcolor{lightgray}$0.0\%$ &\cellcolor{lightgray}& \cellcolor{lightgray}$0.0\%$ & \cellcolor{lightgray}$3.8\%$ &\cellcolor{lightgray}& \cellcolor{lightgray}$0.0\%$ & \cellcolor{lightgray}$0.0\%$ & \cellcolor{lightgray} & \cellcolor{lightgray}$3.4\%$ & \cellcolor{lightgray}$0.0\%$\\
    \cellcolor{lightgray} & \cellcolor{lightgray}2  & \cellcolor{lightgray}$0.0\%$ & \cellcolor{lightgray}$0.0\%$ &\cellcolor{lightgray}& \cellcolor{lightgray}$0.0\%$ & \cellcolor{lightgray}$7.7\%$ &\cellcolor{lightgray}& \cellcolor{lightgray}$0.0\%$ & \cellcolor{lightgray}$0.0\%$ & \cellcolor{lightgray} & \cellcolor{lightgray}$2.2\%$ & \cellcolor{lightgray}$0.0\%$\\
    \multirow{-3}{*}{\cellcolor{lightgray}Background noise (BN)} & \cellcolor{lightgray}3 & \cellcolor{lightgray}$0.0\%$ & \cellcolor{lightgray}$0.0\%$ &\cellcolor{lightgray}& \cellcolor{lightgray}$3.8\%$ & \cellcolor{lightgray}$11.5\%$ &\cellcolor{lightgray}& \cellcolor{lightgray}$0.0\%$ & \cellcolor{lightgray}$0.0\%$ & \cellcolor{lightgray} & \cellcolor{lightgray}$1.1\%$ & \cellcolor{lightgray}$0.0\%$\\
    \bottomrule
    \end{tabular}
    }
    \begin{tablenotes}
        \item[$\dagger$] Configuration 1 and Configuration 2 is tested on LibriSpeech dataset, where $1\%$ data has a size of 26 instances.
        \item[$\diamond$] Configuration 3 and Configuration 4 is tested on CommonVoice-it dataset, where $1\%$ data has a size of 89 instances.
    \end{tablenotes}
    \end{threeparttable}
    \label{tab:speech_ood_diff}
\end{table}

\begin{table}[htbp]
    \scriptsize
    \centering
    \caption{Similarity of top-5\% OOD data detected according two different modules in SpeechBrain system.}
    \begin{threeparttable}
    \resizebox{\linewidth}{!}{
    \begin{tabular}{lcccccccccccc}
    \toprule
    \multirow{2}{*}{\textbf{Corruption}} & \multirow{2}{*}{\textbf{Severity}} & \multicolumn{2}{c}{\textbf{Configuration 1}$^\dagger$} && \multicolumn{2}{c}{\textbf{Configuration 2}$^\dagger$} && \multicolumn{2}{c}{\textbf{Configuration 3}$^\diamond$} && \multicolumn{2}{c}{\textbf{Configuration 4}$^\diamond$}\\
    \cmidrule{3-4} \cmidrule{6-7} \cmidrule{9-10} \cmidrule{12-13}
    & & \multicolumn{1}{c}{Maha-d} & \multicolumn{1}{c}{KDE} && \multicolumn{1}{c}{Maha-d} & \multicolumn{1}{c}{KDE} && \multicolumn{1}{c}{Maha-d} & \multicolumn{1}{c}{KDE} && \multicolumn{1}{c}{Maha-d} & \multicolumn{1}{c}{KDE}\\
    \midrule
    \cellcolor{lightgray} & \cellcolor{lightgray}1  & \cellcolor{lightgray}$0.8\%$ & \cellcolor{lightgray}$6.9\%$ &\cellcolor{lightgray}& \cellcolor{lightgray}$12.3\%$ & \cellcolor{lightgray}$5.4\%$ &\cellcolor{lightgray}& \cellcolor{lightgray}$2.0\%$ & \cellcolor{lightgray}$0.0\%$ & \cellcolor{lightgray} & \cellcolor{lightgray}$0.9\%$ & \cellcolor{lightgray}$1.3\%$\\
    \cellcolor{lightgray} & \cellcolor{lightgray}2  & \cellcolor{lightgray}$4.6\%$ & \cellcolor{lightgray}$6.2\%$ &\cellcolor{lightgray}& \cellcolor{lightgray}$6.9\%$ & \cellcolor{lightgray}$4.6\%$ &\cellcolor{lightgray}& \cellcolor{lightgray}$3.0\%$ & \cellcolor{lightgray}$0.1\%$ & \cellcolor{lightgray} & \cellcolor{lightgray}$0.9\%$ & \cellcolor{lightgray}$4.5\%$\\
    \multirow{-3}{*}{\cellcolor{lightgray}Reverb (RE)} & \cellcolor{lightgray}3 & \cellcolor{lightgray}$3.8\%$ & \cellcolor{lightgray}$4.6\%$ &\cellcolor{lightgray}& \cellcolor{lightgray}$1.5\%$ & \cellcolor{lightgray}$6.9\%$ &\cellcolor{lightgray}& \cellcolor{lightgray}$3.4\%$ & \cellcolor{lightgray}$0.1\%$ & \cellcolor{lightgray} & \cellcolor{lightgray}$0.9\%$ & \cellcolor{lightgray}$3.6\%$\\
    \multirow{3}{*}{Speed (SP)}  & 1        & $4.6\%$ & $5.4\%$ & & $6.2\%$ & $3.8\%$ && $3.1\%$ & $0.0\%$ & & $3.1\%$ & $8.9\%$\\
    & 2        & $1.5\%$ & $3.8\%$ & & $15.4\%$ & $5.4\%$ && $2.0\%$ & $0.1\%$ & & $1.3\%$ & $6.0\%$\\
    & 3        & $5.4\%$ & $5.4\%$ & & $1.5\%$ & $4.6\%$ && $6.3\%$ & $0.0\%$ & & $4.0\%$ & $4.3\%$\\
    \cellcolor{lightgray} & \cellcolor{lightgray}1  & \cellcolor{lightgray}$3.8\%$ & \cellcolor{lightgray}$3.1\%$ &\cellcolor{lightgray}& \cellcolor{lightgray}$15.4\%$ & \cellcolor{lightgray}$3.8\%$ &\cellcolor{lightgray}& \cellcolor{lightgray}$10.1\%$ & \cellcolor{lightgray}$2.2\%$ & \cellcolor{lightgray} & \cellcolor{lightgray}$13.6\%$ & \cellcolor{lightgray}$8.7\%$\\
    \cellcolor{lightgray} & \cellcolor{lightgray}2  & \cellcolor{lightgray}$1.5\%$ & \cellcolor{lightgray}$5.4\%$ &\cellcolor{lightgray}& \cellcolor{lightgray}$16.9\%$ & \cellcolor{lightgray}$5.4\%$ &\cellcolor{lightgray}& \cellcolor{lightgray}$10.2\%$ & \cellcolor{lightgray}$3.1\%$ & \cellcolor{lightgray} & \cellcolor{lightgray}$11.6\%$ & \cellcolor{lightgray}$9.8\%$\\
    \multirow{-3}{*}{\cellcolor{lightgray}Distortion (DI)} & \cellcolor{lightgray}3 & \cellcolor{lightgray}$0.1\%$ & \cellcolor{lightgray}$6.9\%$ &\cellcolor{lightgray}& \cellcolor{lightgray}$7.7\%$ & \cellcolor{lightgray}$6.2\%$ &\cellcolor{lightgray}& \cellcolor{lightgray}$8.9\%$ & \cellcolor{lightgray}$12.1\%$ & \cellcolor{lightgray} & \cellcolor{lightgray}$9.6\%$ & \cellcolor{lightgray}$18.3\%$\\
    \multirow{3}{*}{Pitch shift (PS)}  & 1        & $3.8\%$ & $3.8\%$ & & $16.9\%$ & $8.5\%$ && $3.8\%$ & $0.1\%$ & & $2.7\%$ & $5.6\%$\\
    & 2        & $3.8\%$ & $4.6\%$ & & $26.1\%$ & $12.3\%$ && $3.8\%$ & $0.1\%$ & & $3.4\%$ & $5.1\%$\\
    & 3        & $4.6\%$ & $8.5\%$ & & $17.7\%$ & $13.8\%$ && $2.5\%$ & $0.1\%$ & & $3.1\%$ & $5.6\%$\\
    \cellcolor{lightgray} & \cellcolor{lightgray}1  & \cellcolor{lightgray}$3.1\%$ & \cellcolor{lightgray}$6.9\%$ &\cellcolor{lightgray}& \cellcolor{lightgray}$10.8\%$ & \cellcolor{lightgray}$18.5\%$ &\cellcolor{lightgray}& \cellcolor{lightgray}$1.8\%$ & \cellcolor{lightgray}$0.1\%$ & \cellcolor{lightgray} & \cellcolor{lightgray}$6.5\%$ & \cellcolor{lightgray}$2.5\%$\\
    \cellcolor{lightgray} & \cellcolor{lightgray}2  & \cellcolor{lightgray}$3.1\%$ & \cellcolor{lightgray}$6.2\%$ &\cellcolor{lightgray}& \cellcolor{lightgray}$8.5\%$ & \cellcolor{lightgray}$20.0\%$ &\cellcolor{lightgray}& \cellcolor{lightgray}$2.0\%$ & \cellcolor{lightgray}$1.8\%$ & \cellcolor{lightgray} & \cellcolor{lightgray}$4.3\%$ & \cellcolor{lightgray}$2.2\%$\\
    \multirow{-3}{*}{\cellcolor{lightgray}Background noise (BN)} & \cellcolor{lightgray}3 & \cellcolor{lightgray}$3.8\%$ & \cellcolor{lightgray}$5.4\%$ &\cellcolor{lightgray}& \cellcolor{lightgray}$6.2\%$ & \cellcolor{lightgray}$20.8\%$ &\cellcolor{lightgray}& \cellcolor{lightgray}$1.6\%$ & \cellcolor{lightgray}$3.8\%$ & \cellcolor{lightgray} & \cellcolor{lightgray}$5.1\%$ & \cellcolor{lightgray}$0.1\%$\\
    \bottomrule
    \end{tabular}
    }
    \begin{tablenotes}
        \item[$\dagger$] Configuration 1 and Configuration 2 is tested on LibriSpeech dataset, where $5\%$ data has a size of 130 instances.
        \item[$\diamond$] Configuration 3 and Configuration 4 is tested on CommonVoice-it dataset, where $5\%$ data has a size of 447 instances.
    \end{tablenotes}
    \end{threeparttable}
    \label{tab:speech_ood_diff5}
\end{table}

\begin{table}[htbp]
    \scriptsize
    \centering
    \caption{Similarity of top-10\% OOD data detected according two different modules in SpeechBrain system.}
    \begin{threeparttable}
    \resizebox{\linewidth}{!}{
    \begin{tabular}{lcccccccccccc}
    \toprule
    \multirow{2}{*}{\textbf{Corruption}} & \multirow{2}{*}{\textbf{Severity}} & \multicolumn{2}{c}{\textbf{Configuration 1}$^\dagger$} && \multicolumn{2}{c}{\textbf{Configuration 2}$^\dagger$} && \multicolumn{2}{c}{\textbf{Configuration 3}$^\diamond$} && \multicolumn{2}{c}{\textbf{Configuration 4}$^\diamond$}\\
    \cmidrule{3-4} \cmidrule{6-7} \cmidrule{9-10} \cmidrule{12-13}
    & & \multicolumn{1}{c}{Maha-d} & \multicolumn{1}{c}{KDE} && \multicolumn{1}{c}{Maha-d} & \multicolumn{1}{c}{KDE} && \multicolumn{1}{c}{Maha-d} & \multicolumn{1}{c}{KDE} && \multicolumn{1}{c}{Maha-d} & \multicolumn{1}{c}{KDE}\\
    \midrule
    \cellcolor{lightgray} & \cellcolor{lightgray}1  & \cellcolor{lightgray}$6.5\%$ & \cellcolor{lightgray}$12.3\%$ &\cellcolor{lightgray}& \cellcolor{lightgray}$24.1\%$ & \cellcolor{lightgray}$6.5\%$ &\cellcolor{lightgray}& \cellcolor{lightgray}$7.3\%$ & \cellcolor{lightgray}$1.1\%$ & \cellcolor{lightgray} & \cellcolor{lightgray}$4.0\%$ & \cellcolor{lightgray}$2.8\%$\\
    \cellcolor{lightgray} & \cellcolor{lightgray}2  & \cellcolor{lightgray}$6.5\%$ & \cellcolor{lightgray}$11.5\%$ &\cellcolor{lightgray}& \cellcolor{lightgray}$12.3\%$ & \cellcolor{lightgray}$9.2\%$ &\cellcolor{lightgray}& \cellcolor{lightgray}$8.0\%$ & \cellcolor{lightgray}$1.2\%$ & \cellcolor{lightgray} & \cellcolor{lightgray}$2.5\%$ & \cellcolor{lightgray}$7.5\%$\\
    \multirow{-3}{*}{\cellcolor{lightgray}Reverb (RE)} & \cellcolor{lightgray}3 & \cellcolor{lightgray}$5.7\%$ & \cellcolor{lightgray}$11.1\%$ &\cellcolor{lightgray}& \cellcolor{lightgray}$7.3\%$ & \cellcolor{lightgray}$12.3\%$ &\cellcolor{lightgray}& \cellcolor{lightgray}$7.2\%$ & \cellcolor{lightgray}$0.9\%$ & \cellcolor{lightgray} & \cellcolor{lightgray}$3.1\%$ & \cellcolor{lightgray}$6.1\%$\\
    \multirow{3}{*}{Speed (SP)}  & 1        & $7.7\%$ & $8.8\%$ & & $13.4\%$ & $8.4\%$ && $5.7\%$ & $1.2\%$ & & $8.9\%$ & $17.8\%$\\
    & 2        & $5.4\%$ & $11.1\%$ & & $27.6\%$ & $20.3\%$ && $5.6\%$ & $2.6\%$ & & $5.5\%$ & $7.9\%$\\
    & 3        & $10.0\%$ & $11.5\%$ & & $4.6\%$ & $15.7\%$ && $11.1\%$ & $0.3\%$ & & $6.6\%$ & $7.3\%$\\
    \cellcolor{lightgray} & \cellcolor{lightgray}1  & \cellcolor{lightgray}$6.1\%$ & \cellcolor{lightgray}$9.2\%$ &\cellcolor{lightgray}& \cellcolor{lightgray}$22.6\%$ & \cellcolor{lightgray}$10.3\%$ &\cellcolor{lightgray}& \cellcolor{lightgray}$16.1\%$ & \cellcolor{lightgray}$5.9\%$ & \cellcolor{lightgray} & \cellcolor{lightgray}$19.6\%$ & \cellcolor{lightgray}$11.4\%$\\
    \cellcolor{lightgray} & \cellcolor{lightgray}2  & \cellcolor{lightgray}$4.2\%$ & \cellcolor{lightgray}$8.0\%$ &\cellcolor{lightgray}& \cellcolor{lightgray}$18.4\%$ & \cellcolor{lightgray}$14.6\%$ &\cellcolor{lightgray}& \cellcolor{lightgray}$14.5\%$ & \cellcolor{lightgray}$10.7\%$ & \cellcolor{lightgray} & \cellcolor{lightgray}$18.4\%$ & \cellcolor{lightgray}$15.1\%$\\
    \multirow{-3}{*}{\cellcolor{lightgray}Distortion (DI)} & \cellcolor{lightgray}3 & \cellcolor{lightgray}$3.4\%$ & \cellcolor{lightgray}$10.7\%$ &\cellcolor{lightgray}& \cellcolor{lightgray}$16.1\%$ & \cellcolor{lightgray}$16.1\%$ &\cellcolor{lightgray}& \cellcolor{lightgray}$14.0\%$ & \cellcolor{lightgray}$20.7\%$ & \cellcolor{lightgray} & \cellcolor{lightgray}$15.5\%$ & \cellcolor{lightgray}$19.8\%$\\
    \multirow{3}{*}{Pitch shift (PS)}  & 1        & $8.4\%$ & $11.9\%$ & & $26.4\%$ & $16.5\%$ && $7.7\%$ & $2.1\%$ & & $7.0\%$ & $9.1\%$\\
    & 2        & $10.3\%$ & $12.3\%$ & & $36.0\%$ & $22.2\%$ && $7.2\%$ & $1.7\%$ & & $6.1\%$ & $7.8\%$\\
    & 3        & $6.9\%$ & $12.6\%$ & & $30.7\%$ & $25.7\%$ && $6.5\%$ & $2.6\%$ & & $5.9\%$ & $8.7\%$\\
    \cellcolor{lightgray} & \cellcolor{lightgray}1  & \cellcolor{lightgray}$5.0\%$ & \cellcolor{lightgray}$10.3\%$ &\cellcolor{lightgray}& \cellcolor{lightgray}$18.0\%$ & \cellcolor{lightgray}$34.5\%$ &\cellcolor{lightgray}& \cellcolor{lightgray}$5.4\%$ & \cellcolor{lightgray}$3.8\%$ & \cellcolor{lightgray} & \cellcolor{lightgray}$7.3\%$ & \cellcolor{lightgray}$3.9\%$\\
    \cellcolor{lightgray} & \cellcolor{lightgray}2  & \cellcolor{lightgray}$4.2\%$ & \cellcolor{lightgray}$10.7\%$ &\cellcolor{lightgray}& \cellcolor{lightgray}$19.9\%$ & \cellcolor{lightgray}$34.1\%$ &\cellcolor{lightgray}& \cellcolor{lightgray}$4.5\%$ & \cellcolor{lightgray}$5.4\%$ & \cellcolor{lightgray} & \cellcolor{lightgray}$6.9\%$ & \cellcolor{lightgray}$4.2\%$\\
    \multirow{-3}{*}{\cellcolor{lightgray}Background noise (BN)} & \cellcolor{lightgray}3 & \cellcolor{lightgray}$5.4\%$ & \cellcolor{lightgray}$8.8\%$ &\cellcolor{lightgray}& \cellcolor{lightgray}$16.5\%$ & \cellcolor{lightgray}$27.6\%$ &\cellcolor{lightgray}& \cellcolor{lightgray}$5.4\%$ & \cellcolor{lightgray}$6.1\%$ & \cellcolor{lightgray} & \cellcolor{lightgray}$7.8\%$ & \cellcolor{lightgray}$5.9\%$\\
    \bottomrule
    \end{tabular}
    }
    \begin{tablenotes}
        \item[$\dagger$] Configuration 1 and Configuration 2 is tested on LibriSpeech dataset, where $10\%$ data has a size of 260 instances.
        \item[$\diamond$] Configuration 3 and Configuration 4 is tested on CommonVoice-it dataset, where $10\%$ data has a size of 894 instances.
    \end{tablenotes}
    \end{threeparttable}
    \label{tab:speech_ood_diff10}
\end{table}

While the OOD techniques used in this study can only measure on AI module, hence in this subsection, we mainly investigate the results from the SpeechBrain system and ByteTrack system (with DeepSORT tracker), where two modules in the system are both AI ones. We select the $1\%$ test data with the highest OOD score detected on two different modules, then analyze if they are the same on two modules. We measure such similarity according to Eq.~\ref{eq:sim}.

\begin{equation}
    \label{eq:sim}
    sim = \frac{|\mathcal{S}^{n}_{module 1}\cap\mathcal{S}^{n}_{module 2}|}{|\mathcal{S}^{n}_{module 1}|}
\end{equation}
where $\mathcal{S}^{n}_{module 1}$ and $\mathcal{S}^{n}_{module 2}$ denotes the set of top-$n$\% OOD data detected by module 1 and module 2, respectively.

\begin{table}[htbp]
    \scriptsize
    \centering
    \caption{Similarity of top-1\%, top-5\%, and top-10\% OOD data detected according two different modules in ByteTrack system (YOLO+DeepSORT).}
    \begin{threeparttable}
    \begin{tabular}{lccccccccc}
    \toprule
    \multirow{2}{*}{\textbf{Corruption}} & \multirow{2}{*}{\textbf{Severity}} & \multicolumn{2}{c}{Top-1\%} & & \multicolumn{2}{c}{Top-5\%} & & \multicolumn{2}{c}{Top-10\%}\\
    \cmidrule{3-4} \cmidrule{6-7} \cmidrule{9-10}
    &  & \multicolumn{1}{c}{Maha-d} & \multicolumn{1}{c}{KDE} & & \multicolumn{1}{c}{Maha-d} & \multicolumn{1}{c}{KDE}& & \multicolumn{1}{c}{Maha-d} & \multicolumn{1}{c}{KDE}\\
    \midrule
    \cellcolor{lightgray} & \cellcolor{lightgray}1  & \cellcolor{lightgray}$0.0\%$ & \cellcolor{lightgray}$3.8\%$ & \cellcolor{lightgray} & \cellcolor{lightgray}$2.3\%$ & \cellcolor{lightgray}$0.0\%$ & \cellcolor{lightgray} & \cellcolor{lightgray}$4.2\%$ & \cellcolor{lightgray}$1.9\%$\\
    \cellcolor{lightgray} & \cellcolor{lightgray}2  & \cellcolor{lightgray}$3.8\%$ & \cellcolor{lightgray}$0.0\%$ & \cellcolor{lightgray} & \cellcolor{lightgray}$12.1\%$ & \cellcolor{lightgray}$1.5\%$ & \cellcolor{lightgray} & \cellcolor{lightgray}$33.2\%$ & \cellcolor{lightgray}$1.5\%$\\
    \multirow{-3}{*}{\cellcolor{lightgray}Gaussian noise (GN)} & \cellcolor{lightgray}3 & \cellcolor{lightgray}$3.8\%$ & \cellcolor{lightgray}$0.0\%$ & \cellcolor{lightgray} & \cellcolor{lightgray}$22.0\%$ & \cellcolor{lightgray}$5.3\%$ & \cellcolor{lightgray} & \cellcolor{lightgray}$45.7\%$ & \cellcolor{lightgray}$10.9\%$\\
    \multirow{3}{*}{Motion blur (MB)}  & 1        & $0.0\%$ & $0.0\%$ & & $7.6\%$ & $3.0\%$ & & $14.7\%$ & $6.8\%$\\
    & 2        & $0.0\%$ & $3.8\%$ & & $3.0\%$ & $1.5\%$ & & $6.8\%$ & $6.8\%$\\
    & 3        & $0.0\%$ & $0.0\%$ & & $1.5\%$ & $0.0\%$ & & $5.7\%$ & $2.3\&$\\
    \cellcolor{lightgray} & \cellcolor{lightgray}1  & \cellcolor{lightgray}$0.0\%$ & \cellcolor{lightgray}$3.8\%$ & \cellcolor{lightgray} & \cellcolor{lightgray}$2.3\%$ & \cellcolor{lightgray}$3.0\%$ & \cellcolor{lightgray} & \cellcolor{lightgray}$4.2\%$ & \cellcolor{lightgray}$6.0\%$\\
    \cellcolor{lightgray} & \cellcolor{lightgray}2  & \cellcolor{lightgray}$0.0\%$ & \cellcolor{lightgray}$3.8\%$ & \cellcolor{lightgray} & \cellcolor{lightgray}$6.8\%$ & \cellcolor{lightgray}$0.1$ & \cellcolor{lightgray} & \cellcolor{lightgray}$11.7\%$ & \cellcolor{lightgray}$4.9\%$\\
    \multirow{-3}{*}{\cellcolor{lightgray}Contrast (CO)} & \cellcolor{lightgray}3 & \cellcolor{lightgray}$7.7\%$ & \cellcolor{lightgray}$0.0\%$ & \cellcolor{lightgray} & \cellcolor{lightgray}$15.2\%$ & \cellcolor{lightgray}$3.8$ & \cellcolor{lightgray} & \cellcolor{lightgray}$40.0\%$ & \cellcolor{lightgray}$5.7\%$\\
    \multirow{3}{*}{Snow (SN)}  & 1        & $0.0\%$ & $0.0\%$ & & $1.5\%$ & $0.0\%$ & & $5.7\%$ & $4.2\%$\\
    & 2        & $0.0\%$ & $0.0\%$ & & $7.6\%$ & $0.1\%$ & & $20.0\%$ & $4.2\%$\\
    & 3        & $0.0\%$ & $0.0\%$ & & $6.8\%$ & $2.2\%$ & & $21.9\%$ & $7.5\%$\\
    \cellcolor{lightgray} & \cellcolor{lightgray}1  & \cellcolor{lightgray}$11.5\%$ & \cellcolor{lightgray}$0.0\%$ & \cellcolor{lightgray} & \cellcolor{lightgray}$12.1\%$ & \cellcolor{lightgray}$6.8\%$ & \cellcolor{lightgray} & \cellcolor{lightgray}$15.5\%$ & \cellcolor{lightgray}$14.7\%$\\
    \cellcolor{lightgray} & \cellcolor{lightgray}2  & \cellcolor{lightgray}$3.8\%$ & \cellcolor{lightgray}$0.0\%$ & \cellcolor{lightgray} & \cellcolor{lightgray}$12.9\%$ & \cellcolor{lightgray}$1.5\%$ & \cellcolor{lightgray} & \cellcolor{lightgray}$14.3\%$ & \cellcolor{lightgray}$9.1\%$\\
    \multirow{-3}{*}{\cellcolor{lightgray}Pixelate (PI)} & \cellcolor{lightgray}3 & \cellcolor{lightgray}$0.0\%$ & \cellcolor{lightgray}$0.0\%$ & \cellcolor{lightgray} & \cellcolor{lightgray}$5.3\%$ & \cellcolor{lightgray}$1.5\%$ & \cellcolor{lightgray} & \cellcolor{lightgray}$19.2\%$ & \cellcolor{lightgray}$4.2\%$\\
    \bottomrule
    \end{tabular}
    \end{threeparttable}
    \label{tab:tracking_ood_diff}
\end{table}

{
Table~\ref{tab:speech_ood_diff}, \ref{tab:speech_ood_diff5}, \ref{tab:speech_ood_diff10} and Table~\ref{tab:tracking_ood_diff} summarize the similarity of top-1\%, top-5\%, and top-10\% OOD data detected on two different modules in the SpeechBrain system and ByteTrack system, respectively. In most cases, we can find that two modules in the SpeechBrain system detect different OOD data. When only considering top-1\% data, the highest similarity (11.5\%, Configuration 2 against Background noise corruption, measured by KDE) only suggests that there are 3 OOD instances that both two modules detect. When considering top-10\% data, even the highest similarity only suggest less than half of OOD data are similar (36\%, Configuration 2 against Pitch shift corruption, measured by Maha-d) .We can observe similar symptoms in the ByteTrack system, where the top-1\% OOD data are very different for two modules (\ie, detector and tracker). These results reveal that different modules in an AI system could treat different data instances as OOD data.

To further investigate the difference of OOD data identified by different modules, we use a threshold-independent assessment, \ie, \textit{Spearman's rank correlation coefficient} to measure such differences. We show the results in Table~\ref{tab:speech_ood_spearman} and Table~\ref{tab:tracking_ood_spearmanr}. We can find there is no strong evidence showing that there are correlations between two modules' OOD data ranking in SpeechBrain system when measured by either Maha-d or KDE. These results suggest that the OOD data identified by these two modules could be very different. Similarly, \textit{Spearman's rank correlation coefficient} results of ByteTrack system also suggest that there is no significant correlation between two modules OOD data ranking. Again, these results reveal that OOD sensitivity could be very different among modules even in an identical AI system.
}

\begin{table}[htbp]
    \scriptsize
    \centering
    \caption{Spearman's rank correlation coefficient of OOD score computed according two different modules in SpeechBrain system.}
    \begin{threeparttable}
    \resizebox{\linewidth}{!}{
    \begin{tabular}{lcccccccccccc}
    \toprule
    \multirow{2}{*}{\textbf{Corruption}} & \multirow{2}{*}{\textbf{Severity}} & \multicolumn{2}{c}{\textbf{Configuration 1}} && \multicolumn{2}{c}{\textbf{Configuration 2}} && \multicolumn{2}{c}{\textbf{Configuration 3}} && \multicolumn{2}{c}{\textbf{Configuration 4}}\\
    \cmidrule{3-4} \cmidrule{6-7} \cmidrule{9-10} \cmidrule{12-13}
    & & \multicolumn{1}{c}{Maha-d} & \multicolumn{1}{c}{KDE} && \multicolumn{1}{c}{Maha-d} & \multicolumn{1}{c}{KDE} && \multicolumn{1}{c}{Maha-d} & \multicolumn{1}{c}{KDE} && \multicolumn{1}{c}{Maha-d} & \multicolumn{1}{c}{KDE}\\
    \midrule
    \cellcolor{lightgray} & \cellcolor{lightgray}1  & \cellcolor{lightgray}$0.01$ & \cellcolor{lightgray}$0.20$ &\cellcolor{lightgray}& \cellcolor{lightgray}$0.33$ & \cellcolor{lightgray}$-0.02$ &\cellcolor{lightgray}& \cellcolor{lightgray}$0.03$ & \cellcolor{lightgray}$-0.36$ & \cellcolor{lightgray} & \cellcolor{lightgray}$-0.09$ & \cellcolor{lightgray}$-0.34$\\
    \cellcolor{lightgray} & \cellcolor{lightgray}2  & \cellcolor{lightgray}$0.01$ & \cellcolor{lightgray}$0.33$ &\cellcolor{lightgray}& \cellcolor{lightgray}$0.16$ & \cellcolor{lightgray}$-0.02$ &\cellcolor{lightgray}& \cellcolor{lightgray}$0.06$ & \cellcolor{lightgray}$-0.46$ & \cellcolor{lightgray} & \cellcolor{lightgray}$-0.17$ & \cellcolor{lightgray}$-0.39$\\
    \multirow{-3}{*}{\cellcolor{lightgray}Reverb (RE)} & \cellcolor{lightgray}3 & \cellcolor{lightgray}$<0.01$ & \cellcolor{lightgray}$0.34$ &\cellcolor{lightgray}& \cellcolor{lightgray}$0.03$ & \cellcolor{lightgray}$0.13$ &\cellcolor{lightgray}& \cellcolor{lightgray}$0.03$ & \cellcolor{lightgray}$-0.42$ & \cellcolor{lightgray} & \cellcolor{lightgray}$-0.18$ & \cellcolor{lightgray}$-0.47$\\
    \multirow{3}{*}{Speed (SP)}  & 1        & $-0.02$ & $-0.05$ & & $0.13$ & $0.10$ && $-0.09$ & $-0.24$ & & $0.01$ & $0.24$\\
    & 2        & $-0.02$ & $-0.02$ & & $0.55$ & $0.44$ && $0.03$ & $-0.41$ & & $-0.09$ & $-0.25$\\
    & 3        & $<0.01$ & $-0.01$ & & $0.15$ & $0.15$ && $0.05$ & $-0.42$ & & $-0.13$ & $-0.38$\\
    \cellcolor{lightgray} & \cellcolor{lightgray}1  & \cellcolor{lightgray}$-0.01$ & \cellcolor{lightgray}$-0.04$ &\cellcolor{lightgray}& \cellcolor{lightgray}$0.25$ & \cellcolor{lightgray}$0.13$ &\cellcolor{lightgray}& \cellcolor{lightgray}$-0.05$ & \cellcolor{lightgray}$-0.01$ & \cellcolor{lightgray} & \cellcolor{lightgray}$0.15$ & \cellcolor{lightgray}$-0.12$\\
    \cellcolor{lightgray} & \cellcolor{lightgray}2  & \cellcolor{lightgray}$-0.03$ & \cellcolor{lightgray}$0.05$ &\cellcolor{lightgray}& \cellcolor{lightgray}$0.30$ & \cellcolor{lightgray}$0.19$ &\cellcolor{lightgray}& \cellcolor{lightgray}$-0.11$ & \cellcolor{lightgray}$0.09$ & \cellcolor{lightgray} & \cellcolor{lightgray}$0.21$ & \cellcolor{lightgray}$-0.05$\\
    \multirow{-3}{*}{\cellcolor{lightgray}Distortion (DI)} & \cellcolor{lightgray}3 & \cellcolor{lightgray}$-0.04$ & \cellcolor{lightgray}$<0.01$ &\cellcolor{lightgray}& \cellcolor{lightgray}$0.22$ & \cellcolor{lightgray}$0.21$ &\cellcolor{lightgray}& \cellcolor{lightgray}$-0.22$ & \cellcolor{lightgray}$0.21$ & \cellcolor{lightgray} & \cellcolor{lightgray}$0.23$ & \cellcolor{lightgray}$-0.01$\\
    \multirow{3}{*}{Pitch shift (PS)}  & 1        & $-0.02$ & $-0.01$ & & $0.39$ & $0.20$ && $-0.02$ & $-0.30$ & & $0.01$ & $-0.22$\\
    & 2        & $-0.03$ & $0.01$ & & $0.49$ & $0.38$ && $-0.03$ & $-0.26$ & & $-0.02$ & $-0.24$\\
    & 3        & $0.02$ & $-0.03$ & & $0.56$ & $0.52$ && $-0.03$ & $-0.24$ & & $-0.04$ & $-0.23$\\
    \cellcolor{lightgray} & \cellcolor{lightgray}1  & \cellcolor{lightgray}$0.01$ & \cellcolor{lightgray}$-0.03$ &\cellcolor{lightgray}& \cellcolor{lightgray}$0.37$ & \cellcolor{lightgray}$0.57$ &\cellcolor{lightgray}& \cellcolor{lightgray}$-0.02$ & \cellcolor{lightgray}$-0.20$ & \cellcolor{lightgray} & \cellcolor{lightgray}$-0.06$ & \cellcolor{lightgray}$-0.25$\\
    \cellcolor{lightgray} & \cellcolor{lightgray}2  & \cellcolor{lightgray}$0.03$ & \cellcolor{lightgray}$-0.06$ &\cellcolor{lightgray}& \cellcolor{lightgray}$0.42$ & \cellcolor{lightgray}$0.49$ &\cellcolor{lightgray}& \cellcolor{lightgray}$-0.02$ & \cellcolor{lightgray}$-0.17$ & \cellcolor{lightgray} & \cellcolor{lightgray}$-0.09$ & \cellcolor{lightgray}$-0.22$\\
    \multirow{-3}{*}{\cellcolor{lightgray}Background noise (BN)} & \cellcolor{lightgray}3 & \cellcolor{lightgray}$0.01$ & \cellcolor{lightgray}$-0.04$ &\cellcolor{lightgray}& \cellcolor{lightgray}$0.38$ & \cellcolor{lightgray}$0.43$ &\cellcolor{lightgray}& \cellcolor{lightgray}$-0.01$ & \cellcolor{lightgray}$-0.20$ & \cellcolor{lightgray} & \cellcolor{lightgray}$-0.09$ & \cellcolor{lightgray}$-0.20$\\
    \bottomrule
    \end{tabular}
    }
    \end{threeparttable}
    \label{tab:speech_ood_spearman}
\end{table}

\begin{table}[htbp]
    \scriptsize
    \centering
    \caption{Spearman's rank correlation coefficient of OOD score computed according two different modules in ByteTrack system (YOLO+DeepSORT).}
    \begin{threeparttable}
    \begin{tabular}{lccc}
    \toprule
    \multirow{1}{*}{\textbf{Corruption}} & \multirow{1}{*}{\textbf{Severity}} & \multicolumn{1}{c}{Maha-d} & \multicolumn{1}{c}{KDE}\\
    \midrule
    \cellcolor{lightgray} & \cellcolor{lightgray}1  & \cellcolor{lightgray}$-0.07$ & \cellcolor{lightgray}$-0.20$ \\
    \cellcolor{lightgray} & \cellcolor{lightgray}2  & \cellcolor{lightgray}$0.39$ & \cellcolor{lightgray}$-0.44$ \\
    \multirow{-3}{*}{\cellcolor{lightgray}Gaussian noise (GN)} & \cellcolor{lightgray}3 & \cellcolor{lightgray}$0.39$ & \cellcolor{lightgray}$-0.14$ \\
    \multirow{3}{*}{Motion blur (MB)}  & 1        & $-0.05$ & $<0.01$ \\
    & 2        & $0.04$ & $-0.10$ \\
    & 3        & $0.13$ & $-0.13$ \\
    \cellcolor{lightgray} & \cellcolor{lightgray}1  & \cellcolor{lightgray}$-0.15$ & \cellcolor{lightgray}$<0.01$ \\
    \cellcolor{lightgray} & \cellcolor{lightgray}2  & \cellcolor{lightgray}$0.28$ & \cellcolor{lightgray}$-0.15$ \\
    \multirow{-3}{*}{\cellcolor{lightgray}Contrast (CO)} & \cellcolor{lightgray}3 & \cellcolor{lightgray}$0.31$ & \cellcolor{lightgray}$-0.02$ \\
    \multirow{3}{*}{Snow (SN)}  & 1        & $-0.02$ & $-0.02$ \\
    & 2        & $0.16$ & $0.02$ \\
    & 3        & $0.19$ & $0.06$ \\
    \cellcolor{lightgray} & \cellcolor{lightgray}1  & \cellcolor{lightgray}$-0.18$ & \cellcolor{lightgray}$0.01$ \\
    \cellcolor{lightgray} & \cellcolor{lightgray}2  & \cellcolor{lightgray}$-0.10$ & \cellcolor{lightgray}$0.14$ \\
    \multirow{-3}{*}{\cellcolor{lightgray}Pixelate (PI)} & \cellcolor{lightgray}3 & \cellcolor{lightgray}$0.29$ & \cellcolor{lightgray}$0.08$ \\
    \bottomrule
    \end{tabular}
    \end{threeparttable}
    \label{tab:tracking_ood_spearmanr}
\end{table}

\begin{tcolorbox}[size=title]
{
\textbf{Answer to RQ1:} Both studied OOD detection methods (\ie, Maha-d and KDE) can capture the data distribution difference between clean data and corrupted data, while there are no systematic evidences that one method is better than the other. These two methods are also more effective on the first module in a serial-executed system than later, indicated by Wilcoxon signed-rank test results. Furthermore, different modules of an AI system could detect different OOD data, suggesting that developers could take the OOD sensitivity of different modules into account when performing risk assessments.
}
\end{tcolorbox}

\subsection{RQ2. How does the uncertainty information behave on different modules in AI systems?}

In this research question, we utilize the corrupted datasets created in RQ1 again to evaluate the effectiveness of two uncertainty estimation metrics used in this study---variation ratio (VR) and variation ratio of original prediction (VRO). We first evaluate the uncertainty score's distribution differences between clean data and corrupted data in Sec.~\ref{sec:uncertainty_distribution}, then we investigate if there is any correlation between different modules' uncertainty information of an AI system in Sec.~\ref{sec:uncertainty_relationship}.

\subsubsection{Uncertainty estimation against common corruption patterns.}
\label{sec:uncertainty_distribution}

\begin{table}[htbp]
    \scriptsize
    \centering
    \caption{Wilcoxon signed-rank test on ByteTrack system's uncertainty score distribution difference between clean data and corrupted data on MOT17 dataset.}
    \begin{threeparttable}
    \begin{tabular}{llccccccccc}
    \toprule
    \multirow{2}{*}{\textbf{Corruption}} & \multirow{2}{*}{\textbf{Settings$^\diamond$}} & \multicolumn{4}{c}{${\mathcal{D}_\textbf{VR}}^\dagger$} && \multicolumn{4}{c}{${\mathcal{D}_\textbf{VRO}}^\dagger$}\\
    \cmidrule{3-6} \cmidrule{8-11}
    & & \multicolumn{1}{c}{YOLO} & \multicolumn{1}{c}{Track} & \multicolumn{1}{c}{SORT} & \multicolumn{1}{c}{DeepSORT} && \multicolumn{1}{c}{YOLO} & \multicolumn{1}{c}{Track} & \multicolumn{1}{c}{SORT} & \multicolumn{1}{c}{DeepSORT}\\
    \midrule
    \cellcolor{lightgray} & \cellcolor{lightgray}L1-Ori & \cellcolor{lightgray}$0.83$ & \cellcolor{lightgray}$0.85$ & \cellcolor{lightgray}$0.81$ & \cellcolor{lightgray}$0.68$ &\cellcolor{lightgray}& \cellcolor{lightgray}$0.82$ & \cellcolor{lightgray}$0.84$ & \cellcolor{lightgray}$0.82$ & \cellcolor{lightgray}$0.72$\\
    \cellcolor{lightgray} & \cellcolor{lightgray}L2-L1 & \cellcolor{lightgray}$0.89$ & \cellcolor{lightgray}$0.89$ & \cellcolor{lightgray}$0.92$ & \cellcolor{lightgray}$0.26^*$ &\cellcolor{lightgray}& \cellcolor{lightgray}$0.95$ & \cellcolor{lightgray}$0.93$ & \cellcolor{lightgray}$0.95$ & \cellcolor{lightgray}$0.71$\\
    \multirow{-3}{*}{\cellcolor{lightgray}Gaussian noise (GN)} & \cellcolor{lightgray}L3-L2 & \cellcolor{lightgray}$0.65$ & \cellcolor{lightgray}$0.39^*$ & \cellcolor{lightgray}$0.31^*$ & \cellcolor{lightgray}$0.02^*$ &\cellcolor{lightgray}& \cellcolor{lightgray}$0.97$ & \cellcolor{lightgray}$0.82$ & \cellcolor{lightgray}$0.77$ & \cellcolor{lightgray}$0.63$\\
    \multirow{3}{*}{Motion blur (MB)} & L1-Ori &$0.41^*$ & $0.69$ & $0.69$ & $0.55$ && $0.47^*$ & $0.67$ & $0.69$ & $0.62$\\
    & L2-L1 & $0.67$ & $0.88$ & $0.86$ & $0.62$ && $0.70$ & $0.87$ & $0.87$ & $0.71$\\
    & L3-L2 & $0.78$ & $0.86$ & $0.85$ & $0.48^*$ && $0.82$ & $0.87$ & $0.86$ & $0.64$\\
    \cellcolor{lightgray} & \cellcolor{lightgray}L1-Ori & \cellcolor{lightgray}$0.60$ & \cellcolor{lightgray}$0.66$ & \cellcolor{lightgray}$0.67$ & \cellcolor{lightgray}$0.46^*$ &\cellcolor{lightgray}& \cellcolor{lightgray}$0.67$ & \cellcolor{lightgray}$0.69$ & \cellcolor{lightgray}$0.75$ & \cellcolor{lightgray}$0.61$\\
    \cellcolor{lightgray} & \cellcolor{lightgray}L2-L1 & \cellcolor{lightgray}$0.68$ & \cellcolor{lightgray}$0.81$ & \cellcolor{lightgray}$0.84$ & \cellcolor{lightgray}$0.36^*$ &\cellcolor{lightgray}& \cellcolor{lightgray}$0.75$ & \cellcolor{lightgray}$0.86$ & \cellcolor{lightgray}$0.88$ & \cellcolor{lightgray}$0.63$\\
    \multirow{-3}{*}{\cellcolor{lightgray}Contrast (CO)} & \cellcolor{lightgray}L3-L2 & \cellcolor{lightgray}$0.49^*$ & \cellcolor{lightgray}$0.10^*$ & \cellcolor{lightgray}$0.07^*$ & \cellcolor{lightgray}$0.01^*$ &\cellcolor{lightgray}& \cellcolor{lightgray}$0.96$ & \cellcolor{lightgray}$0.63$ & \cellcolor{lightgray}$0.59$ & \cellcolor{lightgray}$0.59$\\
    \multirow{3}{*}{Snow (SN)} & L1-Ori & $0.86$ & $0.82$ & $0.83$ & $0.44^*$ && $0.91$ & $0.84$ & $0.85$ & $0.66$\\
    & L2-L1 & $0.82$ & $0.80$ & $0.79$ & $0.39^*$ && $0.80$ & $0.79$ & $0.79$ & $0.56$\\
    & L3-L2 & $0.62$ & $0.66$ & $0.65$ & $0.36^*$ && $0.67$ & $0.69$ & $0.67$ & $0.52^*$\\
    \cellcolor{lightgray} & \cellcolor{lightgray}L1-Ori & \cellcolor{lightgray}$0.62$ & \cellcolor{lightgray}$0.67$ & \cellcolor{lightgray}$0.64$ & \cellcolor{lightgray}$0.60$ &\cellcolor{lightgray}& \cellcolor{lightgray}$0.68$ & \cellcolor{lightgray}$0.69$ & \cellcolor{lightgray}$0.71$ & \cellcolor{lightgray}$0.62$\\
    \cellcolor{lightgray} & \cellcolor{lightgray}L2-L1 & \cellcolor{lightgray}$0.62$ & \cellcolor{lightgray}$0.72$ & \cellcolor{lightgray}$0.72$ & \cellcolor{lightgray}$0.58$ &\cellcolor{lightgray}& \cellcolor{lightgray}$0.67$ & \cellcolor{lightgray}$0.73$ & \cellcolor{lightgray}$0.73$ & \cellcolor{lightgray}$0.63$\\
    \multirow{-3}{*}{\cellcolor{lightgray}Pixelate (PI)} & \cellcolor{lightgray}L3-L2 & \cellcolor{lightgray}$0.72$ & \cellcolor{lightgray}$0.82$ & \cellcolor{lightgray}$0.79$ & \cellcolor{lightgray}$0.62$ &\cellcolor{lightgray}& \cellcolor{lightgray}$0.74$ & \cellcolor{lightgray}$0.79$ & \cellcolor{lightgray}$0.78$ & \cellcolor{lightgray}$0.67$\\
    \bottomrule
    \end{tabular}
    \begin{tablenotes}
        \item[$\diamond$] A setting of ``A-B'' indicates that the test is conducted between uncertainty score on A and B, \eg, ``L1-Ori'' means the test is conducted between level 1 corrupted data and clean data.
        \item[$\dagger$] An effect size $\mathcal{D}$ indicates the probability that uncertainty score on corrupted data is larger than it on clean data; larger the effect size, more significant the result is.
        \item[$*$] This result is not statistically significant ($p>0.0001$).
    \end{tablenotes}
    \end{threeparttable}
    \label{tab:tracking_uncertainty}
\end{table}

\begin{table}[htbp]
    \scriptsize
    \centering
    \caption{Wilcoxon signed-rank test on ConvLab-2 system's uncertainty score distribution difference between clean data and corrupted data on MultiWOZ dataset.}
    \begin{threeparttable}
    \begin{tabular}{llccccc}
    \toprule
    \multirow{2}{*}{\textbf{Corruption}} & \multirow{2}{*}{\textbf{Settings$^\diamond$}} & \multicolumn{2}{c}{${\mathcal{D}_\textbf{VR}}^\dagger$} && \multicolumn{2}{c}{${\mathcal{D}_\textbf{VRO}}^\dagger$}\\
    \cmidrule{3-4} \cmidrule{6-7}
    & & \multicolumn{1}{c}{BERT NLU} & \multicolumn{1}{c}{MILU NLU} && \multicolumn{1}{c}{BERT NLU} & \multicolumn{1}{c}{MILU NLU}\\
    \midrule
    \cellcolor{lightgray} & \cellcolor{lightgray}L1-Ori & \cellcolor{lightgray}$0.06*$ & \cellcolor{lightgray}$0.32$ &\cellcolor{lightgray}& \cellcolor{lightgray}$0.07*$ & \cellcolor{lightgray}$0.29$\\
    \cellcolor{lightgray} & \cellcolor{lightgray}L2-L1 & \cellcolor{lightgray}$0.08$ & \cellcolor{lightgray}$0.12*$ &\cellcolor{lightgray}& \cellcolor{lightgray}$0.08$ & \cellcolor{lightgray}$0.12*$\\
    \multirow{-3}{*}{\cellcolor{lightgray}Digits2words (DW)} & \cellcolor{lightgray}L3-L2 & \cellcolor{lightgray}$0.12$ & \cellcolor{lightgray}$0.13*$ &\cellcolor{lightgray}& \cellcolor{lightgray}$0.12$ & \cellcolor{lightgray}$0.12*$\\
    \multirow{3}{*}{Change char (CC)} & L1-Ori &$0.75$ & $0.78$ && $0.73$ & $0.72$\\
    & L2-L1 & $0.95$ & $0.89$ && $0.94$ & $0.83$\\
    & L3-L2 & $0.96$ & $0.95$ && $0.95$ & $0.91$\\
    \cellcolor{lightgray} & \cellcolor{lightgray}L1-Ori & \cellcolor{lightgray}$0.84$ & \cellcolor{lightgray}$0.84$ &\cellcolor{lightgray}& \cellcolor{lightgray}$0.81$ & \cellcolor{lightgray}$0.75$\\
    \cellcolor{lightgray} & \cellcolor{lightgray}L2-L1 & \cellcolor{lightgray}$0.98$ & \cellcolor{lightgray}$0.95$ &\cellcolor{lightgray}& \cellcolor{lightgray}$0.98$ & \cellcolor{lightgray}$0.92$\\
    \multirow{-3}{*}{\cellcolor{lightgray}Remove char (RC)} & \cellcolor{lightgray}L3-L2 & \cellcolor{lightgray}$0.99$ & \cellcolor{lightgray}$0.99$ &\cellcolor{lightgray}& \cellcolor{lightgray}$0.99$ & \cellcolor{lightgray}$0.98$\\
    \multirow{3}{*}{Misspelling (MS)} & L1-Ori & $0.37$ & $0.54$ && $0.37$ & $0.49$\\
    & L2-L1 & $0.65$ & $0.55$ && $0.62$ & $0.53$\\
    & L3-L2 & $0.80$ & $0.67$ && $0.78$ & $0.65$\\
    \cellcolor{lightgray} & \cellcolor{lightgray}L1-Ori & \cellcolor{lightgray}$0.92$ & \cellcolor{lightgray}$0.94$ &\cellcolor{lightgray}& \cellcolor{lightgray}$0.90$ & \cellcolor{lightgray}$0.89$\\
    \cellcolor{lightgray} & \cellcolor{lightgray}L2-L1 & \cellcolor{lightgray}$1.00$ & \cellcolor{lightgray}$1.00$ &\cellcolor{lightgray}& \cellcolor{lightgray}$1.00$ & \cellcolor{lightgray}$1.00$\\
    \multirow{-3}{*}{\cellcolor{lightgray}Swap (SW)} & \cellcolor{lightgray}L3-L2 & \cellcolor{lightgray}$0.99$ & \cellcolor{lightgray}$0.99$ &\cellcolor{lightgray}& \cellcolor{lightgray}$0.94$ & \cellcolor{lightgray}$0.97$\\
    \bottomrule
    \end{tabular}
    \begin{tablenotes}
        \item[$\diamond$] A setting of ``A-B'' indicates that the test is conducted between uncertainty score on A and B, \eg, ``L1-Ori'' means the test is conducted between level 1 corrupted data and clean data.
        \item[$\dagger$] An effect size $\mathcal{D}$ indicates the probability that uncertainty score on corrupted data is larger than it on clean data; larger the effect size, more significant the result is.
        \item[$*$] This result is not statistically significant ($p>0.0001$).
    \end{tablenotes}
    \end{threeparttable}
    \label{tab:convlab_uncertainty}
\end{table}

\begin{table}[htbp]
    \scriptsize
    \centering
    \caption{Wilcoxon signed-rank test on SpeechBrain system's uncertainty score distribution difference between clean data and corrupted data on LibriSpeech dataset.}
    \resizebox{\linewidth}{!}{
    \begin{threeparttable}
    \begin{tabular}{llccccccccccc}
    \toprule
    \multirow{3}{*}{\textbf{Corruption}} & \multirow{3}{*}{\textbf{Settings$^\diamond$}} & \multicolumn{5}{c}{\textbf{Configuration 1}} & & \multicolumn{5}{c}{\textbf{Configuration 2}}\\
    & & \multicolumn{2}{c}{${\mathcal{D}_{VR}}^\dagger$} && \multicolumn{2}{c}{${\mathcal{D}_{VRO}}^\dagger$} && \multicolumn{2}{c}{${\mathcal{D}_{VR}}^\dagger$} && \multicolumn{2}{c}{${\mathcal{D}_{VRO}}^\dagger$}\\
    \cmidrule{3-4} \cmidrule{6-7} \cmidrule{9-10} \cmidrule{12-13}
    & & \multicolumn{1}{c}{CRDNN} & \multicolumn{1}{c}{RNN} && \multicolumn{1}{c}{CRDNN} & \multicolumn{1}{c}{RNN} & & \multicolumn{1}{c}{CRDNN} & \multicolumn{1}{c}{Trans-} && \multicolumn{1}{c}{CRDNN} & \multicolumn{1}{c}{Trans-}\\
    \midrule
    \cellcolor{lightgray} & \cellcolor{lightgray}L1-Ori & \cellcolor{lightgray}$0.65$ & \cellcolor{lightgray}$0.58$ &\cellcolor{lightgray}& \cellcolor{lightgray}$0.65$ & \cellcolor{lightgray}$0.56$ &\cellcolor{lightgray}& \cellcolor{lightgray}$0.72$ & \cellcolor{lightgray}$0.97$ &\cellcolor{lightgray}& \cellcolor{lightgray}$0.76$ & \cellcolor{lightgray}$0.98$\\
    \cellcolor{lightgray} & \cellcolor{lightgray}L2-L1 & \cellcolor{lightgray}$0.83$ & \cellcolor{lightgray}$0.94$ &\cellcolor{lightgray}& \cellcolor{lightgray}$0.83$ & \cellcolor{lightgray}$0.91$ &\cellcolor{lightgray}& \cellcolor{lightgray}$0.93$ & \cellcolor{lightgray}$0.99$ &\cellcolor{lightgray}& \cellcolor{lightgray}$0.95$ & \cellcolor{lightgray}$0.99$\\
    \multirow{-3}{*}{\cellcolor{lightgray}Reverb (RE)} & \cellcolor{lightgray}L3-L2 & \cellcolor{lightgray}$0.66$ & \cellcolor{lightgray}$0.97$ &\cellcolor{lightgray}& \cellcolor{lightgray}$0.65$ & \cellcolor{lightgray}$0.95$ &\cellcolor{lightgray}& \cellcolor{lightgray}$0.70$ & \cellcolor{lightgray}$0.91$ &\cellcolor{lightgray}& \cellcolor{lightgray}$0.72$ & \cellcolor{lightgray}$0.89$\\
    \multirow{3}{*}{Speed (SP)} & L1-Ori &$0.65$ & $0.47$ && $0.64$ & $0.44$ && $0.58$ & $0.64$ && $0.60$ & $0.68$\\
    & L2-L1 & $0.94$ & $0.93$ && $0.94$ & $0.91$ && $0.95$ & $0.98$ && $0.96$ & $0.98$\\
    & L3-L2 & $0.90$ & $0.98$ && $0.89$ & $0.96$ && $0.99$ & $0.97$ && $0.99$ & $0.96$\\
    \cellcolor{lightgray} & \cellcolor{lightgray}L1-Ori & \cellcolor{lightgray}$0.62$ & \cellcolor{lightgray}$0.45$ &\cellcolor{lightgray}& \cellcolor{lightgray}$0.62$ & \cellcolor{lightgray}$0.43^*$ &\cellcolor{lightgray}& \cellcolor{lightgray}$0.50^*$ & \cellcolor{lightgray}$0.60$ &\cellcolor{lightgray}& \cellcolor{lightgray}$0.51^*$ & \cellcolor{lightgray}$0.63$\\
    \cellcolor{lightgray} & \cellcolor{lightgray}L2-L1 & \cellcolor{lightgray}$0.60$ & \cellcolor{lightgray}$0.54$ &\cellcolor{lightgray}& \cellcolor{lightgray}$0.61$ & \cellcolor{lightgray}$0.53$ &\cellcolor{lightgray}& \cellcolor{lightgray}$0.51^*$ & \cellcolor{lightgray}$0.70$ &\cellcolor{lightgray}& \cellcolor{lightgray}$0.52^*$ & \cellcolor{lightgray}$0.75$\\
    \multirow{-3}{*}{\cellcolor{lightgray}Distortion (DI)} & \cellcolor{lightgray}L3-L2 & \cellcolor{lightgray}$0.79$ & \cellcolor{lightgray}$0.88$ &\cellcolor{lightgray}& \cellcolor{lightgray}$0.78$ & \cellcolor{lightgray}$0.84$ &\cellcolor{lightgray}& \cellcolor{lightgray}$0.72$ & \cellcolor{lightgray}$0.96$ &\cellcolor{lightgray}& \cellcolor{lightgray}$0.75$ & \cellcolor{lightgray}$0.97$\\
    \multirow{3}{*}{Pitch shift (PS)} & L1-Ori & $0.79$ & $0.42^*$ && $0.81$ & $0.40^*$ && $0.54$ & $0.68$ && $0.56$ & $0.72$\\
    & L2-L1 & $0.66$ & $0.55$ && $0.66$ & $0.52$ && $0.63$ & $0.75$ && $0.65$ & $0.78$\\
    & L3-L2 & $0.77$ & $0.74$ && $0.77$ & $0.72$ && $0.70$ & $0.91$ && $0.73$ & $0.92$\\
    \cellcolor{lightgray} & \cellcolor{lightgray}L1-Ori & \cellcolor{lightgray}$0.63$ & \cellcolor{lightgray}$0.60$ &\cellcolor{lightgray}& \cellcolor{lightgray}$0.63$ & \cellcolor{lightgray}$0.56$ &\cellcolor{lightgray}& \cellcolor{lightgray}$0.93$ & \cellcolor{lightgray}$0.99$ &\cellcolor{lightgray}& \cellcolor{lightgray}$0.95$ & \cellcolor{lightgray}$0.99$\\
    \cellcolor{lightgray} & \cellcolor{lightgray}L2-L1 & \cellcolor{lightgray}$0.61$ & \cellcolor{lightgray}$0.71$ &\cellcolor{lightgray}& \cellcolor{lightgray}$0.61$ & \cellcolor{lightgray}$0.67$ &\cellcolor{lightgray}& \cellcolor{lightgray}$0.84$ & \cellcolor{lightgray}$0.95$ &\cellcolor{lightgray}& \cellcolor{lightgray}$0.85$ & \cellcolor{lightgray}$0.96$\\
    \multirow{-3}{*}{\cellcolor{lightgray}Background noise (BN)} & \cellcolor{lightgray}L3-L2 & \cellcolor{lightgray}$0.54$ & \cellcolor{lightgray}$0.75$ &\cellcolor{lightgray}& \cellcolor{lightgray}$0.54$ & \cellcolor{lightgray}$0.71$ &\cellcolor{lightgray}& \cellcolor{lightgray}$0.60$ & \cellcolor{lightgray}$0.81$ &\cellcolor{lightgray}& \cellcolor{lightgray}$0.60$ & \cellcolor{lightgray}$0.77$\\
    \bottomrule
    \end{tabular}
    \begin{tablenotes}
        \item[$\diamond$] A setting of ``A-B'' indicates that the test is conducted between uncertainty score on A and B, \eg, ``L1-Ori'' means the test is conducted between level 1 corrupted data and clean data.
        \item[$\dagger$] An effect size $\mathcal{D}$ indicates the probability that uncertainty score on corrupted data is larger than it on clean data; larger the effect size, more significant the result is.
        \item[$*$] This result is not statistically significant ($p>0.0001$).
    \end{tablenotes}
    \end{threeparttable}
    }
    \label{tab:speech_uncertainty1}
\end{table}

\begin{table}[htbp]
    \scriptsize
    \centering
    \caption{Wilcoxon signed-rank test on SpeechBrain system's uncertainty score distribution difference between clean data and corrupted data on CommonVoice-it dataset.}
    \resizebox{\linewidth}{!}{
    \begin{threeparttable}
    \begin{tabular}{llccccccccccc}
    \toprule
    \multirow{3}{*}{\textbf{Corruption}} & \multirow{3}{*}{\textbf{Settings$^\diamond$}} & \multicolumn{5}{c}{\textbf{Configuration 3}} & & \multicolumn{5}{c}{\textbf{Configuration 4}}\\
    & & \multicolumn{2}{c}{${\mathcal{D}_{VR}}^\dagger$} && \multicolumn{2}{c}{${\mathcal{D}_{VRO}}^\dagger$} && \multicolumn{2}{c}{${\mathcal{D}_{VR}}^\dagger$} && \multicolumn{2}{c}{${\mathcal{D}_{VRO}}^\dagger$}\\
    \cmidrule{3-4} \cmidrule{6-7} \cmidrule{9-10} \cmidrule{12-13}
    & & \multicolumn{1}{c}{CRDNN} & \multicolumn{1}{c}{RNN} && \multicolumn{1}{c}{CRDNN} & \multicolumn{1}{c}{RNN} & & \multicolumn{1}{c}{Wav2Vec2} & \multicolumn{1}{c}{RNN} && \multicolumn{1}{c}{Wav2Vec2} & \multicolumn{1}{c}{RNN}\\
    \midrule
    \cellcolor{lightgray} & \cellcolor{lightgray}L1-Ori & \cellcolor{lightgray}$0.69$ & \cellcolor{lightgray}$0.76$ &\cellcolor{lightgray}& \cellcolor{lightgray}$0.72$ & \cellcolor{lightgray}$0.73$ &\cellcolor{lightgray}& \cellcolor{lightgray}$1.00$ & \cellcolor{lightgray}$0.64$ &\cellcolor{lightgray}& \cellcolor{lightgray}$1.00$ & \cellcolor{lightgray}$0.63$\\
    \cellcolor{lightgray} & \cellcolor{lightgray}L2-L1 & \cellcolor{lightgray}$0.95$ & \cellcolor{lightgray}$0.96$ &\cellcolor{lightgray}& \cellcolor{lightgray}$0.96$ & \cellcolor{lightgray}$0.95$ &\cellcolor{lightgray}& \cellcolor{lightgray}$1.00$ & \cellcolor{lightgray}$0.96$ &\cellcolor{lightgray}& \cellcolor{lightgray}$1.00$ & \cellcolor{lightgray}$0.95$\\
    \multirow{-3}{*}{\cellcolor{lightgray}Reverb (RE)} & \cellcolor{lightgray}L3-L2 & \cellcolor{lightgray}$0.85$ & \cellcolor{lightgray}$0.91$ &\cellcolor{lightgray}& \cellcolor{lightgray}$0.86$ & \cellcolor{lightgray}$0.88$ &\cellcolor{lightgray}& \cellcolor{lightgray}$0.97$ & \cellcolor{lightgray}$0.97$ &\cellcolor{lightgray}& \cellcolor{lightgray}$0.93$ & \cellcolor{lightgray}$0.96$\\
    \multirow{3}{*}{Speed (SP)} & L1-Ori &$0.92$ & $0.82$ && $0.93$ & $0.80$ && $0.95$ & $0.45$ && $0.96$ & $0.44$\\
    & L2-L1 & $0.73$ & $0.88$ && $0.74$ & $0.85$ && $0.95$ & $0.77$ && $0.93$ & $0.75$\\
    & L3-L2 & $0.98$ & $0.95$ && $0.98$ & $0.93$ && $1.00$ & $1.00$ && $0.99$ & $1.00$\\
    \cellcolor{lightgray} & \cellcolor{lightgray}L1-Ori & \cellcolor{lightgray}$0.70$ & \cellcolor{lightgray}$0.57$ &\cellcolor{lightgray}& \cellcolor{lightgray}$0.71$ & \cellcolor{lightgray}$0.55$ &\cellcolor{lightgray}& \cellcolor{lightgray}$0.88$ & \cellcolor{lightgray}$0.38$ &\cellcolor{lightgray}& \cellcolor{lightgray}$0.88$ & \cellcolor{lightgray}$0.37$\\
    \cellcolor{lightgray} & \cellcolor{lightgray}L2-L1 & \cellcolor{lightgray}$0.66$ & \cellcolor{lightgray}$0.61$ &\cellcolor{lightgray}& \cellcolor{lightgray}$0.67$ & \cellcolor{lightgray}$0.58$ &\cellcolor{lightgray}& \cellcolor{lightgray}$0.93$ & \cellcolor{lightgray}$0.50$ &\cellcolor{lightgray}& \cellcolor{lightgray}$0.92$ & \cellcolor{lightgray}$0.48$\\
    \multirow{-3}{*}{\cellcolor{lightgray}Distortion (DI)} & \cellcolor{lightgray}L3-L2 & \cellcolor{lightgray}$0.79$ & \cellcolor{lightgray}$0.84$ &\cellcolor{lightgray}& \cellcolor{lightgray}$0.81$ & \cellcolor{lightgray}$0.81$ &\cellcolor{lightgray}& \cellcolor{lightgray}$0.99$ & \cellcolor{lightgray}$0.82$ &\cellcolor{lightgray}& \cellcolor{lightgray}$0.99$ & \cellcolor{lightgray}$0.81$\\
    \multirow{3}{*}{Pitch shift (PS)} & L1-Ori & $0.62$ & $0.56$ && $0.62$ & $0.54$ && $0.97$ & $0.39$ && $0.96$ & $0.37$\\
    & L2-L1 & $0.74$ & $0.65$ && $0.75$ & $0.63$ && $0.92$ & $0.47$ && $0.92$ & $0.45$\\
    & L3-L2 & $0.81$ & $0.81$ && $0.82$ & $0.78$ && $0.98$ & $0.72$ && $0.97$ & $0.70$\\
    \cellcolor{lightgray} & \cellcolor{lightgray}L1-Ori & \cellcolor{lightgray}$0.93$ & \cellcolor{lightgray}$0.75$ &\cellcolor{lightgray}& \cellcolor{lightgray}$0.93$ & \cellcolor{lightgray}$0.73$ &\cellcolor{lightgray}& \cellcolor{lightgray}$1.00$ & \cellcolor{lightgray}$0.60$ &\cellcolor{lightgray}& \cellcolor{lightgray}$1.00$ & \cellcolor{lightgray}$0.59$\\
    \cellcolor{lightgray} & \cellcolor{lightgray}L2-L1 & \cellcolor{lightgray}$0.79$ & \cellcolor{lightgray}$0.78$ &\cellcolor{lightgray}& \cellcolor{lightgray}$0.80$ & \cellcolor{lightgray}$0.75$ &\cellcolor{lightgray}& \cellcolor{lightgray}$1.00$ & \cellcolor{lightgray}$0.75$ &\cellcolor{lightgray}& \cellcolor{lightgray}$0.99$ & \cellcolor{lightgray}$0.74$\\
    \multirow{-3}{*}{\cellcolor{lightgray}Background noise (BN)} & \cellcolor{lightgray}L3-L2 & \cellcolor{lightgray}$0.74$ & \cellcolor{lightgray}$0.79$ &\cellcolor{lightgray}& \cellcolor{lightgray}$0.75$ & \cellcolor{lightgray}$0.75$ &\cellcolor{lightgray}& \cellcolor{lightgray}$0.99$ & \cellcolor{lightgray}$0.84$ &\cellcolor{lightgray}& \cellcolor{lightgray}$0.98$ & \cellcolor{lightgray}$0.82$\\
    \bottomrule
    \end{tabular}
    \begin{tablenotes}
        \item[$\diamond$] A setting of ``A-B'' indicates that the test is conducted between uncertainty score on A and B, \eg, ``L1-Ori'' means the test is conducted between level 1 corrupted data and clean data.
        \item[$\dagger$] An effect size $\mathcal{D}$ indicates the probability that uncertainty score on corrupted data is larger than it on clean data; larger the effect size, more significant the result is.
        \item[$*$] This result is not statistically significant ($p>0.0001$).
    \end{tablenotes}
    \end{threeparttable}
    }
    \label{tab:speech_uncertainty2}
\end{table}

Table~\ref{tab:tracking_uncertainty}, Table~\ref{tab:convlab_uncertainty}, and Table~\ref{tab:speech_uncertainty1}\&\ref{tab:speech_uncertainty2} show the \textit{Wilcoxon signed-rank test} results on differences between uncertainty score of clean data and corrupted data on ByteTrack system, ConvLab-2 system, and SpeechBrain system, respectively. In these tables, an effect size larger than 0.5 indicates that this module's uncertainty score on corrupted data is significantly larger than it on clean data. Larger the value, the more significant the result is. In the ConvLab-2 system (Table~\ref{tab:convlab_uncertainty}) and SpeechBrain system (Table~\ref{tab:speech_uncertainty1}\&\ref{tab:speech_uncertainty2}), we can find that the uncertainty score measured by either VR or VRO on corrupted data is usually larger than it on clean data, demonstrating that uncertainty estimation techniques can capture the system's prediction difference on clean data and corrupted data. Moreover, we also find that in the ConvLab-2 and SpeechBrain systems, a larger severity level of corruption leads to a larger effect size, showing that the system's predictions on high-severity corrupted data are more uncertain than with predictions on clean data.

There are also some exceptions in the results. For instance, we find that in the ByteTrack system, sometimes the uncertainty score measured by VR on corrupted data is not larger than it on clean data (\eg, DeepSORT module against Gaussian noise pattern). Further investigation results suggest that this is because the tracking module's robustness is significantly affected by these common corruption patterns. Therefore, a lot of prediction results (\ie, tracking results) from the tracking module are empty tracklets (``nothing tracked''). Recall that the calculation of VR is based on the major prediction results after applying Monte Carlo dropout. When there are a lot of empty tracklets, it's not surprising that VR on corrupted data is even lower, especially on high severity corrupted data. While in such cases, VRO performs much better than VR. In Table~\ref{tab:tracking_uncertainty}, the effect size suggests that the uncertainty score measured by VRO on corrupted data is significantly larger than it on clean data. We also find that in the ConvLab-2 system, both VR and VRO work only effectively on the NLU module. However, it's hard for them to measure the uncertainty on the other modules, especially DST and NLG.

\subsubsection{Correlations of different modules' uncertainty change}
\label{sec:uncertainty_relationship}

\begin{table}[htbp]
    \scriptsize
    \centering
    \caption{Kendall's $\tau$ test on correlation between YOLO detector's uncertainty score change and different tracker's uncertainty score change in ByteTrack system.}
    \begin{threeparttable}
    \begin{tabular}{lcccccccccc}
    \toprule
    \multirow{2}{*}{\textbf{Corruption}} & \multirow{2}{*}{\textbf{Severity}} & \multicolumn{3}{c}{\textbf{Kendall's} $\tau$\textbf{-statistic of VR}$^\dagger$} && \multicolumn{3}{c}{\textbf{Kendall's} $\tau$\textbf{-statistic of VRO}$^\dagger$}\\
    \cmidrule{3-5} \cmidrule{7-9}
    & & \multicolumn{1}{c}{Track} & \multicolumn{1}{c}{SORT} & \multicolumn{1}{c}{DeepSORT} && \multicolumn{1}{c}{Track} & \multicolumn{1}{c}{SORT} & \multicolumn{1}{c}{DeepSORT}\\
    \midrule
    \cellcolor{lightgray} & \cellcolor{lightgray}1  & \cellcolor{lightgray}$0.139$ &\cellcolor{lightgray}$0.141$ &\cellcolor{lightgray}$-0.01^*$ &\cellcolor{lightgray}& \cellcolor{lightgray}$0.172$ & \cellcolor{lightgray}$0.143$ &\cellcolor{lightgray}$0.025^*$\\
    \cellcolor{lightgray} & \cellcolor{lightgray}2  & \cellcolor{lightgray}$0.236$ &\cellcolor{lightgray}$0.213$ &\cellcolor{lightgray}$-0.08$ &\cellcolor{lightgray}& \cellcolor{lightgray}$0.243$ & \cellcolor{lightgray}$0.256$ &\cellcolor{lightgray}$0.015^*$ \\
    \multirow{-3}{*}{\cellcolor{lightgray}Gaussian noise (GN)} &\cellcolor{lightgray}3 & \cellcolor{lightgray}$0.302$ &\cellcolor{lightgray}$0.174$ &\cellcolor{lightgray}$0.07$ &\cellcolor{lightgray}& \cellcolor{lightgray}$0.454$ & \cellcolor{lightgray}$0.358$ &\cellcolor{lightgray}$0.127$\\
    \multirow{3}{*}{Motion blur (MB)}  & 1        & $0.058$ & $0.036$ & $-0.03^*$ & & $0.058$ & $0.042$ & $-0.02^*$ \\
    & 2        & $0.113$ & $0.101$ & $0.01^*$ & & $0.144$ & $0.141$ & $0.04^*$ \\
    & 3        & $0.070$ & $0.093$ & $-0.186$ & & $0.150$ & $0.153$ & $-0.04$ \\
    \cellcolor{lightgray} & \cellcolor{lightgray}1  & \cellcolor{lightgray}$0.089$ &\cellcolor{lightgray}$0.103$ &\cellcolor{lightgray}$-0.03^*$ &\cellcolor{lightgray}& \cellcolor{lightgray}$0.112$ & \cellcolor{lightgray}$0.083$ &\cellcolor{lightgray}$0.04^*$\\
    \cellcolor{lightgray} & \cellcolor{lightgray}2  & \cellcolor{lightgray}$0.187$ &\cellcolor{lightgray}$0.167$ &\cellcolor{lightgray}$-0.097$ &\cellcolor{lightgray} & \cellcolor{lightgray}$0.211$ & \cellcolor{lightgray}$0.204$ &\cellcolor{lightgray}$-0.02$ \\
    \multirow{-3}{*}{\cellcolor{lightgray}Contrast (CO)} & \cellcolor{lightgray}3 & \cellcolor{lightgray}$0.478$ &\cellcolor{lightgray}$0.382$ &\cellcolor{lightgray}$0.397$ &\cellcolor{lightgray} & \cellcolor{lightgray}$0.381$ & \cellcolor{lightgray}$0.323$ &\cellcolor{lightgray}$0.314$\\
    \multirow{3}{*}{Snow (SN)}  & 1        & $0.134$ & $0.131$ & $-0.039^*$ & & $0.218$ & $0.231$ & $0.041$ \\
    & 2        & $0.237$ & $0.188$ & $-0.144$ & & $0.288$ & $0.268$ & $0.025$ \\
    & 3        & $0.168$ & $0.115$ & $-0.291$ & & $0.276$ & $0.233$ & $-0.084$ \\
    \cellcolor{lightgray} & \cellcolor{lightgray}1  & \cellcolor{lightgray}$0.038$ &\cellcolor{lightgray}$0.055$ &\cellcolor{lightgray}$-0.013^*$ &\cellcolor{lightgray}& \cellcolor{lightgray}$0.064$ & \cellcolor{lightgray}$0.062$ &\cellcolor{lightgray}$0.025^*$\\
    \cellcolor{lightgray} & \cellcolor{lightgray}2  & \cellcolor{lightgray}$0.116$ &\cellcolor{lightgray}$0.086$ &\cellcolor{lightgray}$0.006^*$ &\cellcolor{lightgray}& \cellcolor{lightgray}$0.184$ & \cellcolor{lightgray}$0.160$ &\cellcolor{lightgray}$0.047$ \\
    \multirow{-3}{*}{\cellcolor{lightgray}Pixelate (PI)} & \cellcolor{lightgray}3 & \cellcolor{lightgray}$0.102$ &\cellcolor{lightgray}$0.087$ &\cellcolor{lightgray}$-0.090$ &\cellcolor{lightgray} & \cellcolor{lightgray}$0.162$ & \cellcolor{lightgray}$0.143$ &\cellcolor{lightgray}$0.043$\\
    \bottomrule
    \end{tabular}
    \begin{tablenotes}
        \item[$\dagger$] A positive Kendall's $\tau$-statistic means when detector's uncertainty score increases on corrupted data comparing with it on clean data, then tracker's uncertainty score also increases. A larger $\tau$-statistic suggests more significant correlation.
        \item[$*$] This result is not statistically significant ($p > 0.0001$).
    \end{tablenotes}
    \end{threeparttable}
    \label{tab:tracking_uncertainty_rel}
\end{table}

\begin{table}[htbp]
    \scriptsize
    \centering
    \caption{Kendall's $\tau$ test on correlation between two module's (module 1---module 2) uncertainty score change in SpeechBrain system.}
    \resizebox{\linewidth}{!}{
    \begin{threeparttable}
    \begin{tabular}{lcccccccccccc}
    \toprule
    \multirow{2}{*}{\textbf{Corruption}} & \multirow{2}{*}{\textbf{Severity}} & \multicolumn{4}{c}{\textbf{Kendall's} $\tau$\textbf{-statistic of VR}$^\dagger$} && \multicolumn{4}{c}{\textbf{Kendall's} $\tau$\textbf{-statistic of VRO}$^\dagger$}\\
    \cmidrule{3-6} \cmidrule{8-11}
    & & \multicolumn{1}{c}{Config-1} & \multicolumn{1}{c}{Config-2} & \multicolumn{1}{c}{Config-3} & \multicolumn{1}{c}{Config-4} && \multicolumn{1}{c}{Config-1} & \multicolumn{1}{c}{Config-2} & \multicolumn{1}{c}{Config-3} & \multicolumn{1}{c}{Config-4}\\
    \midrule
    \cellcolor{lightgray} & \cellcolor{lightgray}1  & \cellcolor{lightgray}$-0.036^*$ &\cellcolor{lightgray}$0.089$ &\cellcolor{lightgray}$0.105$ & \cellcolor{lightgray}$0.411$ &\cellcolor{lightgray}& \cellcolor{lightgray}$-0.061$ & \cellcolor{lightgray}$0.091$ &\cellcolor{lightgray}$0.099$ & \cellcolor{lightgray}$0.489$\\
    \cellcolor{lightgray} & \cellcolor{lightgray}2  & \cellcolor{lightgray}$0.012$ &\cellcolor{lightgray}$0.082$ &\cellcolor{lightgray}$0.132$ & \cellcolor{lightgray}$0.355$ &\cellcolor{lightgray}& \cellcolor{lightgray}$0.002^*$ & \cellcolor{lightgray}$0.076$ &\cellcolor{lightgray}$0.131$ & \cellcolor{lightgray}$0.434$\\
    \multirow{-3}{*}{\cellcolor{lightgray}Reverb (RE)} &\cellcolor{lightgray}3 & \cellcolor{lightgray}$0.061$ &\cellcolor{lightgray}$0.014^*$ &\cellcolor{lightgray}$0.111$ & \cellcolor{lightgray}$0.218$ &\cellcolor{lightgray}& \cellcolor{lightgray}$0.075$ & \cellcolor{lightgray}$0.027$ &\cellcolor{lightgray}$0.115$ & \cellcolor{lightgray}$0.319$\\
    \multirow{3}{*}{Speed (SP)}  & 1        & $-0.045$ & $0.024^*$ & $0.197$ &$0.368$ & & $-0.060$ & $0.034^*$ & $0.194$ & $0.506$\\
    & 2        & $0.087$ & $0.303$ & $0.212$ &$0.510$ & & $0.055$ & $0.335$ & $0.207$ & $0.611$\\
    & 3        & $0.132$ & $0.009^*$ & $0.097$ &$0.214$ & & $0.0.134$ & $0.029^*$ & $0.122$ & $0.330$\\
    \cellcolor{lightgray} & \cellcolor{lightgray}1  & \cellcolor{lightgray}$-0.074$ &\cellcolor{lightgray}$0.004^*$ &\cellcolor{lightgray}$0.091$ & \cellcolor{lightgray}$0.374$ &\cellcolor{lightgray}& \cellcolor{lightgray}$-0.088$ & \cellcolor{lightgray}$0.009^*$ &\cellcolor{lightgray}$0.088$ & \cellcolor{lightgray}$0.572$\\
    \cellcolor{lightgray} & \cellcolor{lightgray}2  & \cellcolor{lightgray}$-0.031^*$ &\cellcolor{lightgray}$0.030^*$ &\cellcolor{lightgray}$0.173$ & \cellcolor{lightgray}$0.508$ &\cellcolor{lightgray} & \cellcolor{lightgray}$-0.057$ & \cellcolor{lightgray}$0.044$ &\cellcolor{lightgray}$0.164$ & \cellcolor{lightgray}$0.674$ \\
    \multirow{-3}{*}{\cellcolor{lightgray}Distortion (DI)} & \cellcolor{lightgray}3 & \cellcolor{lightgray}$0.139$ &\cellcolor{lightgray}$0.208$ &\cellcolor{lightgray}$0.234$ & \cellcolor{lightgray}$0.598$ &\cellcolor{lightgray} & \cellcolor{lightgray}$0.109$ & \cellcolor{lightgray}$0.239$ &\cellcolor{lightgray}$0.226$ & \cellcolor{lightgray}$0.704$\\
    \multirow{3}{*}{Pitch shift (PS)}  & 1        & $-0.067$ & $0.014^*$ & $0.047$ &$0.295$ & & $-0.083$ & $0.022^*$ & $0.036$ &$0.381$ \\
    & 2        & $0.007^*$ & $0.086$ & $0.110$ & $0.395$ & & $-0.024^*$ & $0.102$ & $0.106$ & $0.523$\\
    & 3        & $0.174$ & $0.229$ & $0.152$ & $0.502$ & & $0.137$ & $0.271$ & $150$ & $0.592$\\
    \cellcolor{lightgray} & \cellcolor{lightgray}1  & \cellcolor{lightgray}$-0.020^*$ &\cellcolor{lightgray}$0.259$ &\cellcolor{lightgray}$0.141$ & \cellcolor{lightgray} $0.463$ &\cellcolor{lightgray}& \cellcolor{lightgray}$-0.036^*$ & \cellcolor{lightgray}$0.290$ &\cellcolor{lightgray}$0.132$ & \cellcolor{lightgray}$0.641$\\
    \cellcolor{lightgray} & \cellcolor{lightgray}2  & \cellcolor{lightgray}$0.004^*$ &\cellcolor{lightgray}$0.136$ &\cellcolor{lightgray}$0.169$ & \cellcolor{lightgray} $0.535$ &\cellcolor{lightgray}& \cellcolor{lightgray}$-0.003^*$ & \cellcolor{lightgray}$0.158$ &\cellcolor{lightgray}$0.163$ & \cellcolor{lightgray}$0.699$ \\
    \multirow{-3}{*}{\cellcolor{lightgray}Background noise (BN)} & \cellcolor{lightgray}3 & \cellcolor{lightgray}$0.002^*$ &\cellcolor{lightgray}$0.007^*$ &\cellcolor{lightgray}$0.198$ & \cellcolor{lightgray}$0.554$ &\cellcolor{lightgray} & \cellcolor{lightgray}$-0.006^*$ & \cellcolor{lightgray}$0.044^*$ &\cellcolor{lightgray}$0.190$ & \cellcolor{lightgray}$0.711$\\
    \bottomrule
    \end{tabular}
    \begin{tablenotes}
        \item[$\dagger$] A positive Kendall's $\tau$-statistic means when module 1's uncertainty score increases on corrupted data comparing with it on clean data, then module 2's uncertainty score also increases. A larger $\tau$-statistic suggests more significant correlation.
        \item[$*$] This result is not statistically significant ($p > 0.0001$).
    \end{tablenotes}
    \end{threeparttable}
    }
    \label{tab:speech_uncertainty_rel}
\end{table}

We show Kendall's $\tau$ test results on the ByteTrack system and SpeechBrain system in Table~\ref{tab:tracking_uncertainty_rel} and Table~\ref{tab:speech_uncertainty_rel}, while there is no evidence suggesting that there exists any correlation between different modules' uncertainty scores in the ConvLab-2 system. We find that when using Track/SORT as the tracker in the ByteTrack system, Kendall's $\tau$-statistic suggests that there is some correlation between the detector's and tracker's uncertainty score change. However, when using DeepSORT as the tracker, Kendall's $\tau$-statistic is usually around 0, suggesting that there is no correlation. One plausible explanation is that, different from Track/SORT only takes the bounding box from the detector as input, whereas DeepSORT also takes the image features as an extra input (See Fig.~\ref{fig:sys_arch} (B)), resulting in a more complex data flow in the system. In the SpeechBrain system, we find that Kendall's $\tau$-statistic suggests there is some correlation between two modules' uncertainty score change. Specifically, some configurations have strong correlations, \eg, Configuration 4 (Wav2Vec2+RNN on CommonVoice-it dataset).

\begin{tcolorbox}[size=title]
{
\textbf{Answer to RQ2:} In most cases, both studied uncertainty estimation methods (\ie, VR and VRO) can capture the difference in the model's prediction change on clean data and corrupted data according to Wilcoxon signed-rank test results. In some cases (\eg, ByteTrack system), VRO performs better than VR. We also find that one module's uncertainty score change does not always correlate to another module's uncertainty score change according to Kendall's $\tau$ test in an AI system.
}
\end{tcolorbox}

\subsection{RQ3. How do the errors propagate among modules in AI systems?}
In RQ2 and RQ3, we've investigated how our OOD detection and uncertainty estimation techniques work on AI systems. RQ3 aims to investigate whether such OOD/uncertainty information correlates to the system output error, which could be helpful to analyze the error propagation effects. We use Kendall's $\tau$ test to measure if there is any correlation between each module's uncertainty score change and system output error change. Specifically, we leverage the results from the SpeechBrain system as an example in this section. Based on our findings in RQ2 and RQ3, we use VRO as the uncertainty estimation technique and KDE as the OOD detection technique.

\begin{table}[htbp]
    \scriptsize
    \centering
    \caption{Kendall's $\tau$ test on the correlation between module's uncertainty (VRO) and system output error in SpeechBrain system.}
    \begin{threeparttable}
    \resizebox{\linewidth}{!}{
    \begin{tabular}{lcccccccccccc}
    \toprule
    \multirow{2}{*}{\textbf{Corruption}} & \multirow{2}{*}{\textbf{Severity}} & \multicolumn{2}{c}{\textbf{Configuration 1}} && \multicolumn{2}{c}{\textbf{Configuration 2}} && \multicolumn{2}{c}{\textbf{Configuration 3}} && \multicolumn{2}{c}{\textbf{Configuration 4}}\\
    \cmidrule{3-4} \cmidrule{6-7} \cmidrule{9-10} \cmidrule{12-13}
    & & \multicolumn{1}{c}{M-1$^\dagger$} & \multicolumn{1}{c}{M-2$^\diamond$} && \multicolumn{1}{c}{M-1$^\dagger$} & \multicolumn{1}{c}{M-2$^\diamond$} && \multicolumn{1}{c}{M-1$^\dagger$} & \multicolumn{1}{c}{M-2$^\diamond$} && \multicolumn{1}{c}{M-1$^\dagger$} & \multicolumn{1}{c}{M-2$^\diamond$}\\
    \midrule
    \cellcolor{lightgray} & \cellcolor{lightgray}1  & \cellcolor{lightgray}$-0.06$ & \cellcolor{lightgray}$0.34$ &\cellcolor{lightgray}& \cellcolor{lightgray}$0.08$ & \cellcolor{lightgray}$0.19$ &\cellcolor{lightgray}& \cellcolor{lightgray}$0.09$ & \cellcolor{lightgray}$0.27$ & \cellcolor{lightgray} & \cellcolor{lightgray}$0.23$ & \cellcolor{lightgray}$0.34$\\
    \cellcolor{lightgray} & \cellcolor{lightgray}2  & \cellcolor{lightgray}$-0.01^*$ & \cellcolor{lightgray}$0.48$ &\cellcolor{lightgray}& \cellcolor{lightgray}$0.09$ & \cellcolor{lightgray}$0.26$ &\cellcolor{lightgray}& \cellcolor{lightgray}$0.11$ & \cellcolor{lightgray}$0.35$ & \cellcolor{lightgray} & \cellcolor{lightgray}$0.23$ & \cellcolor{lightgray}$0.47$\\
    \multirow{-3}{*}{\cellcolor{lightgray}Reverb (RE)} & \cellcolor{lightgray}3 & \cellcolor{lightgray}$0.01^*$ & \cellcolor{lightgray}$0.40$ &\cellcolor{lightgray}& \cellcolor{lightgray}$0.03^*$ & \cellcolor{lightgray}$0.23$ &\cellcolor{lightgray}& \cellcolor{lightgray}$0.10$ & \cellcolor{lightgray}$0.36$ & \cellcolor{lightgray} & \cellcolor{lightgray}$0.12$ & \cellcolor{lightgray}$0.41$\\
    \multirow{3}{*}{Speed (SP)}  & 1        & $-0.06$ & $0.32$ & & $0.01^*$ & $0.05$ && $0.18$ & $0.38$ & & $0.19$ & $0.29$\\
    & 2        & $0.04^*$ & $0.62$ & & $0.34$ & $0.53$ && $0.19$ & $0.40$ & & $0.29$ & $0.46$\\
    & 3        & $0.03^*$ & $0.27$ & & $-0.01^*$ & $0.81$ && $0.10$ & $0.24$ & & $0.05$ & $0.23$\\
    \cellcolor{lightgray} & \cellcolor{lightgray}1  & \cellcolor{lightgray}$-0.08$ & \cellcolor{lightgray}$0.33$ &\cellcolor{lightgray}& \cellcolor{lightgray}$-0.01^*$ & \cellcolor{lightgray}$0.02$ &\cellcolor{lightgray}& \cellcolor{lightgray}$0.10$ & \cellcolor{lightgray}$0.24$ & \cellcolor{lightgray} & \cellcolor{lightgray}$0.22$ & \cellcolor{lightgray}$0.29$\\
    \cellcolor{lightgray} & \cellcolor{lightgray}2  & \cellcolor{lightgray}$-0.03^*$ & \cellcolor{lightgray}$0.37$ &\cellcolor{lightgray}& \cellcolor{lightgray}$0.01^*$ & \cellcolor{lightgray}$0.11$ &\cellcolor{lightgray}& \cellcolor{lightgray}$0.16$ & \cellcolor{lightgray}$0.35$ & \cellcolor{lightgray} & \cellcolor{lightgray}$0.35$ & \cellcolor{lightgray}$0.43$\\
    \multirow{-3}{*}{\cellcolor{lightgray}Distortion (DI)} & \cellcolor{lightgray}3 & \cellcolor{lightgray}$0.10$ & \cellcolor{lightgray}$0.63$ &\cellcolor{lightgray}& \cellcolor{lightgray}$0.24$ & \cellcolor{lightgray}$0.52$ &\cellcolor{lightgray}& \cellcolor{lightgray}$0.21$ & \cellcolor{lightgray}$0.49$ & \cellcolor{lightgray} & \cellcolor{lightgray}$0.46$ & \cellcolor{lightgray}$0.61$\\
    \multirow{3}{*}{Pitch shift (PS)}  & 1        & $-0.09$ & $0.30$ & & $-0.01^*$ & $0.01$ && $0.03$ & $0.17$ & & $0.12$ & $0.21$\\
    & 2        & $-0.01^*$ & $0.38$ & & $0.06$ & $0.15$ && $0.11$ & $0.28$ & & $0.19$ & $0.30$\\
    & 3        & $0.13$ & $0.53$ & & $0.23$ & $0.42$ && $0.14$ & $0.37$ & & $0.29$ & $0.44$\\
    \cellcolor{lightgray} & \cellcolor{lightgray}1  & \cellcolor{lightgray}$-0.04^*$ & \cellcolor{lightgray}$0.40$ &\cellcolor{lightgray}& \cellcolor{lightgray}$0.30$ & \cellcolor{lightgray}$0.48$ &\cellcolor{lightgray}& \cellcolor{lightgray}$0.12$ & \cellcolor{lightgray}$0.33$ & \cellcolor{lightgray} & \cellcolor{lightgray}$0.28$ & \cellcolor{lightgray}$0.37$\\
    \cellcolor{lightgray} & \cellcolor{lightgray}2  & \cellcolor{lightgray}$-0.01^*$ & \cellcolor{lightgray}$0.46$ &\cellcolor{lightgray}& \cellcolor{lightgray}$0.17$ & \cellcolor{lightgray}$0.43$ &\cellcolor{lightgray}& \cellcolor{lightgray}$0.15$ & \cellcolor{lightgray}$0.41$ & \cellcolor{lightgray} & \cellcolor{lightgray}$0.34$ & \cellcolor{lightgray}$0.47$\\
    \multirow{-3}{*}{\cellcolor{lightgray}Background noise (BN)} & \cellcolor{lightgray}3 & \cellcolor{lightgray}$-0.01^*$ & \cellcolor{lightgray}$0.51$ &\cellcolor{lightgray}& \cellcolor{lightgray}$-0.01^*$ & \cellcolor{lightgray}$0.25$ &\cellcolor{lightgray}& \cellcolor{lightgray}$0.17$ & \cellcolor{lightgray}$0.46$ & \cellcolor{lightgray} & \cellcolor{lightgray}$0.36$ & \cellcolor{lightgray}$0.52$\\
    \bottomrule
    \end{tabular}
    }
    \begin{tablenotes}
        \item[$\dagger$] $\tau$-statistic of the correlation between Module 1's (M-1) uncertainty change (VRO) and system output error (WER) change.
        \item[$\diamond$] $\tau$-statistic of the correlation between Module 2's (M-2) uncertainty change (VRO) and system output error (WER) change.
        \item[$*$] This result is not statistically significant ($p>0.0001$).
    \end{tablenotes}
    \end{threeparttable}
    \label{tab:speech_error1}
\end{table}

\begin{table}[htbp]
    \scriptsize
    \centering
    \caption{Kendall's $\tau$ test on the correlation between module's OOD score (KDE) and system output error in SpeechBrain system.}
    \begin{threeparttable}
    \resizebox{\linewidth}{!}{
    \begin{tabular}{lcccccccccccc}
    \toprule
    \multirow{2}{*}{\textbf{Corruption}} & \multirow{2}{*}{\textbf{Severity}} & \multicolumn{2}{c}{\textbf{Configuration 1}} && \multicolumn{2}{c}{\textbf{Configuration 2}} && \multicolumn{2}{c}{\textbf{Configuration 3}} && \multicolumn{2}{c}{\textbf{Configuration 4}}\\
    \cmidrule{3-4} \cmidrule{6-7} \cmidrule{9-10} \cmidrule{12-13}
    & & \multicolumn{1}{c}{M-1$^\dagger$} & \multicolumn{1}{c}{M-2$^\diamond$} && \multicolumn{1}{c}{M-1$^\dagger$} & \multicolumn{1}{c}{M-2$^\diamond$} && \multicolumn{1}{c}{M-1$^\dagger$} & \multicolumn{1}{c}{M-2$^\diamond$} && \multicolumn{1}{c}{M-1$^\dagger$} & \multicolumn{1}{c}{M-2$^\diamond$}\\
    \midrule
    \cellcolor{lightgray} & \cellcolor{lightgray}1  & \cellcolor{lightgray}$0.04^*$ & \cellcolor{lightgray}$0.03^*$ &\cellcolor{lightgray}& \cellcolor{lightgray}$0.02^*$ & \cellcolor{lightgray}$0.34$ &\cellcolor{lightgray}& \cellcolor{lightgray}$0.01^*$ & \cellcolor{lightgray}$0.23$ & \cellcolor{lightgray} & \cellcolor{lightgray}$-0.01^*$ & \cellcolor{lightgray}$0.27$\\
    \cellcolor{lightgray} & \cellcolor{lightgray}2  & \cellcolor{lightgray}$-0.03^*$ & \cellcolor{lightgray}$-0.01^*$ &\cellcolor{lightgray}& \cellcolor{lightgray}$0.09$ & \cellcolor{lightgray}$0.40$ &\cellcolor{lightgray}& \cellcolor{lightgray}$-0.01^*$ & \cellcolor{lightgray}$0.15$ & \cellcolor{lightgray} & \cellcolor{lightgray}$-0.02^*$ & \cellcolor{lightgray}$0.35$\\
    \multirow{-3}{*}{\cellcolor{lightgray}Reverb (RE)} & \cellcolor{lightgray}3 & \cellcolor{lightgray}$0.06$ & \cellcolor{lightgray}$0.04^*$ &\cellcolor{lightgray}& \cellcolor{lightgray}$0.12$ & \cellcolor{lightgray}$0.22$ &\cellcolor{lightgray}& \cellcolor{lightgray}$0.02$ & \cellcolor{lightgray}$-0.05^*$ & \cellcolor{lightgray} & \cellcolor{lightgray}$-0.01^*$ & \cellcolor{lightgray}$0.13$\\
    \multirow{3}{*}{Speed (SP)}  & 1        & $0.01^*$ & $0.04^*$ & & $0.05^*$ & $0.14$ && $0.06$ & $0.36$ & & $0.02^*$ & $0.21$\\
    & 2        & $-0.01^*$ & $0.01^*$ & & $0.31$ & $0.60$ && $0.03$ & $0.20$ & & $-0.01^*$ & $0.38$\\
    & 3        & $0.02$ & $0.09$ & & $0.04$ & $0.81$ && $0.01^*$ & $-0.01^*$ & & $0.01^*$ & $0.04$\\
    \cellcolor{lightgray} & \cellcolor{lightgray}1  & \cellcolor{lightgray}$0.03^*$ & \cellcolor{lightgray}$0.02^*$ &\cellcolor{lightgray}& \cellcolor{lightgray}$0.07$ & \cellcolor{lightgray}$0.13$ &\cellcolor{lightgray}& \cellcolor{lightgray}$0.10$ & \cellcolor{lightgray}$0.24$ & \cellcolor{lightgray} & \cellcolor{lightgray}$0.06$ & \cellcolor{lightgray}$0.24$\\
    \cellcolor{lightgray} & \cellcolor{lightgray}2  & \cellcolor{lightgray}$0.05^*$ & \cellcolor{lightgray}$0.03^*$ &\cellcolor{lightgray}& \cellcolor{lightgray}$0.09$ & \cellcolor{lightgray}$0.24$ &\cellcolor{lightgray}& \cellcolor{lightgray}$0.13$ & \cellcolor{lightgray}$0.36$ & \cellcolor{lightgray} & \cellcolor{lightgray}$0.08$ & \cellcolor{lightgray}$0.38$\\
    \multirow{-3}{*}{\cellcolor{lightgray}Distortion (DI)} & \cellcolor{lightgray}3 & \cellcolor{lightgray}$0.04*$ & \cellcolor{lightgray}$-0.04^*$ &\cellcolor{lightgray}& \cellcolor{lightgray}$0.14$ & \cellcolor{lightgray}$0.61$ &\cellcolor{lightgray}& \cellcolor{lightgray}$0.21$ & \cellcolor{lightgray}$0.43$ & \cellcolor{lightgray} & \cellcolor{lightgray}$0.11$ & \cellcolor{lightgray}$0.53$\\
    \multirow{3}{*}{Pitch shift (PS)}  & 1        & $-0.01^*$ & $0.01^*$ & & $0.03^*$ & $0.12$ && $0.04$ & $0.14$ & & $0.01^*$ & $0.16$\\
    & 2        & $0.02^*$ & $0.01*$ & & $0.11$ & $0.27$ && $0.06$ & $0.26$ & & $0.01^*$ & $0.25$\\
    & 3        & $0.04^*$ & $-0.01^*$ & & $0.26$ & $0.54$ && $0.06$ & $0.34$ & & $-0.01^*$ & $0.39$\\
    \cellcolor{lightgray} & \cellcolor{lightgray}1  & \cellcolor{lightgray}$0.06$ & \cellcolor{lightgray}$0.02^*$ &\cellcolor{lightgray}& \cellcolor{lightgray}$0.38$ & \cellcolor{lightgray}$0.62$ &\cellcolor{lightgray}& \cellcolor{lightgray}$0.05$ & \cellcolor{lightgray}$0.27$ & \cellcolor{lightgray} & \cellcolor{lightgray}$0.02^*$ & \cellcolor{lightgray}$0.29$\\
    \cellcolor{lightgray} & \cellcolor{lightgray}2  & \cellcolor{lightgray}$0.07$ & \cellcolor{lightgray}$-0.01^*$ &\cellcolor{lightgray}& \cellcolor{lightgray}$0.34$ & \cellcolor{lightgray}$0.49$ &\cellcolor{lightgray}& \cellcolor{lightgray}$0.07$ & \cellcolor{lightgray}$0.29$ & \cellcolor{lightgray} & \cellcolor{lightgray}$0.03$ & \cellcolor{lightgray}$0.38$\\
    \multirow{-3}{*}{\cellcolor{lightgray}Background noise (BN)} & \cellcolor{lightgray}3 & \cellcolor{lightgray}$0.06$ & \cellcolor{lightgray}$0.01^*$ &\cellcolor{lightgray}& \cellcolor{lightgray}$0.23$ & \cellcolor{lightgray}$0.26$ &\cellcolor{lightgray}& \cellcolor{lightgray}$0.08$ & \cellcolor{lightgray}$0.21$ & \cellcolor{lightgray} & \cellcolor{lightgray}$0.06$ & \cellcolor{lightgray}$0.41$\\
    \bottomrule
    \end{tabular}
    }
    \begin{tablenotes}
        \item[$\dagger$] $\tau$-statistic of the correlation between Module 1's (M-1) OOD score change (KDE) and system output error (WER) change.
        \item[$\diamond$] $\tau$-statistic of the correlation between Module 2's (M-2) OOD score change (KDE) and system output error (WER) change.
        \item[$*$] This result is not statistically significant ($p>0.0001$).
    \end{tablenotes}
    \end{threeparttable}
    \label{tab:speech_error2}
\end{table}

Table~\ref{tab:speech_error1} and Table~\ref{tab:speech_error2} summarize the correlation between the module's OOD/uncertainty score change and system output error. Overall, we find that two indicators (\ie, VRO and KDE) behave similarly, where the correlation between OOD/uncertainty score and system output error is more significant when measuring on the second module. Surprisingly, in RQ2, we find that OOD detection methods are more effective in capturing the data distribution shift when measuring on the first module. However, we find that the first module's OOD score change does not have much correlation with the system output error. In fact, both OOD scores and uncertainty scores measured on the first module do not show an obvious correlation with the system output error with configurations 1 and 2. While there is some correlation between the first module's uncertainty score change and system output error with configurations 3 and 4, however, such correlation is always less significant than it between the second module's uncertainty score change and system output error. We can also connect these observations with results in Table~\ref{tab:speech_uncertainty_rel}, where Kendall's $\tau$ test suggests that there is some correlation between two modules' uncertainty with configurations 3 and 4. These results indicate that for the SpeechBrain system, when the second module's OOD/uncertainty score is high, it's relatively possible to trigger an error output. 

Besides, we also find that the correlation does not always become stronger when the severity level increases. For instance, in Table~\ref{tab:speech_error1}, we can observe that when against level-2 \textbf{SP} corruption, the second module's uncertainty score change of configuration 1, 3~\&~4 has a stronger correlation with the system output error compared with them against level-3 corruption. Recall that when against level-3 \textbf{SP} corruption, the system output error is relatively higher (see Table~\ref{tab:speech}). This indicates that when against high severity-level corruption patterns, current uncertainty estimation and OOD detection techniques could be less effective for risk assessments. Therefore, more advanced OOD detection techniques and uncertainty estimation are needed for analyzing AI systems, especially when interpreting system performance against severely corrupted data.

\begin{tcolorbox}[size=title]
{
\textbf{Answer to RQ3:} We find that for a serial executed system, \eg, SpeechBrain, neither the first module's OOD nor uncertainty information strongly correlates to the system output error. However, both the second module's OOD and uncertainty information can indicate potential output errors of the system. Besides, when the corruption severity level increases, OOD detection and uncertainty estimation sometimes become less effective as indicators of the system's potential risk. This result also implies that more advanced OOD detection techniques/uncertainty estimation might be needed at the system level.
}
\end{tcolorbox}

\subsection{RQ4. How do simple combinations of OOD analysis- and uncertainty-based method improve the reliability of AI systems?}

\begin{figure}[t]
    \centering
    \includegraphics[width=0.95\linewidth]{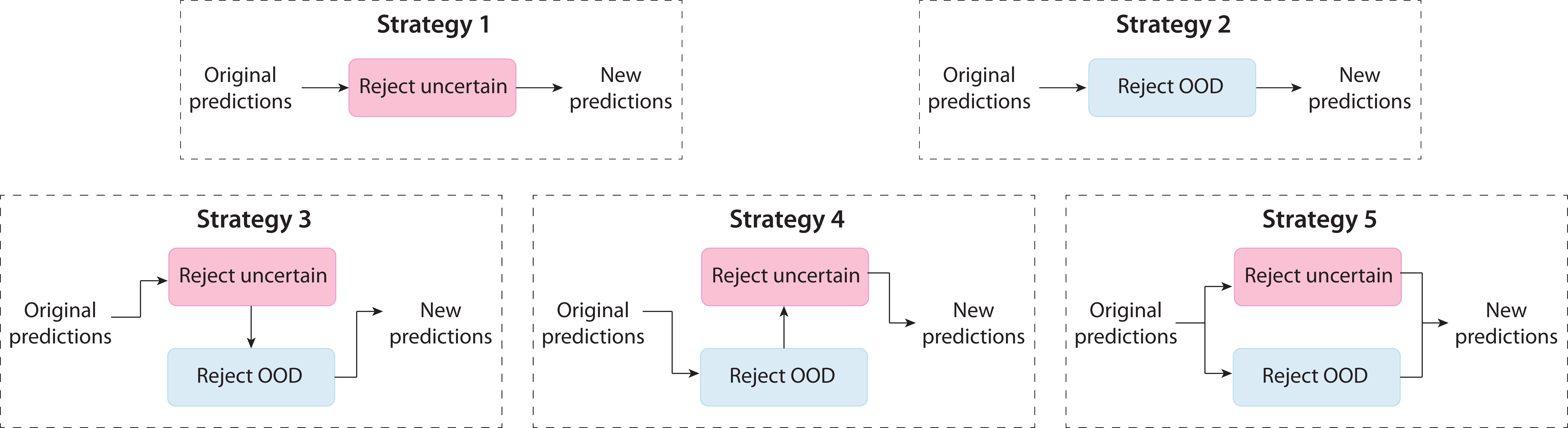}
    \caption{Five simple risk assessment strategies we used in this study.}
    \label{fig:risk_strategy}
\end{figure}

{
In RQ4, we first use the SpeechBrain system as an example to show that a simple combination of OOD analysis- and uncertainty-based methods can improve the reliability of AI systems. We compare the following strategies (See Fig.4 for the workflow of each) to reject a certain number of risky predictions for the systems, suppose we are rejecting $n$ risky predictions: 

\begin{itemize}[leftmargin=*]
    \item \textbf{Strategy 1 (S1)}: Reject only uncertain predictions. Through this strategy, we will reject the predictions with top-$n$ highest uncertainty scores.
    \item \textbf{Strategy 2 (S2)}: Reject only OOD data. Through this strategy, we will reject the predictions with top-$n$ highest OOD scores.
    \item \textbf{Strategy 3 (S3)}: First reject uncertain predictions, then reject OOD data. Through this strategy, we will first reject the predictions with top-$\frac{n}{2}$ highest uncertainty scores, then reject the predictions with top-$\frac{n}{2}$ highest OOD scores.
    \item \textbf{Strategy 4 (S4)}: First reject OOD data, then reject uncertain predictions. Through this strategy, we will first reject the predictions with top-$\frac{n}{2}$ highest OOD scores, then reject the predictions with top-$\frac{n}{2}$ highest uncertainty scores.
    \item \textbf{Strategy 5 (S5)}: Reject either uncertain predictions or OOD data simultaneously. Through this strategy, we will first reject rank the predictions according to uncertainty and OOD scores separately, then we will reject top-$n$ common predictions between this two ranked lists.
\end{itemize}
}

\begin{table}[htbp]
    \scriptsize
    \centering
    \caption{SpeechBrain’s performance improvement on LibriSpeech dataset by rejecting 1\% predictions.}
    \begin{threeparttable}
        \centering
        \resizebox{\linewidth}{!}{
        \begin{tabular}{lccccccccccccc}
        \toprule
        \multirow{2}{*}{\textbf{Corruption}} & \multirow{2}{*}{\textbf{Severity}} & \multicolumn{5}{c}{\textbf{Configuration 1}} && \multicolumn{5}{c}{\textbf{Configuration 2}} \\
        \cmidrule{3-7} \cmidrule{9-13}
        & & S1 (\%) & S2 (\%) & S3 (\%) & S4 (\%) & S5 (\%) && S1 (\%) & S2 (\%) & S3 (\%) & S4 (\%) & S5 (\%) \\
        \midrule
        \cellcolor{lightgray} & \cellcolor{lightgray}1 & \cellcolor{tab_red}$27.55$ & \cellcolor{lightgray}$-0.61$ & \cellcolor{lightgray}$-0.61$ & \cellcolor{lightgray}$27.55$ & \cellcolor{lightgray}$27.21$ &\cellcolor{lightgray}& \cellcolor{lightgray}$2.74$ & \cellcolor{lightgray}$4.53$ & \cellcolor{lightgray}$4.55$ & \cellcolor{lightgray}$2.62$ & \cellcolor{tab_red}$6.69$\\
        \cellcolor{lightgray} & \cellcolor{lightgray}2 & \cellcolor{lightgray}$0.30$ & \cellcolor{lightgray}$0.05$ & \cellcolor{lightgray}$0.05$ & \cellcolor{lightgray}$0.30$ & \cellcolor{tab_red}$9.66$ &\cellcolor{lightgray}& \cellcolor{lightgray}$0.39$ & \cellcolor{lightgray}$0.50$ & \cellcolor{lightgray}$0.53$ & \cellcolor{lightgray}$0.44$ & \cellcolor{tab_red}$0.88$\\
        \multirow{-3}{*}{\cellcolor{lightgray}Reverb (RE)} & \cellcolor{lightgray}3 & \cellcolor{lightgray}$0.33$ & \cellcolor{lightgray}$0.28$ & \cellcolor{lightgray}$0.28$ & \cellcolor{lightgray}$0.33$ & \cellcolor{tab_red}$1.99$ &\cellcolor{lightgray}& \cellcolor{lightgray}$0.01$ & \cellcolor{lightgray}$0.03$ & \cellcolor{lightgray}$0.04$ & \cellcolor{lightgray}$0.04$ & \cellcolor{tab_red}$0.17$\\
        \multirow{3}{*}{Speed (SP)} & 1 &$36.01$ & $1.49$ & $1.49$ & $36.01$ & \cellcolor{tab_red}$37.88$ && $4.77$ & $7.76$ & $7.59$ & $4.22$ & \cellcolor{tab_red}$9.82$\\
        & 2 & $-0.29$ & $0.54$ & $0.54$ & $-0.29$ & \cellcolor{tab_red}$7.12$ && $-0.06$ & $0.36$ & $0.34$ & $0.19$ & \cellcolor{tab_red}$2.51$\\
        & 3 & $-0.05$ & $-0.08$ & $-0.08$ & $-0.05$ & \cellcolor{tab_red}$0.65$ && $0.06$ & \cellcolor{tab_red}$0.11$ & $0.08$ & $0.03$ & \cellcolor{tab_red}$0.11$\\
        \cellcolor{lightgray} & \cellcolor{lightgray}1 & \cellcolor{tab_red}$28.54$ & \cellcolor{lightgray}$-0.47$ & \cellcolor{lightgray}$-0.47$ & \cellcolor{tab_red}$28.54$ & \cellcolor{lightgray}$28.35$ &\cellcolor{lightgray}& \cellcolor{lightgray}$2.28$ & \cellcolor{lightgray}$4.95$ & \cellcolor{lightgray}$5.19$ & \cellcolor{lightgray}$1.78$ & \cellcolor{tab_red}$6.50$\\
        \cellcolor{lightgray} & \cellcolor{lightgray}2 & \cellcolor{lightgray}$13.28$ & \cellcolor{lightgray}$-0.60$ & \cellcolor{lightgray}$-0.60$ & \cellcolor{lightgray}$13.28$ & \cellcolor{tab_red}$28.61$ &\cellcolor{lightgray}& \cellcolor{lightgray}$2.48$ & \cellcolor{lightgray}$5.50$ & \cellcolor{lightgray}$5.49$ & \cellcolor{lightgray}$1.67$ & \cellcolor{tab_red}$6.95$\\
        \multirow{-3}{*}{\cellcolor{lightgray}Distortion (DI)} & \cellcolor{lightgray}3 & \cellcolor{lightgray}$1.21$ & \cellcolor{lightgray}$-0.14$ & \cellcolor{lightgray}$-0.14$ & \cellcolor{lightgray}$1.21$ & \cellcolor{tab_red}$10.10$ &\cellcolor{lightgray}& \cellcolor{lightgray}$1.29$ & \cellcolor{lightgray}$2.14$ & \cellcolor{lightgray}$2.02$ & \cellcolor{lightgray}$1.07$ & \cellcolor{tab_red}$5.29$\\
        \multirow{3}{*}{Pitch shift (PS)} & 1 & $30.18$ & $4.27$ & $0.22$ & $26.48$ & \cellcolor{tab_red}$30.62$ && $3.31$ & $6.41$ & $5.79$ & $2.93$ & \cellcolor{tab_red}$7.61$\\
        & 2 & $7.11$ & $-0.10$ & $-0.25$ & $7.05$ & \cellcolor{tab_red}$27.58$ && $4.14$ & $7.02$ & $6.14$ & $2.62$ & \cellcolor{tab_red}$12.27$\\
        & 3 & $5.44$ & $0.52$ & $0.29$ & $5.58$ & \cellcolor{tab_red}$20.01$ && $2.87$ & $3.08$ & $2.86$ & $2.32$ & \cellcolor{tab_red}$6.12$\\
        \cellcolor{lightgray} & \cellcolor{lightgray}1 & \cellcolor{lightgray}$30.02$ & \cellcolor{lightgray}$-0.25$ & \cellcolor{lightgray}$-0.25$ & \cellcolor{lightgray}$30.24$ & \cellcolor{tab_red}$30.03$ &\cellcolor{lightgray}& \cellcolor{lightgray}$1.27$ & \cellcolor{lightgray}$1.56$ & \cellcolor{lightgray}$1.56$ & \cellcolor{lightgray}$1.27$ & \cellcolor{tab_red}$2.71$\\
        \cellcolor{lightgray} & \cellcolor{lightgray}2 & \cellcolor{lightgray}$2.37$ & \cellcolor{lightgray}$-0.38$ & \cellcolor{lightgray}$-0.38$ & \cellcolor{lightgray}$2.55$ & \cellcolor{tab_red}$20.22$ &\cellcolor{lightgray}& \cellcolor{lightgray}$0.31$ & \cellcolor{lightgray}$0.36$ & \cellcolor{lightgray}$0.36$ & \cellcolor{lightgray}$0.31$ & \cellcolor{tab_red}$1.08$\\
        \multirow{-3}{*}{\cellcolor{lightgray}Background noise (BN)} & \cellcolor{lightgray}3 & \cellcolor{lightgray}$1.85$ & \cellcolor{lightgray}$-0.42$ & \cellcolor{lightgray}$-0.42$ & \cellcolor{lightgray}$1.99$ & \cellcolor{tab_red}$8.50$ &\cellcolor{lightgray}& \cellcolor{lightgray}$0.09$ & \cellcolor{lightgray}$0.09$ & \cellcolor{lightgray}$0.09$ & \cellcolor{lightgray}$0.09$ & \cellcolor{tab_red}$0.16$\\
        \bottomrule
        \end{tabular}
        }
    \end{threeparttable}
    \label{tab:speech_improve1}
\end{table}

\begin{table}[htbp]
    \scriptsize
    \centering
    \caption{SpeechBrain’s performance improvement on CommonVoice-it dataset by rejecting 1\% predictions.}
    \begin{threeparttable}
        \centering
        \resizebox{\linewidth}{!}{
        \begin{tabular}{lccccccccccccc}
        \toprule
        \multirow{2}{*}{\textbf{Corruption}} & \multirow{2}{*}{\textbf{Severity}} & \multicolumn{5}{c}{\textbf{Configuration 3}} && \multicolumn{5}{c}{\textbf{Configuration 4}} \\
        \cmidrule{3-7} \cmidrule{9-13}
        & & S1 (\%) & S2 (\%) & S3 (\%) & S4 (\%) & S5 (\%) && S1 (\%) & S2 (\%) & S3 (\%) & S4 (\%) & S5 (\%) \\
        \midrule
        \cellcolor{lightgray} & \cellcolor{lightgray}1 & \cellcolor{lightgray}$1.15$ & \cellcolor{lightgray}$0.63$ & \cellcolor{lightgray}$0.67$ & \cellcolor{lightgray}$1.26$ & \cellcolor{tab_red}$1.72$ &\cellcolor{lightgray}& \cellcolor{lightgray}$2.21$ & \cellcolor{lightgray}$2.13$ & \cellcolor{lightgray}$1.04$ & \cellcolor{lightgray}$1.86$ & \cellcolor{tab_red}$3.13$\\
        \cellcolor{lightgray} & \cellcolor{lightgray}2 & \cellcolor{lightgray}$0.47$ & \cellcolor{lightgray}$0.37$ & \cellcolor{lightgray}$0.36$ & \cellcolor{lightgray}$0.47$ & \cellcolor{tab_red}$1.13$ &\cellcolor{lightgray}& \cellcolor{lightgray}$0.83$ & \cellcolor{lightgray}$0.85$ & \cellcolor{lightgray}$0.73$ & \cellcolor{lightgray}$0.80$ & \cellcolor{tab_red}$1.89$\\
        \multirow{-3}{*}{\cellcolor{lightgray}Reverb (RE)} & \cellcolor{lightgray}1 & \cellcolor{lightgray}$0.15$ & \cellcolor{lightgray}$0.07$ & \cellcolor{lightgray}$0.06$ & \cellcolor{lightgray}$0.14$ & \cellcolor{tab_red}$0.69$ &\cellcolor{lightgray}& \cellcolor{lightgray}$0.25$ & \cellcolor{lightgray}$0.27$ & \cellcolor{lightgray}$0.24$ & \cellcolor{lightgray}$0.27$ & \cellcolor{tab_red}$1.63$\\
        \multirow{3}{*}{Speed (SP)} & 1 &$1.99$ & $1.51$ & $1.32$ & $1.83$ & \cellcolor{tab_red}$3.25$ && $3.13$ & $1.72$ & $1.69$ & $2.96$ & \cellcolor{tab_red}$4.17$\\
        & 2 & $0.11$ & $0.21$ & $0.19$ & $0.09$ & \cellcolor{tab_red}$1.07$ && $0.30$ & $0.56$ & $0.64$ & $0.34$ & \cellcolor{tab_red}$2.61$\\
        & 3 & $0.00$ & $0.04$ & $0.02$ & $-0.00$ & \cellcolor{tab_red}$0.27$ && $0.01$ & $0.03$ & $0.08$ & $0.02$ & \cellcolor{tab_red}$1.05$\\
        \cellcolor{lightgray} & \cellcolor{lightgray}1 & \cellcolor{lightgray}$1.65$ & \cellcolor{lightgray}$1.79$ & \cellcolor{lightgray}$1.71$ & \cellcolor{lightgray}$1.65$ & \cellcolor{tab_red}$3.13$ &\cellcolor{lightgray}& \cellcolor{lightgray}$1.99$ & \cellcolor{lightgray}$1.69$ & \cellcolor{lightgray}$1.60$ & \cellcolor{lightgray}$1.83$ & \cellcolor{tab_red}$3.21$\\
        \cellcolor{lightgray} & \cellcolor{lightgray}2 & \cellcolor{lightgray}$1.20$ & \cellcolor{lightgray}$1.73$ & \cellcolor{lightgray}$1.64$ & \cellcolor{lightgray}$1.19$ & \cellcolor{tab_red}$3.00$ &\cellcolor{lightgray}& \cellcolor{lightgray}$2.02$ & \cellcolor{lightgray}$1.76$ & \cellcolor{lightgray}$1.51$ & \cellcolor{lightgray}$1.94$ & \cellcolor{tab_red}$3.75$\\
        \multirow{-3}{*}{\cellcolor{lightgray}Distortion (DI)} & \cellcolor{lightgray}1 & \cellcolor{lightgray}$1.05$ & \cellcolor{lightgray}$1.03$ & \cellcolor{lightgray}$0.94$ & \cellcolor{lightgray}$1.06$ & \cellcolor{tab_red}$2.50$ &\cellcolor{lightgray}& \cellcolor{lightgray}$1.38$ & \cellcolor{lightgray}$1.75$ & \cellcolor{lightgray}$1.71$ & \cellcolor{lightgray}$1.33$ & \cellcolor{tab_red}$3.02$\\
        \multirow{3}{*}{Pitch shift (PS)} & 1 & $2.76$ & $1.81$ & $1.57$ & $1.95$ & \cellcolor{tab_red}$3.55$ && $1.70$ & $3.78$ & $3.62$ & $1.60$ & \cellcolor{tab_red}$5.06$\\
        & 2 & $1.84$ & $1.76$ & $1.58$ & $1.47$ & \cellcolor{tab_red}$2.38$ && $3.94$ & $2.09$ & $2.03$ & $3.78$ & \cellcolor{tab_red}$5.40$\\
        & 3 & $0.87$ & $0.76$ & $0.63$ & $0.79$ & \cellcolor{tab_red}$3.01$ && $2.25$ & $2.11$ & $1.27$ & $1.41$ & \cellcolor{tab_red}$4.53$\\
        \cellcolor{lightgray} & \cellcolor{lightgray}1 & \cellcolor{lightgray}$1.63$ & \cellcolor{lightgray}$0.69$ & \cellcolor{lightgray}$0.67$ & \cellcolor{lightgray}$1.62$ & \cellcolor{tab_red}$2.28$ &\cellcolor{lightgray}& \cellcolor{lightgray}$1.94$ & \cellcolor{lightgray}$1.76$ & \cellcolor{lightgray}$1.28$ & \cellcolor{lightgray}$1.91$ & \cellcolor{tab_red}$3.05$\\
        \cellcolor{lightgray} & \cellcolor{lightgray}2 & \cellcolor{lightgray}$1.00$ & \cellcolor{lightgray}$0.50$ & \cellcolor{lightgray}$0.47$ & \cellcolor{lightgray}$0.98$ & \cellcolor{tab_red}$1.70$ &\cellcolor{lightgray}& \cellcolor{lightgray}$1.51$ & \cellcolor{lightgray}$0.99$ & \cellcolor{lightgray}$1.01$ & \cellcolor{lightgray}$1.52$ & \cellcolor{tab_red}$2.48$\\
        \multirow{-3}{*}{\cellcolor{lightgray}Background noise (BN)} & \cellcolor{lightgray}1 & \cellcolor{lightgray}$0.72$ & \cellcolor{lightgray}$0.39$ & \cellcolor{lightgray}$0.39$ & \cellcolor{lightgray}$0.72$ & \cellcolor{tab_red}$1.18$ &\cellcolor{lightgray}& \cellcolor{lightgray}$1.24$ & \cellcolor{lightgray}$0.84$ & \cellcolor{lightgray}$0.84$ & \cellcolor{lightgray}$1.23$ & \cellcolor{tab_red}$2.55$\\
        \bottomrule
        \end{tabular}
        }
    \end{threeparttable}
    \label{tab:speech_improve2}
\end{table}

\begin{figure}[htbp]
    \centering
    \begin{subfigure}[b]{0.95\textwidth}
        \centering
        \includegraphics[width=\textwidth]{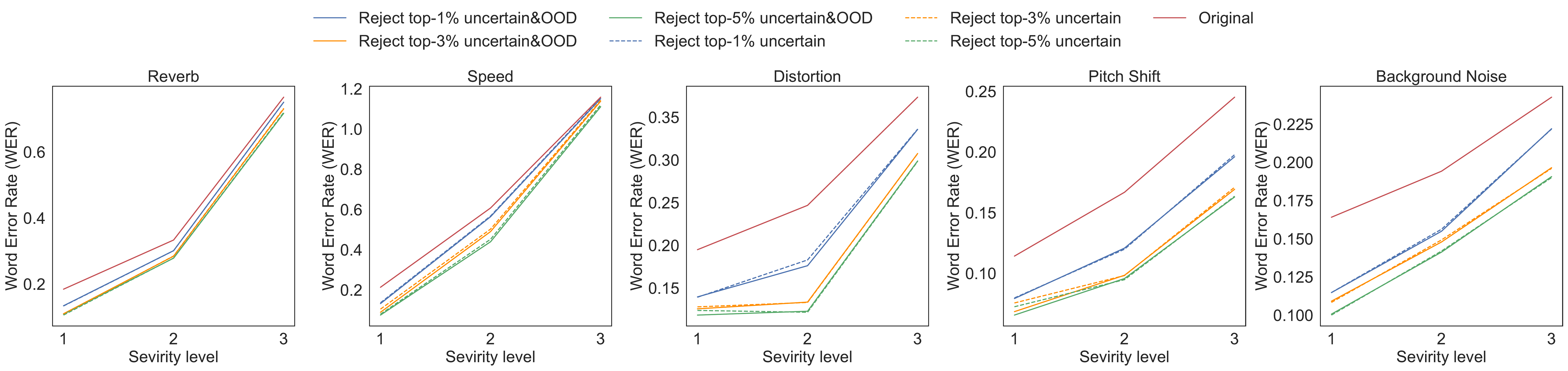}
        \caption{Configuration 1: CRDNN+RNN on LibriSpeech dataset}
        \label{fig:speech_config1}
    \end{subfigure}
     \vfill
     \begin{subfigure}[b]{0.95\textwidth}
        \centering
        \includegraphics[width=\textwidth]{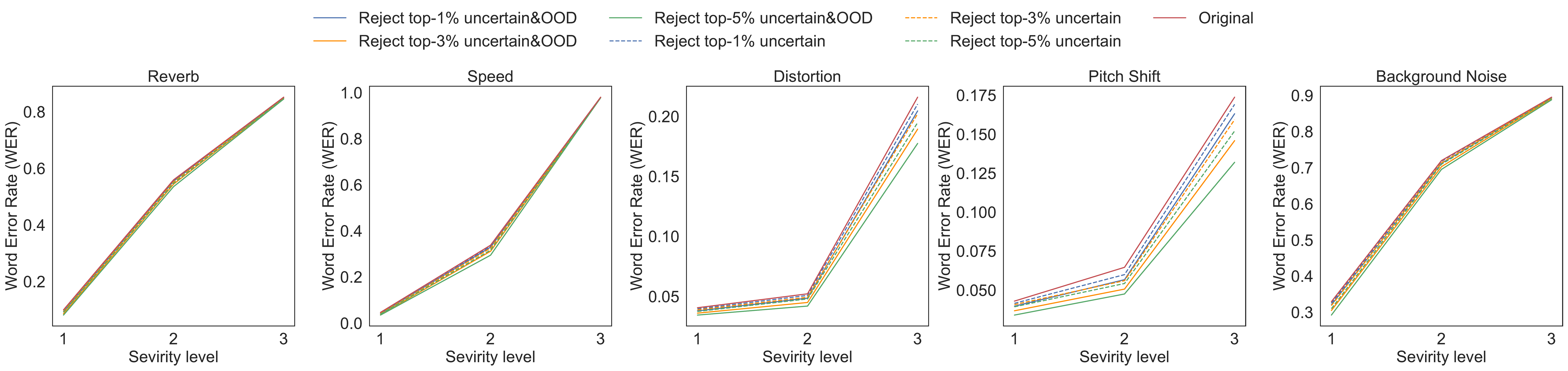}
        \caption{Configuration 2: CRDNN+Transformer on LibriSpeech dataset}
        \label{fig:speech_config2}
    \end{subfigure}
    \vfill
    \begin{subfigure}[b]{0.95\textwidth}
        \centering
        \includegraphics[width=\textwidth]{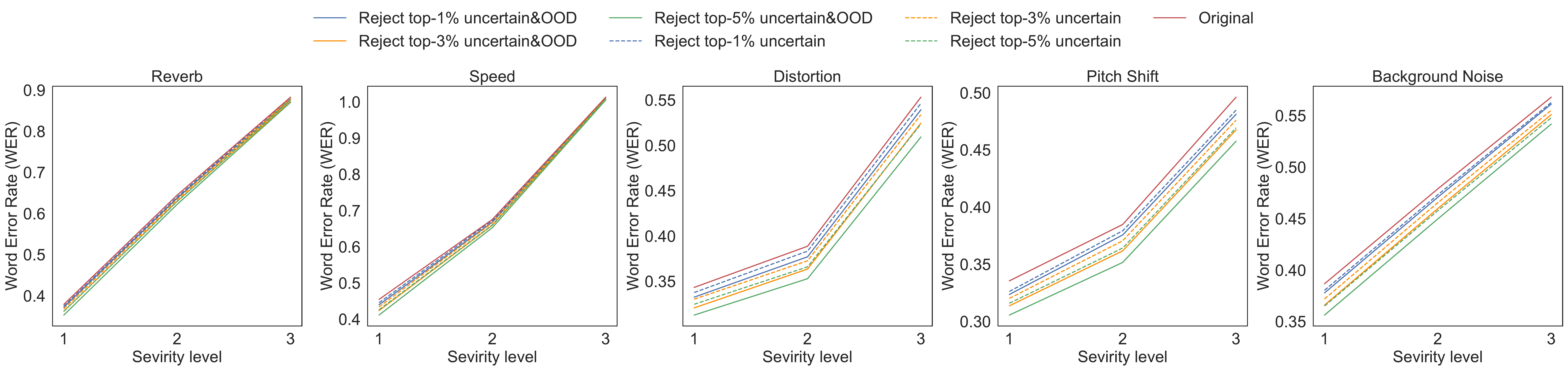}
        \caption{Configuration 3: CRDNN+RNN on CommonVoice-it dataset}
        \label{fig:speech_config3}
    \end{subfigure}
    \vfill\begin{subfigure}[b]{0.95\textwidth}
        \centering
        \includegraphics[width=\textwidth]{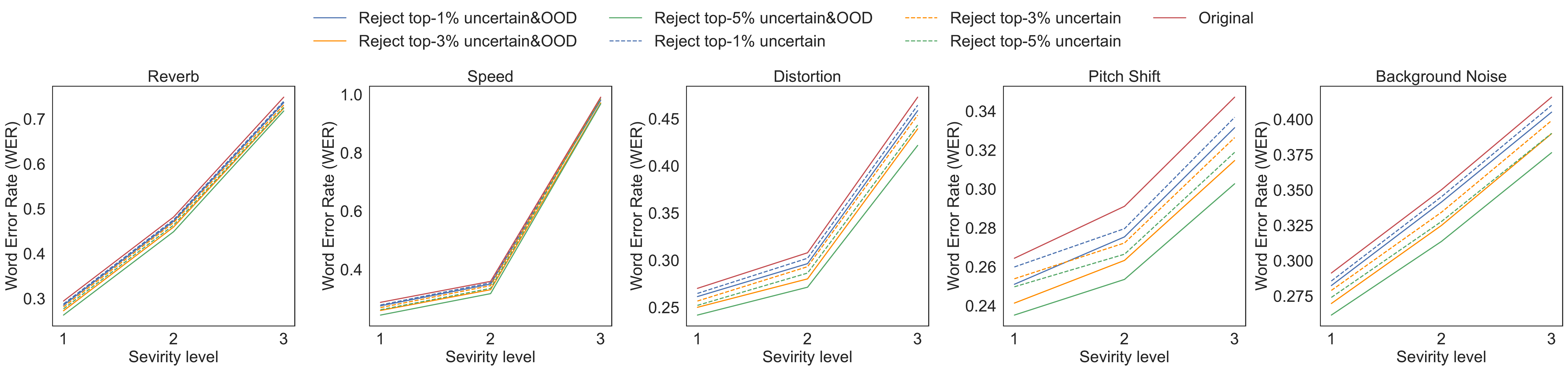}
        \caption{Configuration 4: Wav2Vec2+RNN on CommonVoice-it dataset}
        \label{fig:speech_config4}
    \end{subfigure}
    \caption{SpeechBrain's output error (WER, lower indicates better) changes when rejecting a certain number of uncertain/OOD data.}
    \label{fig:speech_risk}
\end{figure}

Based on our findings in RQ2-4, we use the OOD score measured by KDE on module 2 and the uncertainty score calculated by VRO on module 2 when investigating this research question. In practice, we use these strategies to reject 1\%, 3\%, and 5\% unreliable predictions for different configurations of the SpeechBrain system accordingly. While reporting all the results might take too much space, here we mainly report the full results of rejecting 1\% unreliable predictions in Table~\ref{tab:speech_improve1} and Table~\ref{tab:speech_improve2}. As we can see, the system's performance can be improved by these strategies to different degrees. Overall, we find that the combinations of OOD and uncertainty (S3, S4, and S5) usually perform better than only using one of them (S1 and S2). Among S3, S4, and S5, the improvement through S5 ({\em reject either uncertain predictions or OOD data simultaneously}) is the most significant one compared with other strategies. For instance, in configuration 1, S5 can improve the system's performance by up to 37.9\% (when against level 1 Speed corruption). We also observe that S1 performs better than S2 in some configurations (configuration 1, 3, and 4), while S2 performs better than S1 in configuration 2. These results also suggest that neither OOD nor uncertainty score is always better than the other one, considering a combination of them could be a better choice. We further show the system's performance improvement by S5 compared with S1 when rejecting a different number of unreliable predictions (1\%, 3\%, and 5\%) in Fig.~\ref{fig:speech_risk}. Compared with S1 (dashed blue, orange, and green line vs. the solid red line), S5 improves system performance (solid blue, orange, and green line vs. the solid red line). 

We also find that these strategies are usually more effective in improving system performance against lower severity level (\ie, level-1) corruption compared with higher ones. For instance, in configuration 1, S1 can usually improve the system performance by over 27\% when against level-1 corruption. However, the improvement on level-2 and level-3 ones is limited, especially on level-3. When using a combination of OOD and uncertainty, \ie, S5, the system performance improvement is still limited when against level-3 corruption. These results align with our findings in RQ4, implying that OOD detection techniques and uncertainty estimation for AI systems can be further improved against highly corrupted data.

\begin{tcolorbox}[size=title]
{
\textbf{Answer to RQ4:} We demonstrate that simple strategies of combining OOD analysis- and uncertainty-based methods can improve the system's performance and reliability. In the SpeechBrain system, we find that only rejecting 1\% uncertain/OOD data can improve the system's performance by up to 37.8\%. These results could motivate the need for more advanced research strategies for AI system risk assessments.
}
\end{tcolorbox}

\section{Discussions and Future Opportunities}\label{sec:discussion}
In this section, we discuss the implications of our empirical studies as well as the identified future opportunities.

\subsection{Common corruption patterns bring risks to AI systems that should be seriously treated towards enhancing the system quality.}
Common corruption patterns have been shown threatening to the performance of unit-level AI modules (\ie, a single DNN model), and performance on corrupted data thus becomes an important indicator of analyzing a single DNN's robustness. Yet limited investigation has been made on how these corruption patterns can put AI systems at risk, in this study, we leverage common corruption patterns from three different domains (\ie, image, text, and audio) to create corrupted datasets with different levels of severity. Our results from RQ0 show that, even at the system level, these common corruption patterns still pose threats to the system's quality. For instance, in the SpeechBrain system (Table~\ref{tab:speech}), even a low level of severity could significantly decrease the system performance by 185.7\%~\footnote{We manually check the audio clips with level-1 corruption and find that such corruption cannot affect human's perception.}. We also find that for an identical system, \eg, ByteTrack, both fully AI-enabled configuration (\eg, YOLO+DeepSORT) and partial AI-enabled one (YOLO+Track) can be threatened by common corruption patterns. Therefore, our empirical study results imply that addressing the AI system's robustness issue should be an essential and urgent need. Robustness enhancement techniques and software repair can be two possible solutions to address this issue. However, even though there have been a lot of studies on a single DNN model's robustness enhancement or repair~\cite{Yu2021,xie2021rnnrepair,wang2019repairing,sotoudeh2019correcting,zhang2019apricot}, yet there is no study on the system-level robustness enhancement or repair. Hence, both these two topics can be interesting research opportunities.

Another interesting finding from analyzing an AI system's performance on corrupted data is that different configurations can have different performances when against the same corruption pattern for an identical system and the same dataset. One example is the SpeechBrain system's performance on the corrupted LibriSpeech dataset, where configuration 1 has much better performance against \textit{Background noise} corruption than configuration 2. By contrast, configuration 2 is much better at against \textit{Distortion} corruption. These results imply that different system configurations can bring different robustness performances. As suggested by international standards ISO 26262~\cite{iso26262} and ISO/PAS 21448 (SOTIF)~\cite{iso21448}, modular redundancy is an important strategy to improve system quality and reliability. Therefore, developers can try to deploy differently configured systems in parallel to achieve better system reliability.

\subsection{OOD detection and uncertainty estimation techniques are still effective when analyzing AI systems, while not every module's OOD or uncertainty information could reveal the system's potential risks.}
From our empirical study results in RQ1 and RQ2, we find that two general OOD detection approaches and two general uncertainty estimation metrics are effective when analyzing subject AI systems, while the choice of each can be varied. Two general OOD detection techniques used in this study can effectively measure the data distribution difference between clean data and corrupted data. Surprisingly, we find that applying OOD detection on different modules can reveal different OOD data for an identical system. For instance, in the SpeechBrain system, we can observe that OOD data identified by two modules are very different, whereas in most cases, the similarity between top-1\% OOD data identified by two modules is less than 5\%. Hence the results imply that when utilizing OOD detection techniques as an approach to risk assessment, it's needed to measure the data distribution difference based on different AI modules in a system.

In the SpeechBrain system and ByteTrack system, two uncertainty estimation metrics show promising performance in capturing the system's unstable performance on corrupted data. While in the ConvLab-2 system, we find that these uncertainty estimation techniques are only effective in analyzing NLU (Natural language understanding) module. Different from other systems, in ConvLab-2, each module in the system will be executed several times in a complete dialogue loop. Certain modules like DST (dialogue state tracking) and POL (dialogue policy) will maintain complex internal state information throughout the dialogue. Thus the system becomes much more complex compared with other ones. These results imply that more advanced uncertainty estimation techniques are needed when analyzing a more complex AI system.

Nevertheless, we also find that not every module's OOD or uncertainty information in an AI system can be used as indicators of the system's potential risks. For instance, in the SpeechBrain system, both the OOD and uncertainty scores from the second module have a strong correlation with system output error. However, there is almost no correlation when it comes to the first module. These results reveal that, when using OOD or uncertainty to do a risk assessment, it's important for developers to decide which module in the system should be measured. Besides, even though the second module shows more obvious effects on the system output error, it's still unclear how the output from the previous module has affected it. When the system triggers a prediction error, how to do fault localization inside the system can be another promising research direction.

\subsection{More advanced risk assessment techniques are needed to enhance the AI system's quality.}
In RQ4, as an early attempt or AI system risk assessment, we use simple strategies of combining OOD and uncertainty information to reject data instances that the system might not be able to handle. Our results show that even such a simple strategy can further improve the system's performance against common corruption patterns by up to 38\% (Table~\ref{tab:speech_improve1}~\&~\ref{tab:speech_improve2}). These empirical study results show the potential of making risk assessments for improving the AI system's robustness, as well as the potential of using OOD detection techniques and uncertainty estimation as risk assessment approaches. While in future work, more complex strategies can be explicitly designed for different AI systems and usage scenarios toward building a more reliable AI system. 

Besides, in this study, we try to focus on making the risk assessment from the data distribution and uncertainty perspectives. However, the general AI system quality assurance can also be achieved from some other perspectives, \eg, leveraging testing and debugging methods to identify an AI system's ability boundaries. In the unit-level analysis, DNN debugging and testing techniques have been widely used to improve a single DNN model's quality and reliability. Yet there is neither an exploratory study on how these unit-level testing/debugging approaches perform on system-level analysis nor methods specifically designed for AI systems at multiple levels. Therefore, we believe studying the quality assurance of AI systems from multiple angles and levels could be another important research direction.

\section{Threat to Validity}\label{sec:threat}

In this section, we summarize the threats to the validity of our study.

{\noindent\textbf{Internal validity.}} The internal threats to validity mainly lie in both the choice of measurement metrics for OOD detection and uncertainty and the choice of performance metrics for different AI systems. For the measurement metrics, specifically Maha-d and KDE in OOD detection and VR and VRO in uncertainty, their effectiveness has mostly been shown in image classification tasks but not in other domains. They may not be suitable for tasks like audio and object tracking. Besides, for modules that do not directly handle prediction tasks, e.g., the first module of the SpeechBrain system, uncertainty calculation may not be representative enough. To mitigate these threats, we first extend the definition of uncertainty measurement in Section~\ref{subsec:analysis_techniques} and perform comprehensive experiments for all the measurement metrics in RQ1 and RQ2, demonstrating that they are indeed capable of capturing the differences between clean data and corrupted data regardless of different kinds of modules. For the performance metrics, if they are not chosen properly, there may be bias when evaluating the AI systems. To reduce the bias, we report the standard metrics commonly used in their own domains.

{\noindent\textbf{Construct validity.}} The construct threats mainly lie in the randomness inherent in our experiments. Specifically, the measurement of uncertainty for AI systems and creating corrupted datasets both incur randomness in our evaluation. To reduce the influence of randomness on uncertainty measurement, we repeat our evaluations on a large number of corrupted test instances for multiple different configurations of each AI system. We additionally analyze the results using two uncertainty metrics, namely VR and VRO. To reduce the influence of randomness in the data corruption process, we select multiple (\ie, three) different levels for each corruption and apply the corruption to diverse datasets. We further perform a large-scale statistical analysis using tests like {\em Wilcoxon signed-rank test} and {\em Kendall's $\tau$ test} on the results to reduce the threats.

{\noindent \textbf{External validity.}} The external threats to validity include the selection of AI systems and the selection of data corruption patterns. Improper selection may hinder the generalization of the results. To mitigate the threat, we first perform a survey on available AI systems and select representative systems across diverse input domains and tasks under the criteria described in Sec.~\ref{subsec:aisystems}. To reduce the bias of the AI systems towards one specific corruption pattern, we select 5 common image corruption patterns, 5 text corruption patterns and 5 audio corruption patterns with three different severity levels. Overall, our large-scale evaluation takes 5000+ (NVIDIA A4000 16GB VRAM) GPU hours, on which we tried our greatest extent based on the available computational resource.
We hope the results obtained through such a large-scale evaluation could be helpful to inspire future research toward designing more advanced quality assurance techniques for building trustworthy AI systems. Finally, our research process and empirical results on selected AI systems might not generalize to all AI systems. To mitigate this threat, we tried our best to select AI systems from four different domains.

\section{Conclusion}\label{sec:conclusion}

In this paper, we propose an exploratory study of AI system risk assessment. We first investigate a wide range of AI-enabled systems and select four representative ones from different domains and applications as our study subjects: Udacity~\cite{udacitychallenge}, ByteTrack~\cite{zhang2021bytetrack}, SpeechBrain~\cite{ravanelli2021speechbrain}, and ConvLab-2~\cite{lee2019convlab}. Then we leverage the common corruption patterns from different data domains, \ie, image, audio, and text, to create large-scale corrupted datasets to evaluate these AI systems' robustness. Overall, we find that these common corruption patterns pose certain quality and reliability threats to AI systems. As an early exploratory study of AI systems at the system level, we then use two general OOD detection techniques and two uncertainty estimation metrics to analyze the AI systems. We find that these two different aspects' techniques are still effective in analyzing different modules in the system. We further investigate the error propagation in AI systems. We also find that not every module in the AI system strongly correlates to the system output error. Based on these experimental results and findings, as an early attempt at AI system risk assessment, we use several simple combination strategies of OOD and uncertainty information to show their potential to improve the reliability of AI systems. At the end of this paper, we also discuss several implications from our exploratory study as well as future research opportunities.

As an early step in studying the complex AI system, our work confirms that the quality assurance of AI systems is an urgent demand. Yet the effectiveness of unit-level analysis techniques can be either limited or hasn't been demonstrated. Considering AI systems have been widely used in real-world environments, we call for software engineering researchers and practitioners’ attention to step forward to not only focusing on the unit (model) level but also on a more advanced system level which is a largely explored realm. When moving towards the AI system level, we believe experience and practice from the software engineering community could bring much potential and played a key important role towards building safe, reliable and trustworthy AI systems.

%
%
%
\bibliographystyle{splncs04}
\bibliography{ref_llncs}

\end{document}